\documentclass[journal]{vgtc}                % final (journal style)
\ifpdf%                                % if we use pdflatex
  \pdfoutput=1\relax                   % create PDFs from pdfLaTeX
  \usepackage{graphicx}                % allow us to embed graphics files
  \DeclareGraphicsExtensions{.pdf,.png,.jpg,.jpeg} % for pdflatex we expect .pdf, .png, or .jpg files
\else%                                 % else we use pure latex
  \usepackage{graphicx}                % allow us to embed graphics files
  \DeclareGraphicsExtensions{.eps}     % for pure latex we expect eps files
\fi%

\usepackage{microtype}                 % use micro-typography (slightly more compact, better to read)
\PassOptionsToPackage{warn}{textcomp}  % to address font issues with \textrightarrow
\usepackage{textcomp}                  % use better special symbols
\usepackage{mathptmx}                  % use matching math font
\usepackage{times}                     % we use Times as the main font
\usepackage{cite}                      % needed to automatically sort the references
\usepackage{tabu}                      % only used for the table example
\usepackage{booktabs}                  % only used for the table example
\usepackage{url}

\usepackage{amsfonts}
\usepackage{amsmath}
\usepackage{amssymb}
\usepackage[title]{appendix}

\usepackage{color}
\usepackage{xcolor}

\definecolor{gray}{gray}{0.5}

\graphicspath{{Figures/}{./}} % where to search for the images

\newcommand{\SE}{{\mathcal{H}}} % Shannon entropy
\newcommand{\CE}{{\mathcal{H}_{\text{CE}}}} % cross entropy
\newcommand{\MI}{{\mathcal{I}}} % mutual information
\newcommand{\DKL}{{\mathcal{D}_{\text{KL}}}} % KL-divergence
\newcommand{\DJS}{{\mathcal{D}_{\text{JS}}}} % JS-divergence
\newcommand{\Dnew}{{\mathcal{D}^k_{\text{new}}}} % the new divergence by Chen and Sbert
\newcommand{\Dncm}{{\mathcal{D}^k_{\text{ncm}}}} % the new asymmetric divergence by Chen and Sbert
\newcommand{\DM}{{D^k_{\text{M}}}} %

\newcommand{\DnewA}{{\mathcal{D}^{k=1}_{\text{new}}}}
\newcommand{\DnewB}{{\mathcal{D}^{k=2}_{\text{new}}}}
\newcommand{\DncmA}{{\mathcal{D}^{k=1}_{\text{ncm}}}}
\newcommand{\DncmB}{{\mathcal{D}^{k=2}_{\text{ncm}}}}

\onlineid{0}

\vgtccategory{Research}
\vgtcpapertype{Theory and Model}

\title{A Bounded Measure for Estimating the Benefit of Visualization}

\author{%
Min Chen, \textit{Member, IEEE},
Mateu Sbert,
Alfie Abdul-Rahman, \textit{Member, IEEE},
Deborah Silver
}
\authorfooter{
\item
Min Chen is with University of Oxford, UK --- min.chen@oerc.ox.ac.uk
\item
Mateu Sbert is with University of Girona, Spain --- mateu@ima.udg.edu
\item
Alfie Abdul-Rahman is with King's College London, UK --- alfie.abdulrahman@kcl.ac.uk
\item
Deborah Silve is with Rutgers University, USA --- silver@jove.rutgers.edu
}

\shortauthortitle{M. Chen et al.: A Bounded Measure for Estimating the Benefit of Visualization}
\abstract{%
Information theory can be used to analyze the cost-benefit of visualization processes. However, the current measure of benefit contains an unbounded term that is neither easy to estimate nor intuitive to interpret. In this work, we propose to revise the existing cost-benefit measure by replacing the unbounded term with a bounded one. We examine a number of bounded measures that include the Jenson-Shannon divergence and a new divergence measure formulated as part of this work. We use visual analysis to support the multi-criteria comparison, narrowing the search down to those options with better mathematical properties.
We apply those remaining options to two visualization case studies to instantiate their uses in practical scenarios, while the collected real world data further informs the selection of a bounded measure, which can be used to estimate the benefit of visualization.
} % end of abstract


\teaser{
  \centering
  \begin{tabular}{@{}c@{\hspace{4mm}}c@{}}
    \includegraphics[height=40mm]{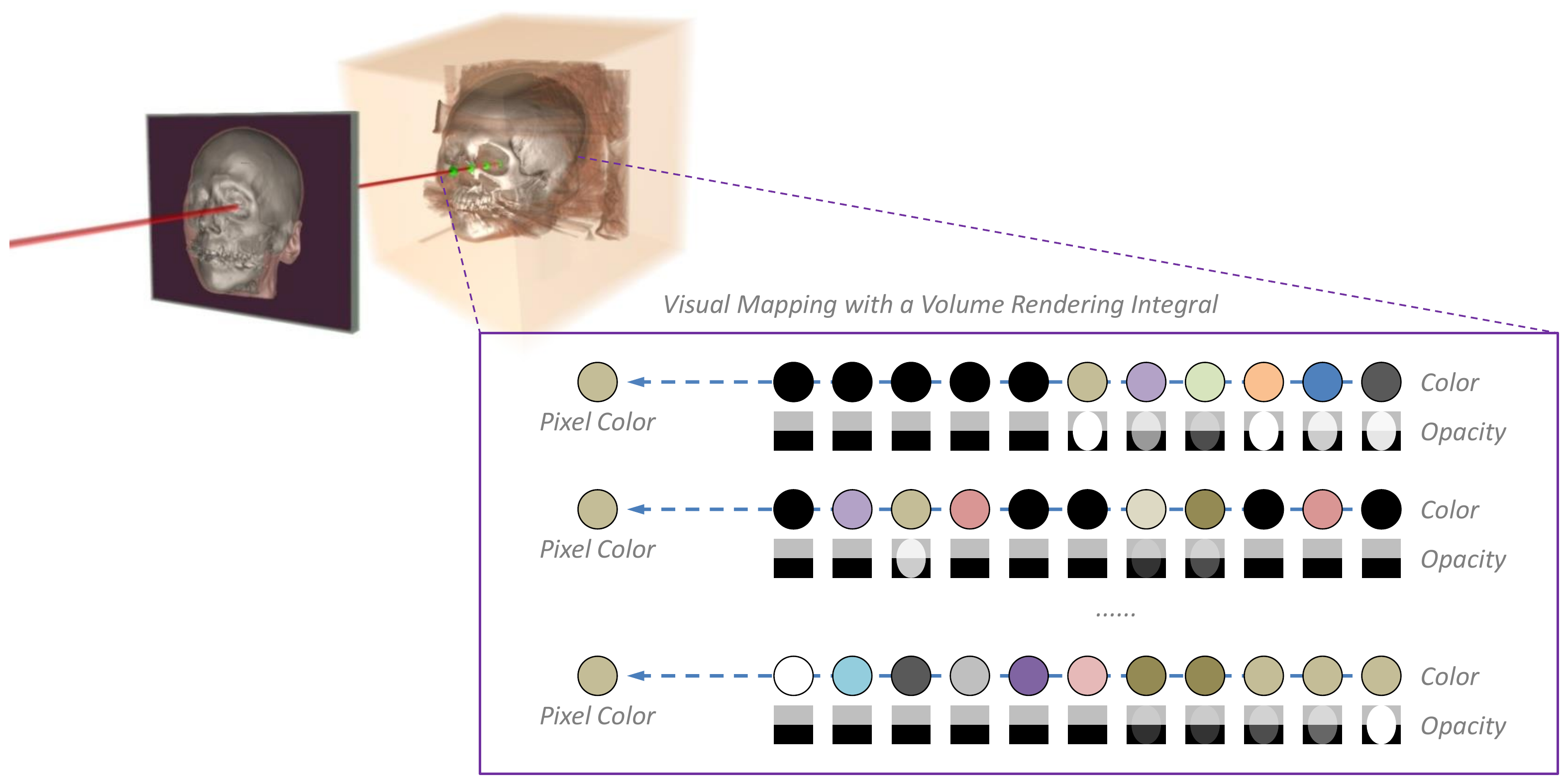} &
    \includegraphics[height=40mm]{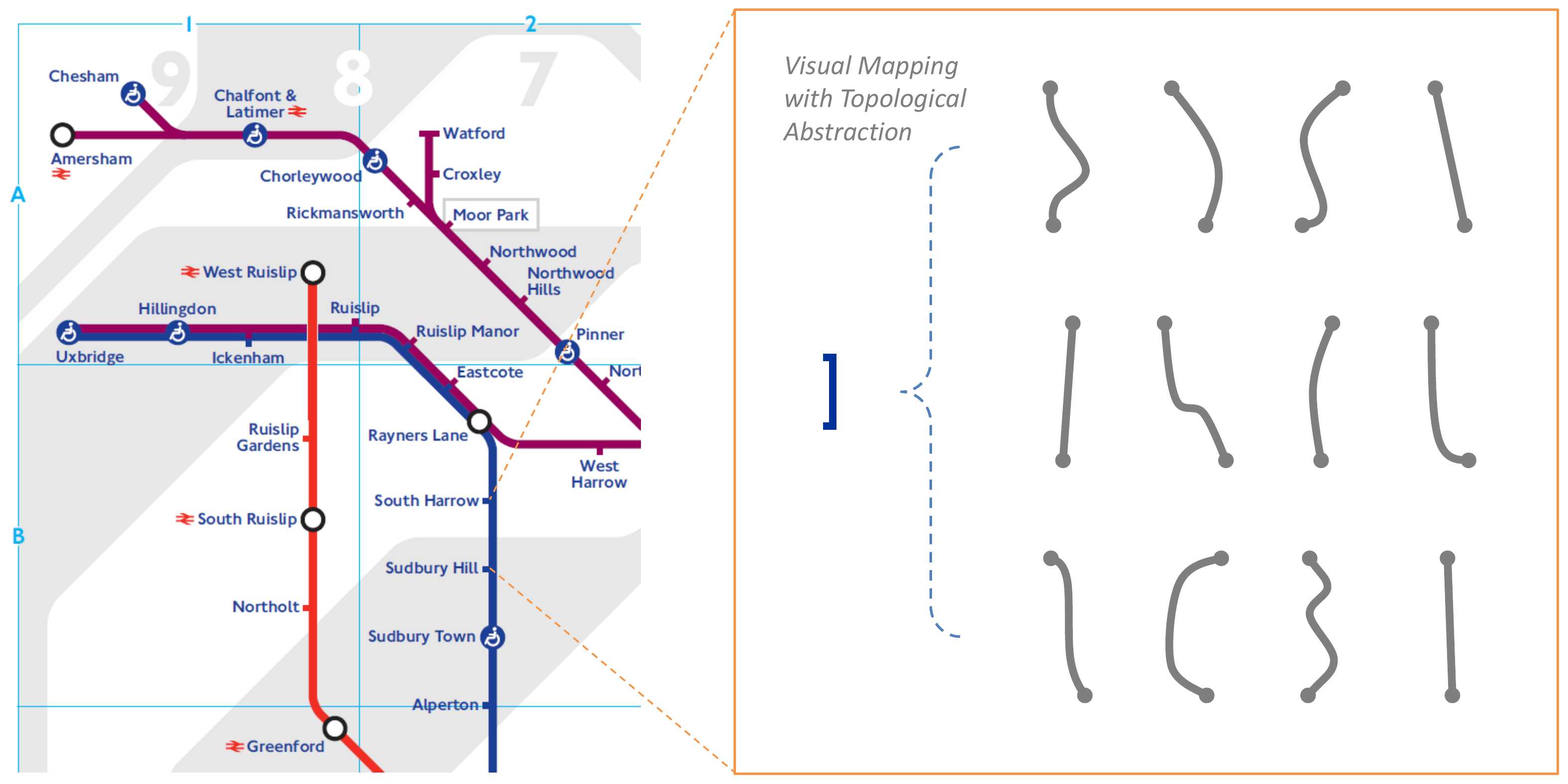}\\
    \small{(a) mapping from different sets of voxel values to the same pixel color} &
    \small{(b) mapping from different geographical paths to the same line segment}
  \end{tabular}
  \caption{Visual encoding typically features many-to-one mapping from data to visual representations, hence information loss. The significant amount of information loss in volume visualization and metro maps suggests that viewers not only can abide the information loss but also benefit from it. Measuring such benefits can lead to new advancements of visualization, in theory and practice.}
  \label{fig:InfoLoss}
}

\begin{document}

% \orcid{0000-0001-5320-5729}
% <concept_id>10003120.10003121.10003126</concept_id>
% <concept_desc>Human-centered computing~HCI theory, concepts and models</concept_desc>
% <concept_id>10002950.10003712</concept_id>
% <concept_desc>Mathematics of computing~Information theory</concept_desc>
% <concept_id>10002951.10003227.10003241.10003244</concept_id>
% <concept_desc>Information systems~Data analytics</concept_desc>
% <concept_significance>300</concept_significance>
% <concept_id>10003120.10003145.10011768</concept_id>
% <concept_desc>Human-centered computing~Visualization theory, concepts and paradigms</concept_desc>
% <concept_significance>100</concept_significance>
% \ccsdesc[500]{Human-centered computing~HCI theory, concepts and models}
% \ccsdesc[500]{Mathematics of computing~Information theory}
% \ccsdesc[300]{Information systems~Data analytics}
% \ccsdesc[100]{Human-centered computing~Visualization theory, concepts and paradigms}


\firstsection{Introduction\label{sec:Introduction}}
\maketitle

To most of us, it seems rather intuitive that visualization should be accurate, different data values should be visually encoded differently, and visual distortion should be disallowed.
However, when we closely examine most (if not all) visualization images, we can notice that inaccuracy is ubiquitous.
The two examples in Fig. \ref{fig:InfoLoss} evidence the presence of such inaccuracy.
In volume visualization, when a pixel is used to depict a set of voxels along a ray, many different sets of voxel values may result in the same pixel color.
In a metro map, a variety of complex geographical paths may be distorted and depicted as a straight line.
Since there is little doubt that volume visualization and metro maps are useful, some ``inaccurate'' visualization must be beneficial.

% It is now widely understood among visualization researchers and practitioners that the effectiveness of a visualization process depends on \emph{data}, \emph{user}, and \emph{task}.
% One important aspect of \emph{user} is a user's knowledge, which plays a critical role in reconstructing the information lost during visualization processes (e.g., data transformation and visual mapping).
% One major challenge in appreciating the significance of such knowledge is the difficulty to measure or estimate the knowledge used by a user during visualization.

In terms of information theory, the types of inaccuracy featured in Fig. \ref{fig:InfoLoss} are different forms of information loss (or many-to-one mapping).
Chen and Golan proposed an information-theoretic measure \cite{Chen:2016:TVCG} for analyzing the cost-benefit of data intelligence workflows.
It enables us to consider the positive impact of information loss (e.g., reducing the cost of storing, processing, displaying, perceiving, and reasoning about the information) as well as its negative impact (e.g., being mislead by the information).
The measure provides a concise explanation about the benefit of visualization because visualization and other data intelligence processes (e.g., statistics and algorithms) all typically cause information loss and visualization allows human users to reduce the negative impact of information loss effectively using their knowledge. 

The mathematical formula of the measure features a term based on the Kullback-Leibler (KL) divergence \cite{Kullback:1951:AMS} for measuring the potential distortion of a user or a group of users in reconstructing the information that may have been lost or distorted during a visualization process.
% The cost-benefit ratio instigates that a user with more knowledge about the source data and its visual representation is likely to suffer less distortion.
While using the KL-divergence is mathematically intrinsic for measuring the potential distortion, its unboundedness property has some undesirable consequences.
The simplest phenomenon of making a false representation (i.e., always displaying 1 when a binary value is 0 or always 0 when it is 1) happens to be a singularity condition of the KL-divergence.
The amount of distortion measured by the KL-divergence often has much more bits than the entropy of the information space itself.
This is not intuitive to interpret and hinders practical applications.%

% Kijmongkolchai et al. applied the formula of Chen and Golan to the results of an empirical study for estimating users' knowledge used in visualization processes, and used a bounded approximation of the KL-divergence in their estimation \cite{Kijmongkolchai:2017:CGF}.

In this work, we propose to replace the KL-divergence with a bounded term.
We first confirm the boundedness is a necessary property.
We then conduct multi-criteria decision analysis (MCDA) \cite{Ishizaka:2013:book} to compare a number of bounded measures, which include the Jensen–Shannon (JS) divergence \cite{Lin:1991:TIT} and a new divergence measure, $\Dnew$, formulated as part of this work.
We use visual analysis to aid the observation of the mathematical properties of these candidate measures, narrowing down from eight options to five.
We use two visualization case studies to instantiate values that may be returned by the remaining options.
The numerical calculations in visualization contexts further inform us about the relative merits of these measures, enabling us to make the final selection while demonstrating its uses in practical scenarios.

% ====================
\section{Related Work}
\label{sec:RelatedWork}

Claude Shannon's landmark article in 1948 \cite{Shannon:1948:BSTJ} signifies the birth of information theory.
It has been underpinning the fields of data communication, compression, and encryption since.
As a mathematical framework, information theory provides a collection of useful measures, many of which, such as Shannon entropy \cite{Shannon:1948:BSTJ}, cross entropy \cite{Cover:2006:book}, mutual information \cite{Cover:2006:book}, and Kullback-Leibler divergence \cite{Kullback:1951:AMS} are widely used in applications of
physics, biology, neurology, psychology, and computer science
(e.g., visualization, computer graphics, computer vision, data mining, machine learning), and so on.
In this work, we also consider Jensen-Shannon divergence \cite{Lin:1991:TIT} in detail.

Information theory has been used extensively in visualization \cite{Chen:2016:book}.
It has enabled many applications in visualization, including
scene and shape complexity analysis by Feixas et al. \cite{Feixas:2001:CGF} and Rigau et al. \cite{Rigau:2005:SMA},
light source placement by Gumhold \cite{Gumhold:2002:Vis},
view selection in mesh rendering by V\'{a}zquez et al. \cite{Vazquez:2004:CGF} and Feixas et al. \cite{Feixas:2009:AP},
attribute selection by Ng and Martin \cite{Ng:2004:IV},
view selection in volume rendering by Bordoloi and Shen \cite{Bordoloi:2005:Vis}, and Takahashi and Takeshima \cite{Takahashi:2005:Vis},
multi-resolution volume visualization by Wang and Shen \cite{Wang:2006:TVCG},
focus of attention in volume rendering by Viola et al. \cite{Viola:2006:TVCG},
feature highlighting by J\"anicke and Scheuermann \cite{Jaenicke:2007:TVCG,Jaenicke:2010:CGA},
    and Wang et al. \cite{Wang:2008:TVCG},
transfer function design by Bruckner and M\"{o}ller \cite{Bruckner:2010:CGF},
	and Ruiz et al. \cite{Ruiz:2011:TVCG,Bramon:2013:JBHI},
multi-modal data fusion by Bramon et al. \cite{Bramon:2012:TVCG},
isosurface evaluation by Wei et al. \cite{Wei:2013:CGF},
measuring observation capacity by Bramon et al. \cite{Bramon:2013:CGF},
measuring information content by Biswas et al. \cite{Biswas:2013:TVCG},
proving the correctness of ``overview first, zoom, details-on-demand'' by Chen and J\"anicke \cite{Chen:2010:TVCG} and Chen et al. \cite{Chen:2016:book}, and
confirming visual multiplexing by Chen et al. \cite{Chen:2014:CGF}.

Ward first suggested that information theory might be an underpinning theory for visualization \cite{Purchase:2008:LNCS}.
Chen and J\"anicke \cite{Chen:2010:TVCG} outlined an information-theoretic framework for visualization, and it was further enriched by Xu et al. \cite{Xu:2010:TVCG} and Wang and Shen \cite{Wang:2011:E} in the context of scientific visualization.
Chen and Golan proposed an information-theoretic measure for analyzing the cost-benefit of visualization processes and visual analytics workflows \cite{Chen:2016:TVCG}.
It was used to frame an observation study showing that human developers usually entered a huge amount of knowledge into a machine learning model \cite{Tam:2017:TVCG}.
It motivated an empirical study confirming that knowledge could be detected and measured quantitatively via controlled experiments \cite{Kijmongkolchai:2017:CGF}. 
It was used to analyze the cost-benefit of different virtual reality applications
\cite{Chen:2019:TVCG}.
It formed the basis of a systematic methodology for improving the cost-benefit of visual analytics workflows \cite{Chen:2019:CGF}.
It survived qualitative falsification by using arguments in visualization \cite{Streeb:2019:TVCG}.
It offered a theoretical explanation of ``visual abstraction'' \cite{Viola:2019:book}.
This work continues the path of theoretical developments in visualization \cite{Chen:2017:CGA}, and is intended to improve the original cost-benefit formula \cite{Chen:2016:TVCG}, in order to make it more intuitive and usable in practical visualization applications.

\begin{figure}[t]
  \centering
  \includegraphics[width=80mm]{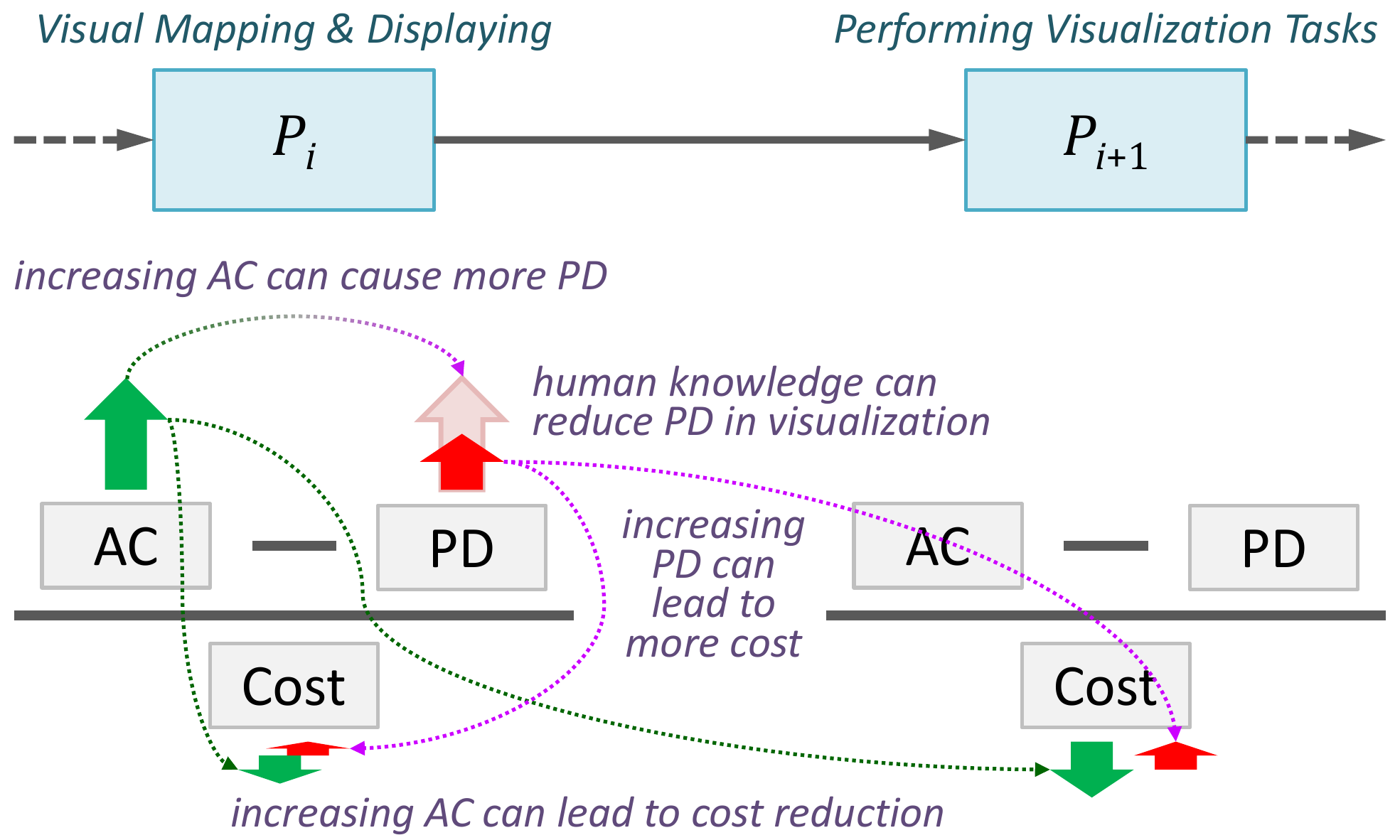}
  \caption{Alphabet compression may reduce the cost of $P_i$ and $P_{i+1}$, especially when human knowledge can reduce the potential distortion.}
  \label{fig:Benefit}
  \vspace{-4mm}
\end{figure}

% ====================
\section{Overview, Motivation, and Problem Statement}
\label{sec:Motivation}
Visualization is useful in most data intelligence workflows, but the usefulness is not universally true because the effectiveness of visualization is usually data-, user-, and task-dependent.
The cost-benefit ratio proposed by Chen and Golan \cite{Chen:2016:TVCG} captures the essence of such dependency.
Below is the qualitative expression of the measure:
\begin{equation} \label{eq:CBM-1}
    \frac{\text{Benefit}}{\text{Cost}} = \frac{\text{Alphabet Compression} - \text{Potential Distortion}}{\text{Cost}}
\end{equation}

Consider the scenario of viewing some data through a particular visual representation.
The term \emph{Alphabet Compression} (AC) measures the amount of information loss due to visual abstraction \cite{Viola:2019:book}.
Since the visual representation is fixed in the scenario, AC is thus largely data-dependent.
AC is a positive measure reflecting the fact that visual abstraction must be useful in many cases though it may result in information loss.
This apparently counter-intuitive term is essential for asserting why visualization is useful. (Note that the term also helps assert the usefulness of statistics, algorithms, and interaction since they all usually cause information loss \cite{Chen:2019:CGF}. See also Appendix \ref{app:OriginalTheory}.)

The positive implication of the term AC is counterbalanced by the term \emph{Potential Distortion}, while both being moderated by the term \emph{Cost}.
The term \emph{Cost} encompasses all costs of the visualization process, including computational costs (e.g., visual mapping and rendering), cognitive costs (e.g., cognitive load), and consequential costs (e.g., impact of errors).
As illustrated in Fig. \ref{fig:Benefit}, increasing AC typically enables the reduction of cost (e.g., in terms of energy, time, or money).
% The measure of cost (e.g., in terms of energy, time, or money) is thus data-, user-, and task-dependent.

The term \emph{Potential Distortion} (PD) measures the informative divergence between viewing the data through visualization with information loss and reading the data without any information loss. The latter might be ideal but is usually at an unattainable cost except for values in a very small data space (i.e., in a small alphabet as discussed in \cite{Chen:2016:TVCG}).
As shown in Fig. \ref{fig:Benefit}, increasing AC typically causes more PD.
PD is data-dependent or user-dependent.
Given the same data visualization with the same amount of information loss, one can postulate that a user with more knowledge about the data or visual representation usually suffers less distortion.
This postulation is a focus of this paper.

\begin{figure}[t]
\centering
\includegraphics[width=\linewidth]{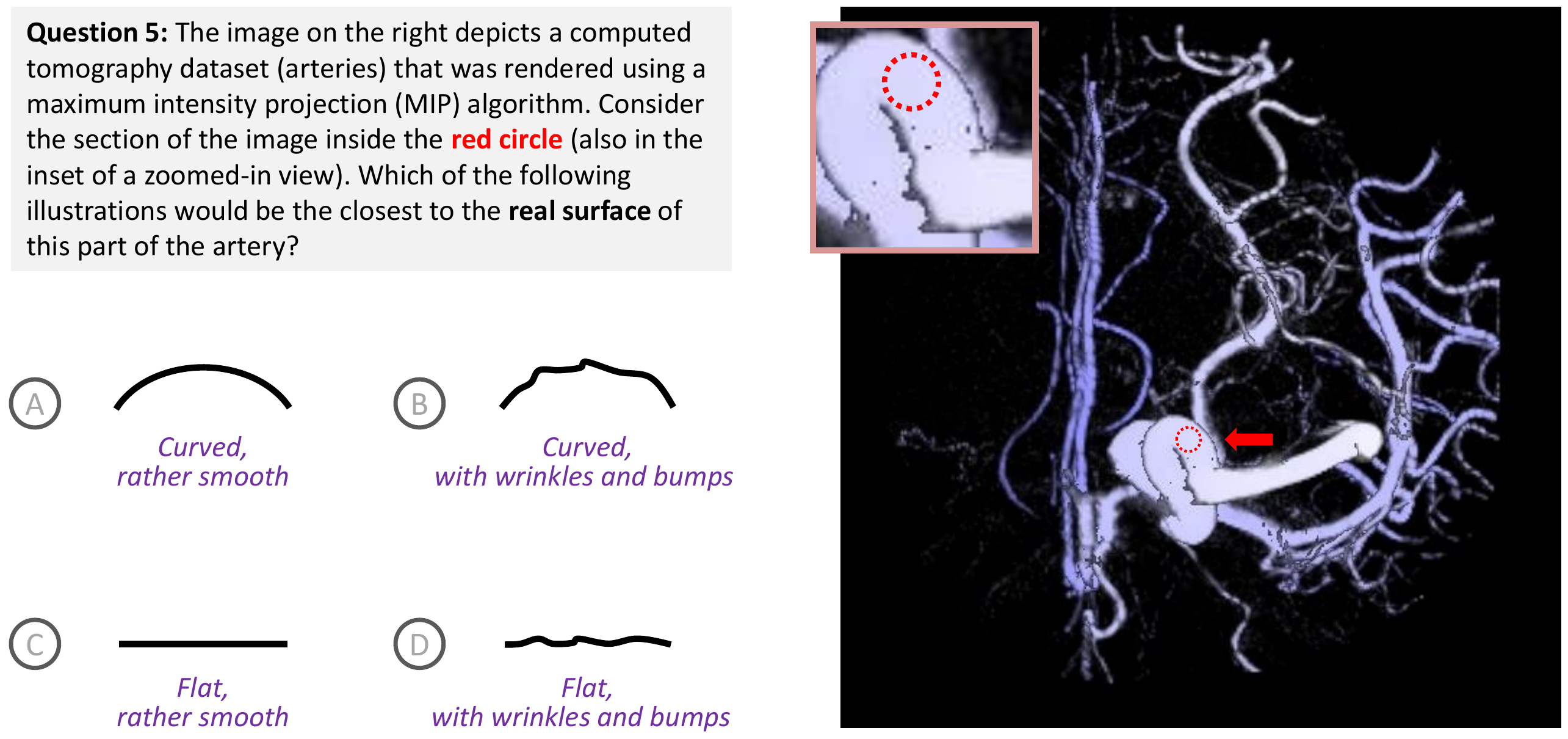}
\caption{A volume dataset was rendered using the MIP method. A question about a ``flat area'' in the image can be used to tease out a viewer's knowledge that is useful in a visualization process.}
\label{fig:Arteries}
\vspace{-4mm}
\end{figure}

Consider the visual representation of a network of arteries in Fig. \ref{fig:Arteries}.
The image was generated from a volume dataset using the maximum intensity projection (MIP) method.
While it is known that MIP cannot convey depth information well, it has been widely used for observing some classes of medical imaging data, such as arteries.
The highlighted area in Fig. \ref{fig:Arteries} shows an apparently flat area, which is a distortion from the actuality of a tubular surface likely with some small wrinkles and bumps.
The doctors who deal with such medical data are expected to have sufficient knowledge to reconstruct the reality adequately from the ``distorted'' visualization, while being able to focus on the more important task of making diagnostic decisions, e.g., about aneurysm.

As shown in some recent works, it is possible for visualization designers to estimate AC, PD, and Cost qualitatively \cite{Chen:2019:TVCG,Chen:2019:CGF} and quantitatively \cite{Tam:2017:TVCG,Kijmongkolchai:2017:CGF}.
It is highly desirable to advance the scientific methods for quantitative estimation, towards the eventual realization of computer-assisted analysis and optimization in designing visual representations.
This work focuses on one challenge of quantitative estimation, i.e., how to estimate the benefit of visualization to human users with different knowledge about the depicted data and visual encoding.

Building on the methods of observational estimation in \cite{Tam:2017:TVCG} and controlled experiment in \cite{Kijmongkolchai:2017:CGF}, one may reasonably anticipate a systematic method based on a short interview by asking potential viewers a few questions.
For example, one may use the question in Fig. \ref{fig:Arteries} to estimate the knowledge of doctors, patients, and any other people who may view such a visualization.
The question is intended to tease out two pieces of knowledge that may help reduce the potential distortion due to the ``flat area'' depiction.
One piece is about the general knowledge that associates arteries with tube-like shapes.
Another, which is more advanced, is about the surface texture of arteries and the limitations of the MIP method.

Let the binary options about whether the ``flat area'' is actually flat or curved be an alphabet $\mathbb{A} = \{ \textit{curved}, \textit{flat} \}$.
The likelihood of the two options is represented by a probability distribution or probability mass function (PMF) $P(\mathbb{A}) = \{1-\epsilon, 0+\epsilon\}$, where $0<\epsilon<1$.
Since most arteries in the real world are of tubular shapes, one can imagine that a ground truth alphabet $\mathbb{A}_\text{G.T.}$ might have a PMF $P(\mathbb{A}_\text{G.T.})$ strongly in favor of the \textit{curved} option.
However, the visualization seems to suggest the opposite, implying a PMF $P(\mathbb{A}_\text{MIP})$ strongly in favor of the \textit{flat} option.
It is not difficult to interview some potential viewers, enquiring how they would answer the question.
One may estimate a PMF $P(\mathbb{A}_\text{doctors})$ from doctors' answers, and another $P(\mathbb{A}_\text{patients})$ from patients' answers.

\begin{table}[t]
\caption{Imaginary scenarios where probability data is collected for estimating knowledge related to alphabet
$\mathbb{A} = \{ \textit{curved}, \, \textit{flat} \}$.
The ground truth (G.T.) PMFs are defined with $\epsilon = 0.01$ and $0.0001$ respectively.
The potential distortion (as ``$\rightarrow$ value'') is computed using the KL-divergence.}
\label{tab:KL-ex1}
\centering
\begin{tabular}{@{}l@{\hspace{3mm}}l@{\hspace{4mm}}l@{}}
  & \textbf{Scenario 1} & \textbf{Scenario 2}\\
  \hline
  $Q(\mathbb{A}_\text{G.T.})$:
    & $\{0.99, 0.01\}$
    & $\{0.9999, 0.0001\}$ \\
  $P(\mathbb{A}_\text{MIP})$:
    & $\{0.01, 0.99\} \rightarrow 6.50$
    & $\{0.0001, 0.9999\} \rightarrow 13.28$\\
  $P(\mathbb{A}_\text{doctors})$:
    & $\{0.99, 0.01\} \rightarrow 0.00$
    & $\{0.99, 0.01\} \rightarrow 0.05$ \\
  $P(\mathbb{A}_\text{patients})$:
    & $\{0.7, 0.3\} \rightarrow 1.12$
    & $\{0.7, 0.3\} \rightarrow 3.11$\\
  \hline
\end{tabular}
\vspace{0mm}
\end{table}

\begin{table}[t]
\caption{Imaginary scenarios for estimating knowledge related to alphabet $\mathbb{B} = \{ \textit{wrinkles-and-bumps}, \, \textit{smooth} \}$.
The ground truth (G.T.) PMFs are defined with $\epsilon = 0.1$ and $0.001$ respectively.
The potential distortion (as ``$\rightarrow$ value'') is computed using the KL-divergence.}
\label{tab:KL-ex2}
\centering
\begin{tabular}{@{}l@{\hspace{7mm}}l@{\hspace{8mm}}l@{}}
  & \textbf{Scenario 3} & \textbf{Scenario 4}\\
  \hline
  $Q(\mathbb{B}_\text{G.T.})$:
    & $\{0.9, 0.1\}$
    & $\{0.001, 0.999\}$ \\
  $P(\mathbb{B}_\text{MIP})$:
    & $\{0.1, 0.9\} \rightarrow 2.54$
    & $\{0.001, 0.999\} \rightarrow 9.94$\\
  $P(\mathbb{B}_\text{doctors})$:
    & $\{0.8, 0.2\} \rightarrow 0.06$
    & $\{0.8, 0.2\} \rightarrow 1.27$ \\
  $P(\mathbb{B}_\text{patients})$:
    & $\{0.1, 0.9\} \rightarrow 2.54$
    & $\{0.1, 0.9\} \rightarrow 8.50$\\
  \hline
\end{tabular}
\vspace{-4mm}
\end{table}

Table \ref{tab:KL-ex1} shows two scenarios where different probability data is obtained. The values of PD are computed using the most well-known divergence measure, KL-divergence \cite{Kullback:1951:AMS}, and are of unit \emph{bit}.
In Scenario 1, without any knowledge, the visualization process would suffer 6.50 bits of PD.
As doctors are not fooled by the ``flat area'' shown in the MIP visualization, their knowledge is worth 6.50 bits.
Meanwhile, patients would suffer 1.12 bits of PD on average, their knowledge is worth $5.38 = 6.50 - 1.12$ bits.

In Scenario 2, the PMFs of $P(\mathbb{A}_\text{G.T.})$ and $P(\mathbb{A}_\text{MIP})$ depart further away, while $P(\mathbb{A}_\text{doctors})$ and $P(\mathbb{A}_\text{patients})$ remain the same.
Although doctors and patients would suffer more PD, their knowledge is worth more than that in Scenario 1 (i.e., $13.28 - 0.05 = 13.23$ bits and $13.28 - 3.11 = 10.17$ bits respectively).

Similarly, the binary options about whether the ``flat area'' is actually smooth or not can be defined by an alphabet
$\mathbb{A} = \{ \textit{wrinkles-and-bumps}, \, \textit{smooth} \}$.
Table \ref{tab:KL-ex2} shows two scenarios about collected probability data.
In these two scenarios, doctors exhibit much more knowledge than patients, indicating that the surface texture of arteries is a piece of specialized knowledge.

The above example demonstrates that using the KL-divergence to estimate PD can differentiate the knowledge variation between doctors and patients regarding the two pieces of knowledge that may reduce the distortion due to the ``flat area''.
When it is used in Eq.\, \ref{eq:CBM-1} in a relative or qualitative context (e.g., \cite{Chen:2019:TVCG,Chen:2019:CGF}), the unboundedness of the KL-divergence does not pose an issue.

However, this does become an issue when the KL-divergence is used to measure PD in an absolute and quantitative context.
From the two diverging PMFs $P(\mathbb{A}_\text{G.T.})$ and $P(\mathbb{A}_\text{MIP})$ in Table \ref{tab:KL-ex1}, or $P(\mathbb{B}_\text{G.T.})$ and $P(\mathbb{B}_\text{MIP})$ in Table \ref{tab:KL-ex2}, we can observe that the smaller $\epsilon$ is, the more divergent the two PMFs become and the higher value the PD has.
Indeed, consider an arbitrary alphabet $\mathbb{Z} = \{z_1, z_2 \}$, and two PMFs defined upon $\mathbb{Z}$: $P=[0+\epsilon, \, 1-\epsilon]$ and $Q=[1-\epsilon, \, 0+\epsilon]$.
When $\epsilon \rightarrow 0$, we have the KL-divergence $\DKL(Q||P) \rightarrow \infty$.

Meanwhile, the Shannon entropy of $\mathbb{Z}$, $\mathcal{H}(\mathbb{Z})$, has an upper bound of 1 bit.
It is thus not intuitive or practical to relate the value of $\DKL(Q||P)$ to that of $\mathcal{H}(\mathbb{Z})$.
Many applications of information theory do not relate these two types of values explicitly.
When reasoning such relations is required, the common approach is to impose a lower-bound threshold for $\epsilon$ (e.g., \cite{Kijmongkolchai:2017:CGF}).
However, there is yet a consistent method for defining such a threshold for various alphabets in different applications, while preventing a range of small or large values (i.e., $[0, \epsilon)$ or $(1-\epsilon, 1]$) in a PMF is often inconvenient in practice.
In the following section, we discuss several approaches to defining a bounded measure for PD.

Note: for an information-theoretic measure, we use an alphabet $\mathbb{Z}$ and its PMF $P$ interchangeably, e.g., $\SE(P(\mathbb{Z})) = \SE(P) = \SE(\mathbb{Z})$.

% =====================
\section{Bounded Measures for Potential Distortion (PD)} 
Let $\mathbf{P}_i$ be a process in a data intelligence workflow, $\mathbb{Z}_i$ be its input alphabet, and $\mathbb{Z}_{i+1}$ be its output alphabet.
$\mathbf{P}_i$ can be a human-centric process (e.g., visualization and interaction) or a machine-centric process (e.g., statistics and algorithms).
In the original proposal \cite{Chen:2016:TVCG}, the value of Benefit in Eq.\,\ref{eq:CBM-1} is measured using:
\begin{equation} \label{eq:CBM-2}
  \text{Benefit} = \text{AC} - \text{PD}
                 = \SE(\mathbb{Z}_i) - \SE(\mathbb{Z}_{i+1})
                 - \DKL(\mathbb{Z}'_i||\mathbb{Z}_i)
\end{equation}
\noindent where $\SE()$ is the Shannon entropy of an alphabet and $\DKL()$ is KL-divergence of an alphabet from a reference alphabet.
Because the Shannon entropy of an alphabet with a finite number of letters is bounded, AC, which is the entropic difference between the input and output alphabets, is also bounded.
On the other hand, as discussed in the previous section, PD is unbounded.
Although Eq.\,\ref{eq:CBM-2} can be used for relative comparison, it is not quite intuitive in an absolute context, and it is difficult to imagine that the amount of informative distortion can be more than the maximum amount of information available.

In this section, we present the unpublished work by Chen and Sbert \cite{Chen:2019:arXiv}, which shows mathematically that for alphabets of a finite size, the KL-divergence used in Eq.\,\ref{eq:CBM-2} should ideally be bounded.
In their arXiv report, they also outlined a new divergence measure and compare it with a few other bounded measures.
Building on initial comparison in \cite{Chen:2019:arXiv}, we use visualization in Section \ref{sec:NewMetric} and real world data in Section \ref{sec:CaseStudies} to assist the multi-criteria analysis and selection of a bounded divergence measure to replace the KL-divergence used in Eq.\,\ref{eq:CBM-2}.
Appendices \ref{app:InfoTheory} and \ref{app:Proof} provide more mathematical background and details.

% --------------------------------------------------
\subsection{A Mathematical Proof of Boundedness}
\label{sec:Proof}
Let $\mathbb{Z}$ be an alphabet with a finite number of letters, $\{ z_1, z_2, \ldots, z_n \}$, and $\mathbb{Z}$ is associated with a PMF, $Q$, such that: 
\begin{equation} \label{eq:CodeP}
\begin{split}
  q(z_n) &= \epsilon, \quad\text{(where $0 < \epsilon < 2^{-(n-1)}$}),\\
  q(z_{n-1}) &= (1-\epsilon)2^{-(n-1)},\\
  q(z_{n-2}) &= (1-\epsilon)2^{-(n-2)},\\
  &\cdots\\
  q(z_{2}) &= (1-\epsilon)2^{-2},\\
  q(z_{1}) &= (1-\epsilon)2^{-1} + (1-\epsilon)2^{-(n-1)}.
\end{split}
\end{equation}
When we encode this alphabet using an entropy binary coding scheme \cite{Moser:2012:book}, we can be assured to achieve an optimal code with the lowest average length for
codewords. One example of such a code for the above probability is:
\begin{equation} \label{eq:Code}
\begin{split}
  z_1&: 0, \qquad z_2: 10, \qquad z_3: 110\\
  &\cdots\\
  z_{n-1}&: 111\ldots10 \quad\text{(with $n-2$ ``1''s and one ``0'') }\\
  z_n&: 111\ldots11 \quad\text{(with $n-1$ ``1''s and no ``0'') }
\end{split}
\end{equation}
In this way, $z_n$, which has the smallest probability, will always be assigned a codeword with the maximal length of $n-1$.
Entropy coding is designed to minimize the average number of bits per letter when one transmits a ``very long'' sequence of letters in the alphabet over a communication channel.
Here the phrase ``very long'' implies that the string exhibits the above PMF $Q$ (Eq.\,\ref{eq:CodeP}).

Suppose that $\mathbb{Z}$ is actually of PMF $P$, but is encoded as Eq.\,\ref{eq:Code} based on $Q$.
The transmission of $\mathbb{Z}$ using this code will have inefficiency.
The inefficiency is usually measured using cross entropy $\CE(P, Q)$:
\begin{equation} \label{eq:CrossEntropy}
   \CE(P, Q) = \SE(P) + \DKL(P||Q)
\end{equation}
Clearly, the worst case is that the letter, $z_n$, which was encoded using $n-1$ bits, turns out to be the most frequently used letter in $P$ (instead of the least in $Q$).
It is so frequent that all letters in the long string are of $z_n$.
So the average codeword length per letter of this string is $n-1$.
The situation cannot be worse.
Therefore, $n-1$ is the upper bound of the cross entropy.
From Eq.\,\ref{eq:CrossEntropy}, we can also observe that $\DKL(P||Q)$ must also be bounded since $\CE(P, Q)$ and $\SE(P)$ are both bounded as long as $\mathbb{Z}$ has a finite number of letters.
Let $\top_{\text{CE}}$ be the upper bound of $\CE(P, Q)$.
The upper bound for $\DKL(P||Q)$, $\top_{\text{KL}}$, is thus:
\begin{equation}\label{eq:CEandKL}
	\DKL(P||Q) = \CE(P, Q) - \SE(P) \le \top_{\text{\text{CE}}} - \min_{\forall P(\mathbb{Z})}\bigl( \SE(P) \bigr)
\end{equation}

There is a special case worth noting. In practice, it is common to
assume that $Q$ is a uniform distribution, i.e., $q_i = 1/n, \forall q_i \in Q$, typically because $Q$ is unknown or varies frequently.
Hence the assumption leads to a code with an average length equaling $\log_2 n$ (or in practice, the smallest integer $\ge \log_2 n$).
Under this special (but rather common) condition, all letters in a very long string have codewords of the same length.
The worst case is that all letters in the string turn out to the same letter. Since there is no informative variation in the PMF $P$ for this very long string, i.e., $\mathcal{H}(P) = 0$, in principle, the transmission of this string is unnecessary.
The maximal amount of inefficiency is thus $\log_2 n$.
This is indeed much lower than the upper bound $\top_{\text{CE}} = n-1$, justifying the assumption or use of a uniform $Q$ in many situations.

% --------------------
\begin{figure*}[t]
\centering
\begin{tabular}{@{}c@{\hspace{2mm}}c@{\hspace{2mm}}c@{\hspace{2mm}}c@{}}
    \includegraphics[width=42mm]{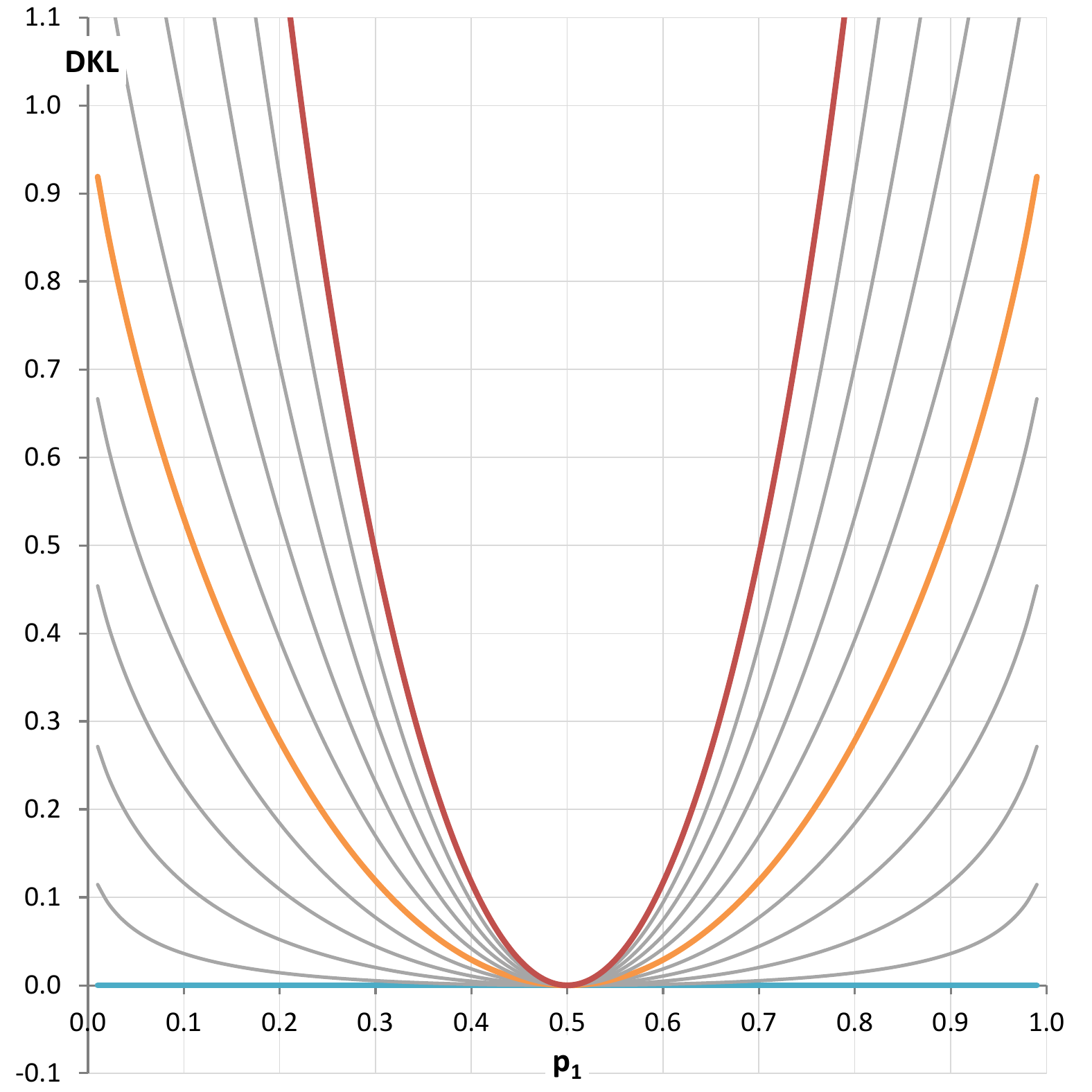} &
    \includegraphics[width=42mm]{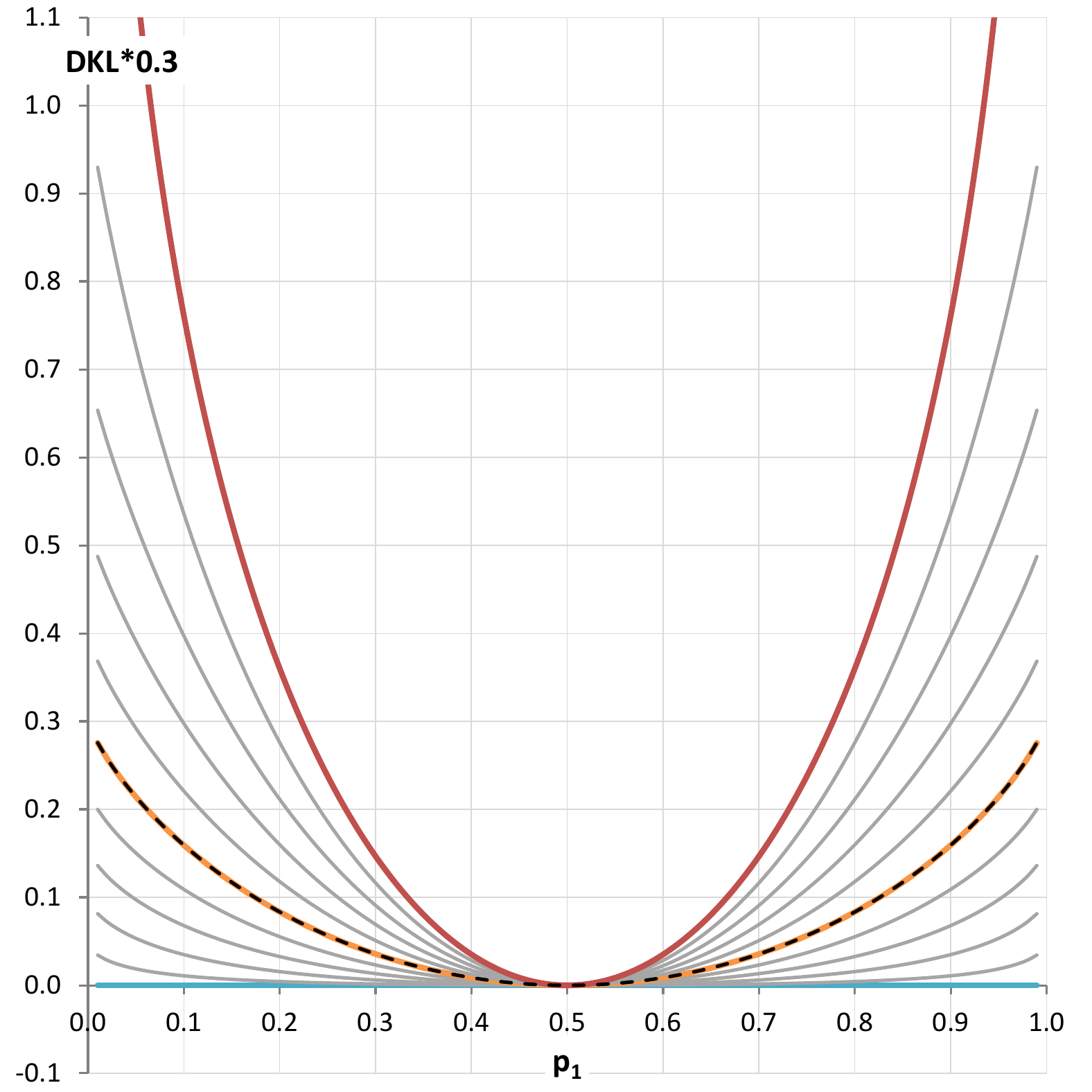} &
    \includegraphics[width=42mm]{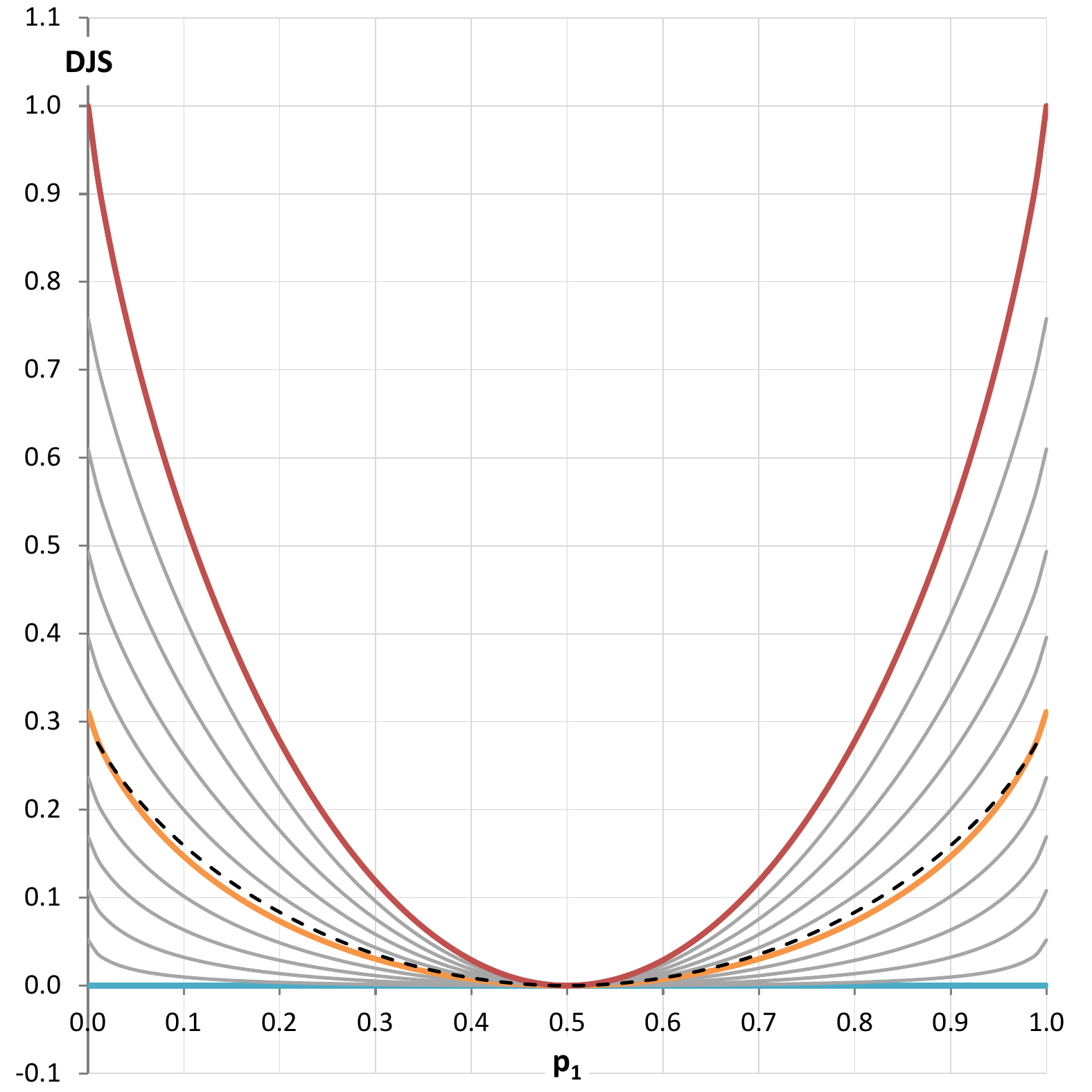} &
    \includegraphics[width=42mm]{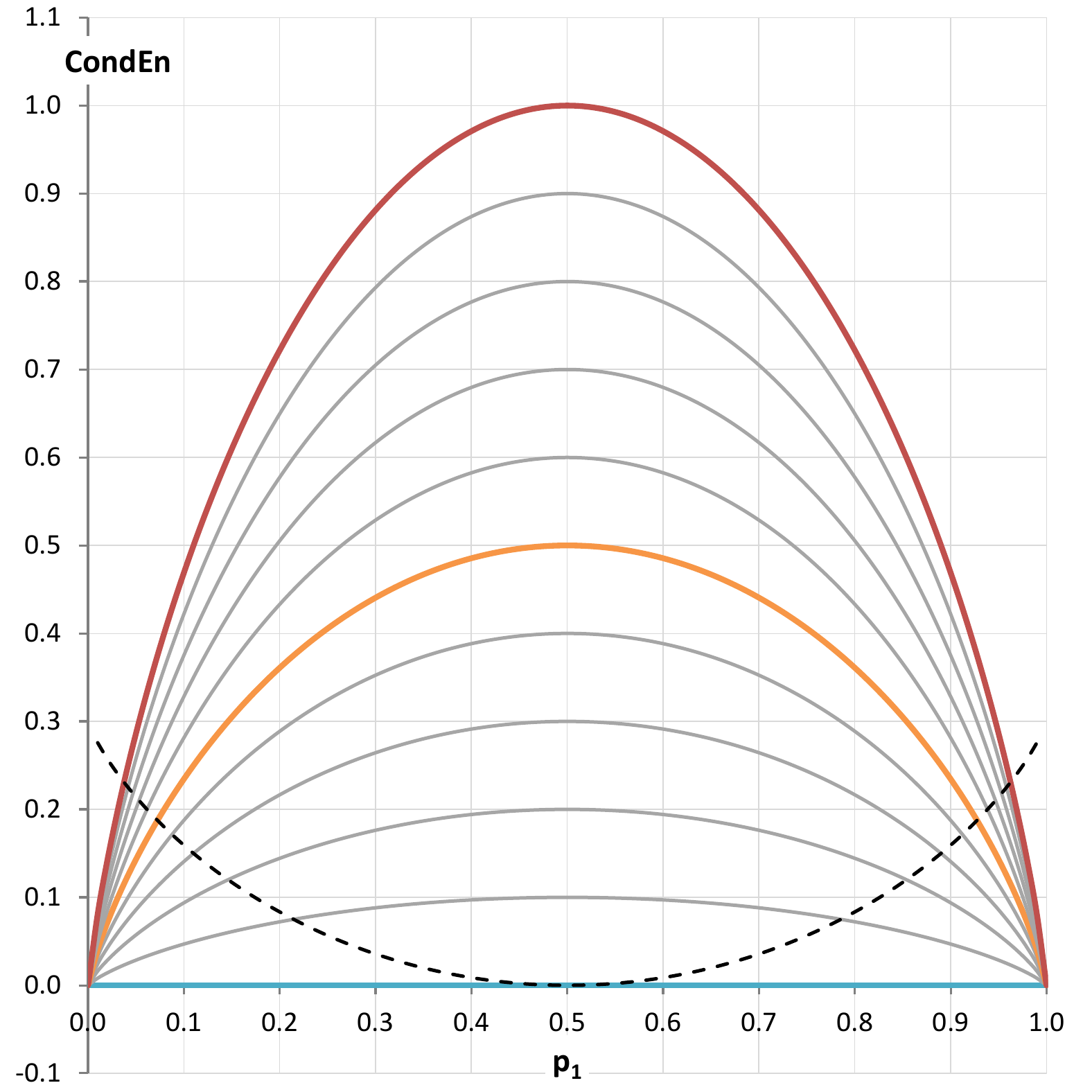} \\
    (a) $\DKL(P||Q)$ & (b) $0.3\DKL(P||Q)$ & (c) $\DJS(P||Q) $ & (d) $\SE(P|Q)$\\[2mm]
    \includegraphics[width=42mm]{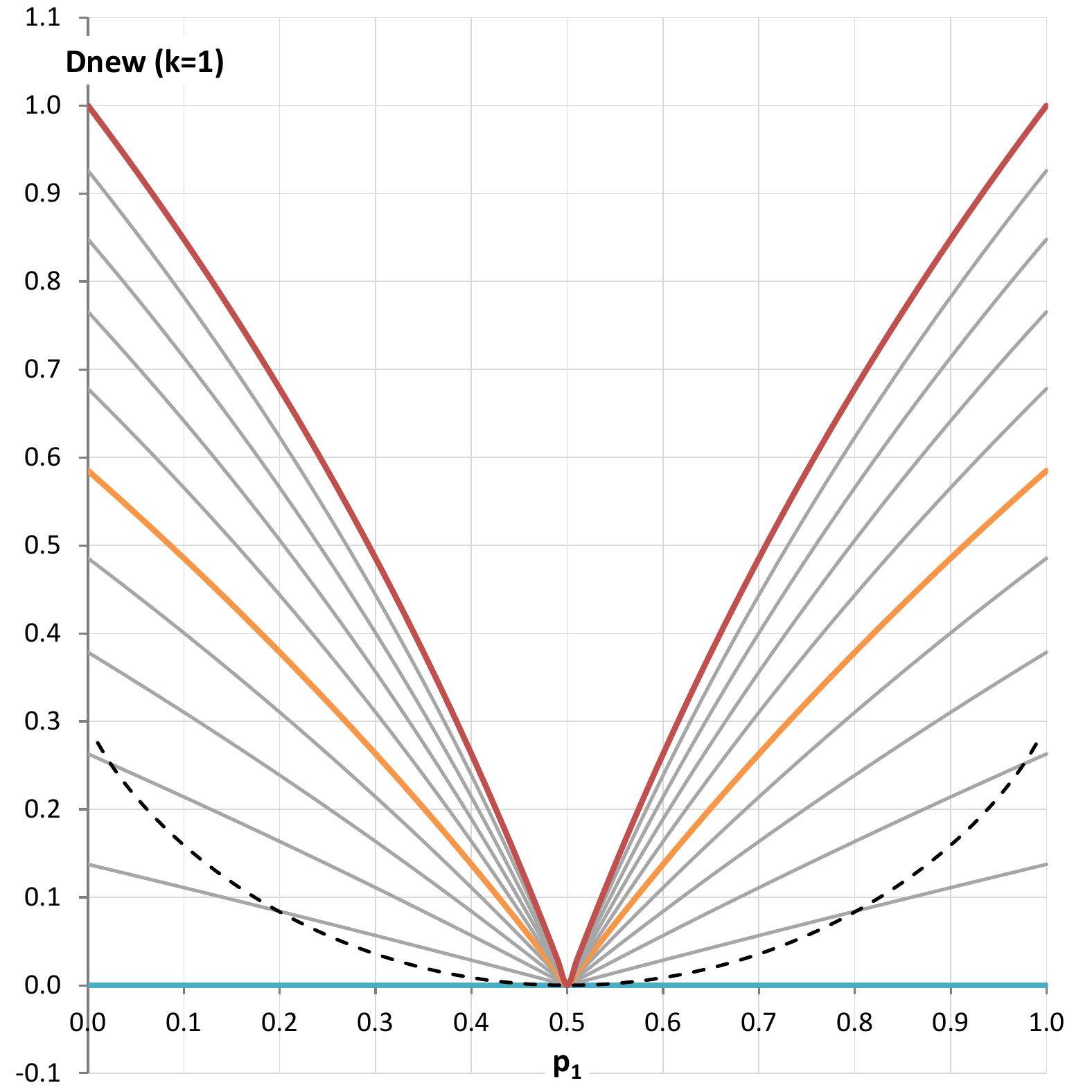} &
    \includegraphics[width=42mm]{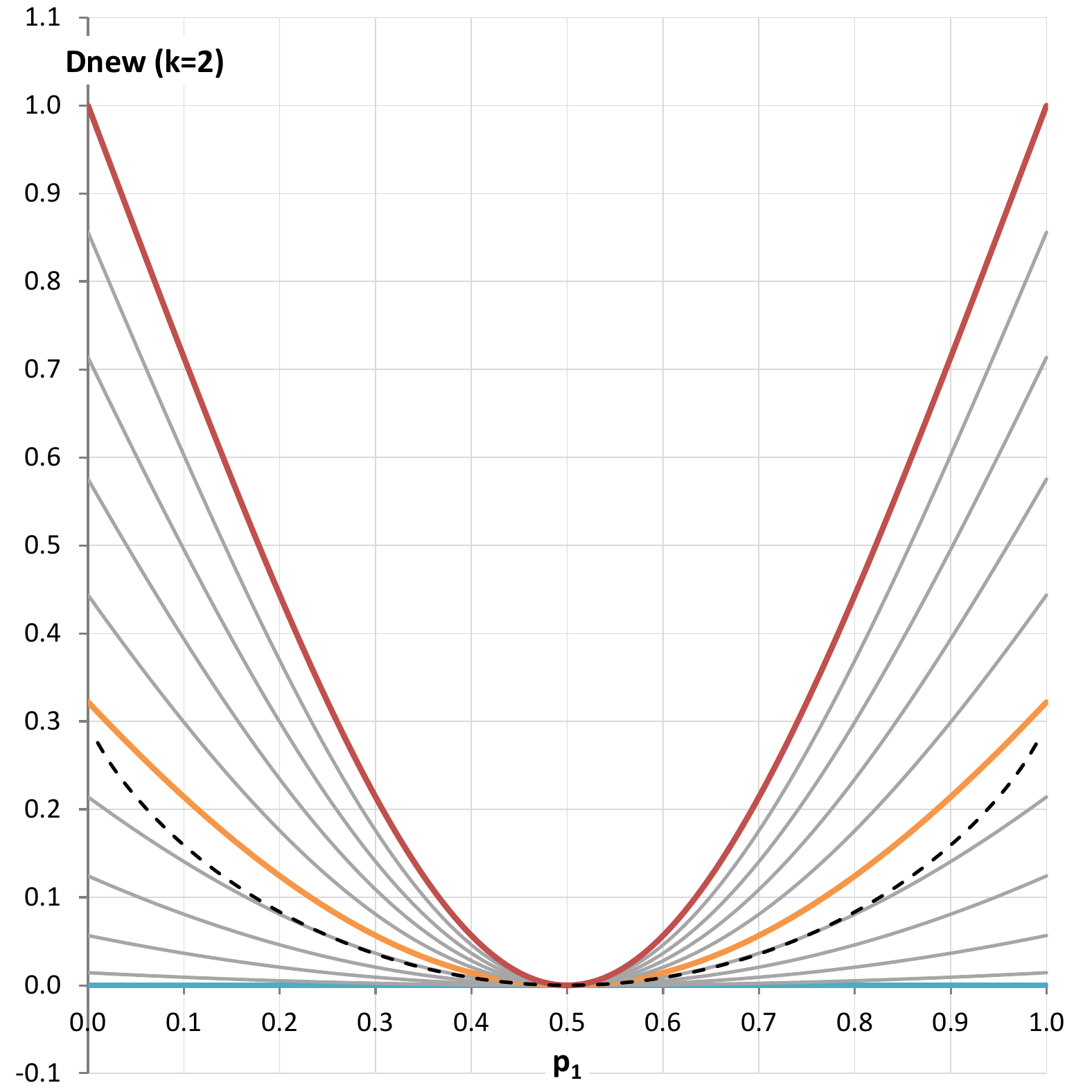} &
    \includegraphics[width=42mm]{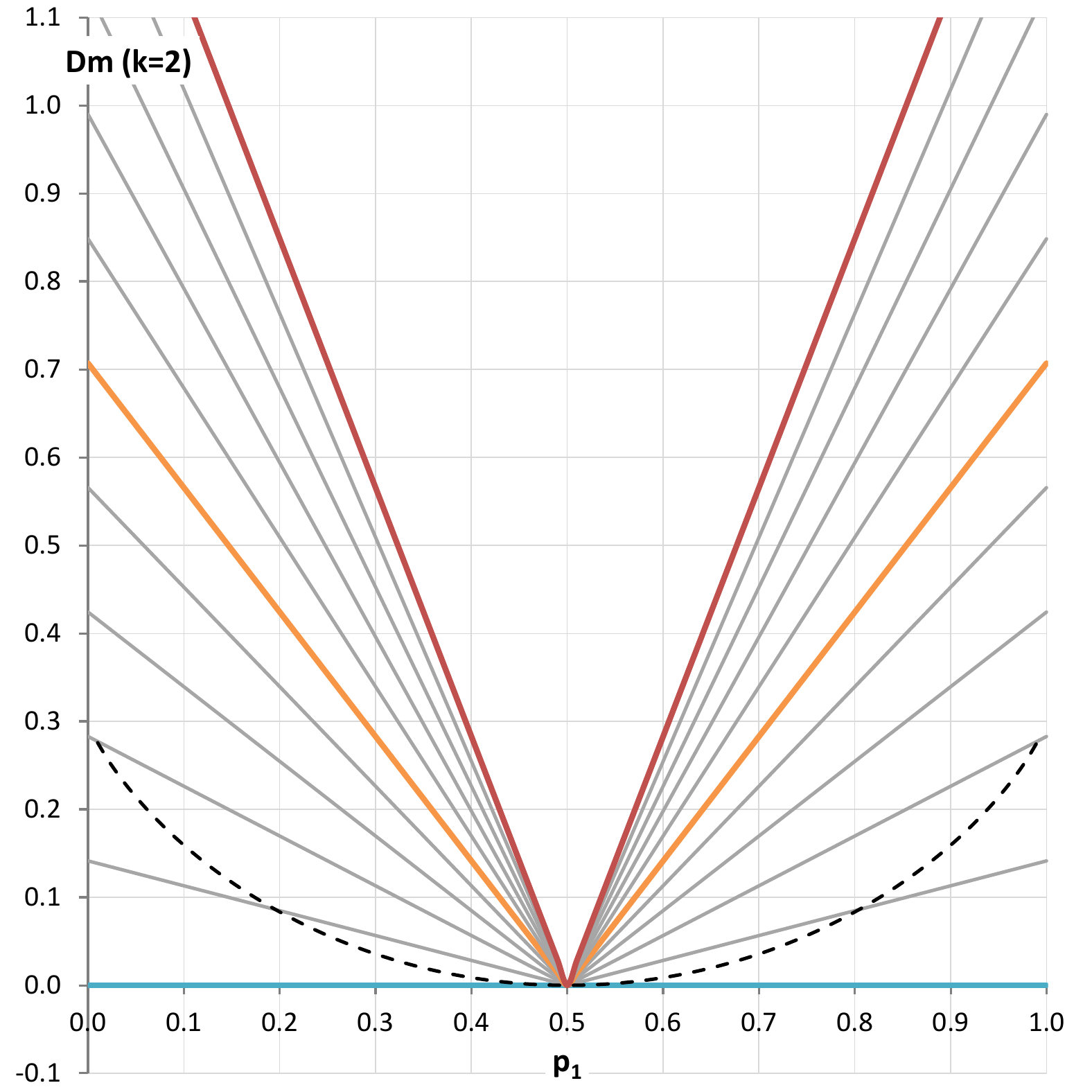} &
    \includegraphics[width=42mm]{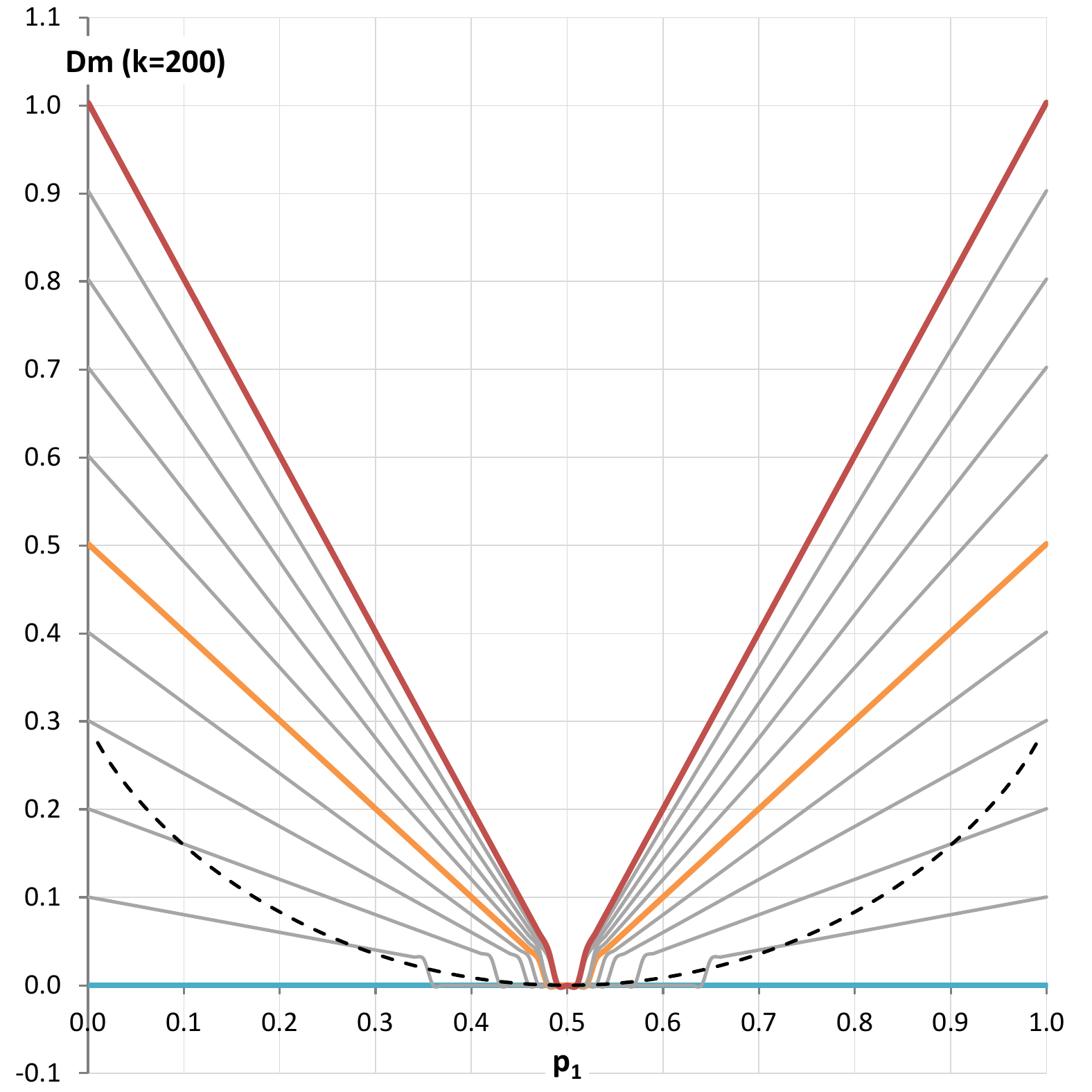} \\
    (e) $\Dnew(P||Q)$, $k=1$ & (f) $\Dnew(P||Q)$, $k=2$ &
    (g) $D^k_\text{M}(P,Q)$, $k=2$ & (h) $D^k_\text{M}(P,Q)$, $k=200$ \\[2mm]
\end{tabular}
\includegraphics[width=180mm]{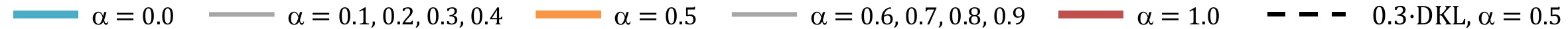}
\caption{The different measurements of the divergence of two PMFs, $P=\{ p_1, 1-p_1 \}$ and $Q=\{ q_1, 1-q_1 \}$. The $x$-axis shows $p_1$, varying from 0 to 1, while we set $q_1 = (1-\alpha)p_1 + \alpha(1-p_1), \alpha \in [0, 1]$. When $\alpha = 1$, $Q$ is most divergent away from $P$.}
\label{fig:P1_0_1}
\vspace{-4mm}
\end{figure*}

% --------------------------------------------------
\subsection{Candidates of Bounded Measures}
\label{sec:NewMetric}
While numerical approximation may provide a bounded KL-divergence, it is not easy to determine the value of $\epsilon$ and it is difficult to ensure everyone to use the same $\epsilon$ for the same alphabet or comparable alphabets.
It is therefore desirable to consider bounded measures that may be used in place of $\DKL$.

Jensen-Shannon divergence is such a measure:
\begin{equation} \label{eq:DJS}
\begin{split}
    \DJS(P||Q) &= \frac{1}{2} \bigl( \DKL(P||M) + \DKL(Q||M) \bigr) = \DJS(Q||P)\\
    &= \frac{1}{2} \sum_{i=1}^n \biggl(p_i \log_2 \frac{2 p_i}{p_i + q_i} + q_i \log_2 \frac{2 q_i}{p_i + q_i} \biggr)
\end{split}
\end{equation}
\noindent where $P$ and $Q$ are two PMFs associated with the same alphabet $\mathbb{Z}$ and $M$ is the average distribution of $P$ and $Q$.
Each letter $z_i \in \mathbb{Z}$ is associated with a probability value $p_i \in P$ and another $q_i \in Q$. 
With the base 2 logarithm as in Eq.\,\ref{eq:DJS}, $\DJS(P||Q)$ is bounded by 0 and 1.

Another bounded measure is the conditional entropy $\SE(P|Q)$:
\begin{equation} \label{eq:CondSE}
    \SE({P|Q}) = \SE(P) - \MI(P;Q) =
    \SE(P) - \sum_{i=1}^n\sum_{j=1}^n r_{i,j} \log_2 \frac{r_{i,j}}{p_i q_j}
\end{equation}
\noindent where $\MI(P;Q)$ is the mutual information between $P$ and $Q$ and $r_{i,j}$ is the joint probability of the two conditions of $z_i, z_j \in \mathbb{Z}$ that are associated with $P$ and $Q$.
$\SE(P|Q)$ is bounded by 0 and $\SE(P)$. 

The third bounded measure was proposed as part of this work, which is denoted as $\Dnew$ and is defined as follows:
\begin{equation} \label{eq:New}
    \Dnew(P||Q) = \frac{1}{2} \sum_{i=1}^n (p_i + q_i) \log_2 \bigl( |p_i - q_i|^k + 1 \bigr)
\end{equation}
\noindent where $k>0$. Because $0 \leq |p_i - q_i|^k \leq 1$, we have
\[
    \frac{1}{2} \sum_{i=1}^n (p_i + q_i) \log_2 (0+1)
    \leq \Dnew(P||Q) \leq \frac{1}{2} \sum_{i=1}^n (p_i + q_i) \log_2 (1+1)
\]
Since $\log_2 1=0$, $\log_2 2=1$, $\sum p_i = 1$, $\sum q_i = 1$, $\Dnew(P||Q)$ is thus bounded by 0 and 1.
The formulation of $\Dnew(P||Q)$ was derived from its non-commutative version: 
\begin{equation} \label{eq:NewA}
    \Dncm(P||Q) = \sum_{i=1}^n p_i \log_2 \bigl( |p_i - q_i|^k + 1 \bigr)
\end{equation}
which captures the non-commutative property of $\DKL$.
For each letter $z_i \in \mathbb{Z}$, $\DKL$ calculates the difference between two probability values in the logarithmic domain (i.e., $\log_2 p_i - \log_2 q_i$), while $\Dncm(P||Q)$ calculates the absolute difference in its original probabilistic domain (i.e., $x=|p_i - q_i|$), and then converts the difference $x$ to the logarithmic domain (i.e., $\log_2 x$ or $\log_2 f(x)$), where $f(x)$ is a transformation of $x$.
As $\log_2 x \in (\infty, 0]$ when $x \in (0, 1]$, there are still the problems of unboundedness when $x \rightarrow 0$ and invalidity when $x=0$.
By using $f(x)=x+1$, we have $\log_2 f(x) \in [0, 1]$ when $x \in [0, 1]$, resolving both problems.
Although $\log_2 f(x)$ and $\log_2 x$ are both ascending measures, they have different gradient functions, or visually, different shapes.
We thus introduce a power parameter $k$ to enable our investigation into different shapes.
Meanwhile, deriving $\Dnew(P||Q)$ from $\Dncm(P||Q)$ is easier than deriving $\DJS$ from $\DKL$ \cite{Lin:1991:TIT}.
In this work, we focus on two options of $\Dnew$ and $\Dncm(P||Q)$, i.e., when $k=1$ and $k=2$.

As $\DJS$, $\Dnew$, and $\Dncm$ are bounded by [0, 1], if any of them is selected to replace $\DKL$, Eq.\,\ref{eq:CBM-2} can be rewritten as 
\begin{equation} \label{eq:CBM-3}
  \text{Benefit} = \SE(\mathbb{Z}_i) - \SE(\mathbb{Z}_{i+1})
                 - \SE_{\text{max}}(\mathbb{Z}_i) \mathcal{D}(\mathbb{Z}'_i||\mathbb{Z}_i)
\end{equation}
\noindent where $\SE_{\text{max}}$ denotes maximum entropy, while $\mathcal{D}$ is a placeholder for $\DJS$, $\Dnew$, or $\Dncm$.

The four measures in Eqs.\,\ref{eq:DJS}, \ref{eq:CondSE}, \ref{eq:New}, \ref{eq:NewA} all consist of logarithmic scaling of probability values, in the same form of Shannon entropy.
They are entropic measures.
In addition, we also considered a set of non-entropic measures in the form of Minkowski distances, which have the following general form:
\begin{equation}
    D^k_{\text{M}}(P,Q) = \sqrt[\leftroot{2}\uproot{10} k]{\sum_{i=1}^n |p_i - q_i|^k}
    \quad (k > 0)
\end{equation}
\noindent where we use symbol $D$ instead of $\mathcal{D}$ because it is not entropic.

% --------------------------------------------------
\subsection{Comparing Bounded Measures: Visual Analysis}
\label{sec:VisualAnalysis}
Given those bounded candidates in the previous section, we would like to select the most suitable measure to be used in Eq.\,\ref{eq:CBM-3}.
Similar to selecting many other measures (e.g., metric vs. imperial), there is no ground truth as to which is correct.
We thus devised a set of criteria and conducted multi-criteria decision analysis (MCDA) \cite{Ishizaka:2013:book}.

As shown in Table \ref{tab:MultiCriteria}, we have considered nine criteria.
Criteria 1-5 concern general mathematical properties of these measures, while criteria 6-9 are assessments based on numerical instances.
For criteria 1, 4-7, we use visualization plots to aid our analysis of the mathematical properties, while for criteria 8 and 9, we use visualization applications to provide instances of using divergence measures.
We detail our analysis of criteria 1-7 in this section, and that of criteria 8 and 9 in Section \ref{sec:CaseStudies}.
Based on our analysis, we score each divergence measure against a criterion using ordinal values between 0 and 5 (0 unacceptable, 1 fall-short, 2 inadequate, 3 mediocre, 4 good, 5 best).
We draw our conclusion about the multi-criteria in Section \ref{sec:Conclusions}.

\textbf{Criterion 1.}
This is essential since the selected divergence measure is to be bounded.
Otherwise we could just use the KL-divergence.
Let us consider a simple alphabet $\mathbb{Z} = \{z_1, z_2 \}$, which is associated with two PMFs, $P=\{ p_1, 1-p_1 \}$ and $Q=\{ q_1, 1-q_1 \}$.
We set $q_1 = (1-\alpha)p_1 + \alpha(1-p_1), \alpha \in [0, 1]$, such that when $\alpha = 1$, $Q$ is most divergent away from $P$.
The entropy values of $P$ and $Q$ fall into the range of [0, 1].
Hence semantically, it is more intuitive to reason an unsigned value representing their divergence within the same range.

Fig. \ref{fig:P1_0_1} shows several measures by varying the values of $p_1$ in the range of $[0, 1]$.
We can obverse that $\DKL$ raises its values quickly above 1 when $\alpha = 1, p_1 \leq 0.22$.
Its scaled version, $0.3\DKL$, does not rise up as quick as $\DKL$ but raises above 1 when $\alpha = 1, p_1 \leq 0.18$.
In fact $\DKL$ and $0.3\DKL$ are not only unbounded, they do not return valid values  when $p_1 = 0$ or $p_1 = 1$.
We therefore score them 0 for Criterion 1.

$\DJS$, $\SE(P|Q)$, $\Dnew$, and $\Dncm$ are all bounded by [0, 1], and
semantically intuitive.
We score them 5.
Although $\DM$ is a bounded measure, its semantic interpretation is not ideal, because its upper bound depends on $k$ and is always $> 1$.
We thus score it 3.
Although $0.3\DKL$ is eliminated based on criterion 1, it is kept in Table \ref{tab:MultiCriteria} as a benchmark in analyzing criteria 2-5.
Meanwhile, we carry all other scores forward to the next stage of analysis.

% We can visualize how different measures numerically convey the divergence between $P$ and $Q$ by observing their relationship with $0.3\DKL$.

\begin{table*}[t]
  \centering
  \caption{A summary of multi-criteria analysis. Each measure is scored against a criterion using an integer in [0, 5] with 5 being the best.}
  \label{tab:MultiCriteria}
  \begin{tabular}{@{}l@{\hspace{3mm}}c@{\hspace{3mm}}c@{\hspace{3mm}}c@{\hspace{2mm}}c@{\hspace{2mm}}c@{\hspace{3mm}}c@{\hspace{3mm}}c@{\hspace{3mm}}c@{\hspace{3mm}}c@{\hspace{3mm}}c@{}}
  \textbf{Criteria} & \textbf{Importance}
  & $0.3\DKL$ & $\DJS$ & $\SE(P|Q)$
  & $\DnewA$ & $\DnewB$
  & $\DncmA$ & $\DncmB$
  & $D^{k=2}_{\text{M}}$ & $D^{k=200}_{\text{M}}$\\[0.5mm]
  \hline
  1. Boundedness & critical
        & 0 & 5 & 5 & 5 & 5 & 5 & 5 & 3 & 3 \\
        \multicolumn{11}{l}{$\blacktriangleright$
        \emph{$0.3\DKL$ is eliminated but used below only for comparison.
        The other scores are carried forward.}} \\
  \hline
  2. Number of PMFs & important
        & \textcolor{gray}{5} & 5 & 2 & 5 & 5 & 5 & 5 & 5 & 5\\ 
  3. Entropic measures & important
        & \textcolor{gray}{5} & 5 & 5 & 5 & 5 & 5 & 5 & 1 & 1 \\
  4. Curve shapes (Fig. \ref{fig:P1_0_1}) & helpful
        & \textcolor{gray}{5} & 5 & 1 & 2 & 4 & 2 & 4 & 3 & 3 \\
  5. Curve shapes (Fig. \ref{fig:NearZero}) & helpful
        & \textcolor{gray}{5} & 4 & 1 & 3 & 5 & 3 & 5 & 2 & 3\\[0.5mm]
  \multicolumn{2}{l}{$\blacktriangleright$
        \emph{Eliminate} \emph{$\SE(P|Q)$}, $\mathcal{D}^2_{\text{M}}$, $\mathcal{D}^{200}_{\text{M}}$ \emph{based on criteria 1-5}}
        & \textbf{sum:} & \textbf{24} & \textcolor{gray}{\textbf{14}}
        & \textbf{20} & \textbf{24} & \textbf{20} & \textbf{24}
        & \textcolor{gray}{\textbf{14}} & \textcolor{gray}{\textbf{15}}\\[0.5mm]
  \hline
  6. Scenario: \emph{good} and \emph{bad} (Fig. \ref{fig:Compare-GoodBad}) & helpful
        & $-$ & 3 & $-$ & 5 & 4 & 5 & 4 & $-$ & $-$\\
  7. Scenario: A, B, C, D (Fig. \ref{fig:Compare-ABCD}) & helpful
        & $-$ & 4 & $-$ & 5 & 3 & 2 & 1 & $-$ & $-$\\
  8. Case Study 1 (Section \ref{sec:VolVis}) & important
        & $-$ & 5 & $-$ & 1 & 5 & 5 & 5 & $-$ & $-$\\
  9. Case Study 2: (Section \ref{sec:London}) & important
        & $-$ & 3 & $-$ & 1 & 5 & 3 & 3 & $-$ & $-$\\[0.5mm]
        \multicolumn{2}{l}{$\blacktriangleright$
        \emph{$\DnewB$ has the highest score based on criteria 6-9 (1-9)}}
        & \textbf{sum:} & \textcolor{gray}{\textbf{15}(39)} &
        & \textcolor{gray}{\textbf{12}(32)} & \textbf{17}(41)
        & \textcolor{gray}{\textbf{15}(35)} & \textcolor{gray}{\textbf{13}(37)}\\[0.5mm]
   \hline
  \end{tabular}
  \vspace{-0mm}
\end{table*}

\textbf{Criterion 2.}
For criteria 2-5, we follow the base-criterion method \cite{Haseli:2020:IJMSEM} by considering 
$\DKL$ and $0.3\DKL$ as the benchmark.
Criterion 2 concerns the number of PMFs as the input variables of each measure.
$\DKL$ and $0.3\DKL$ depend on two PMFs, $P$ and $Q$.
All candidates of the bounded measures depend on two PMFs, except the conditional entropy $\SE(P|Q)$ that depends on three.
Because it requires some effort to obtain a PMF, especially a joint probability distribution, this makes $\SE(P|Q)$ less favourable and it is scored 2.

\textbf{Criterion 3.}
In addition, we prefer to have an entropic measure as it is more compatible with the measure of alphabet compression as well as $\DKL$ that is to be replaced.
For this reason, $\DM$ is scored 1.

\textbf{Criterion 4.}
One may wish for a bounded measure to have a geometric behaviour similar to $\DKL$ since it is the most popular divergence measure.
Since $\DKL$ rises up far too quickly as shown in Fig. \ref{fig:P1_0_1}, we use $0.3\DKL$ as a benchmark, though it is still unbounded.
As Fig. \ref{fig:P1_0_1} plots the curves for $\alpha = 0.0, 0.1, \ldots, 1.0$, we can visualize the ``geometric shape'' of each bounded measure, and compare it with that of $0.3\DKL$.  

From Fig. \ref{fig:P1_0_1}, we can observe that $\DJS$ has almost a perfect match when $\alpha = 0.5$, while $\Dnew (k=2)$ is also fairly close.
They thus score 5 and 4 respectively in Table \ref{tab:MultiCriteria}.
Meanwhile, the lines of $\SE(P|Q)$ curve in the opposite direction of $0.3\DKL$.
We score it 1.
$\Dnew (k=1)$ and $\DM (k=2, k=200)$ are of similar shapes, with $\DM$ correlating with $0.3\DKL$ slightly better.
We thus score $\Dnew (k=1)$ 2 and $\DM (k=2, k=200)$ 3.
For the PMFs $P$ and $Q$ concerned, $\Dncm$ has the same curves as $\Dnew$.
Hence $\Dncm$ has the same score as $\Dnew$ in Table \ref{tab:MultiCriteria}.

\begin{figure}[h!]
  \centering
  \includegraphics[width=80mm]{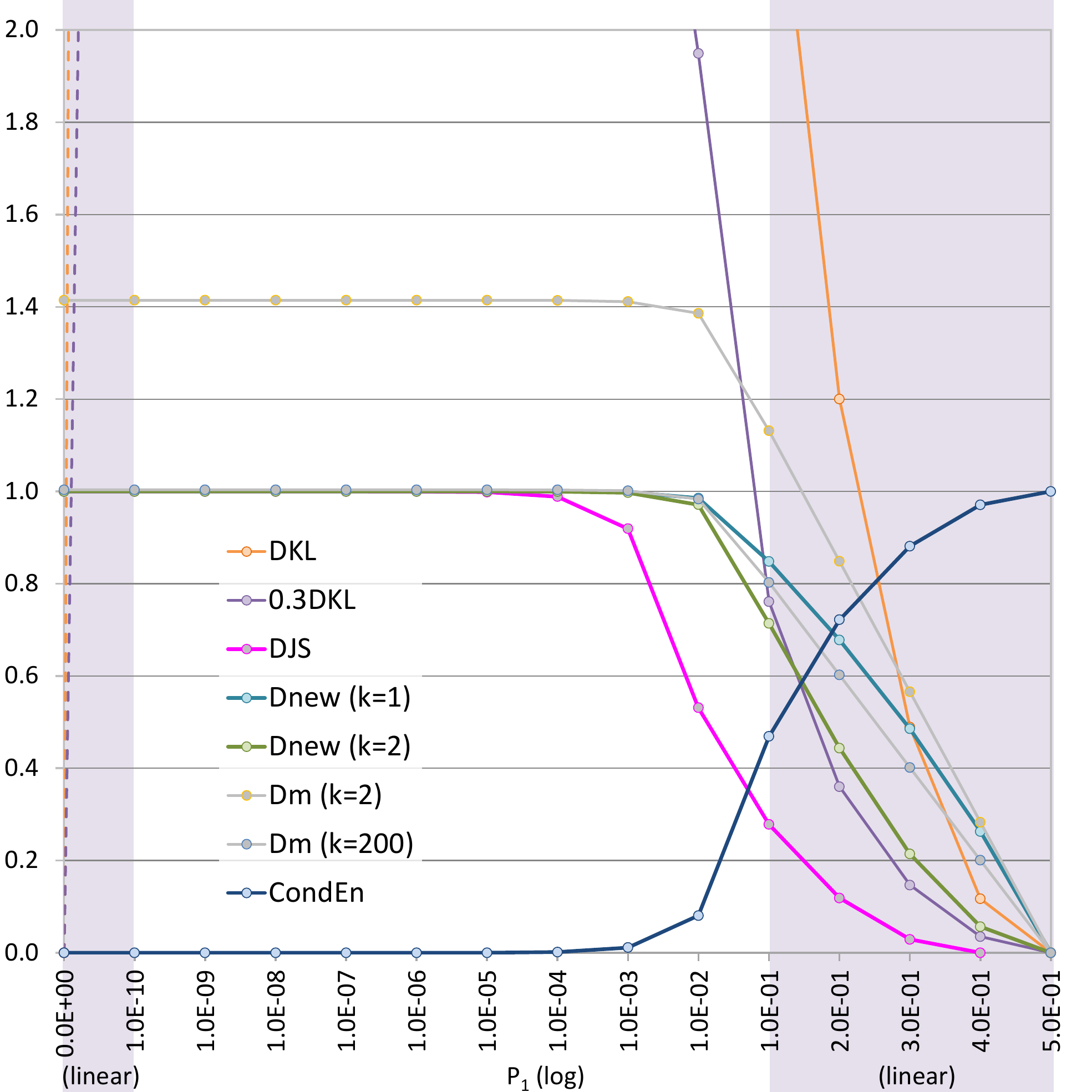}
  \caption{A visual comparison of the candidate measures in a range near zero. Similar to Fig. \ref{fig:P1_0_1}, $P=\{ p_1, 1-p_1 \}$ and $Q=\{ q_1, 1-q_1 \}$, but only the curve $\alpha=1$ is shown, i.e., $q_1 = 1-p_1$. The line segments of $\DKL$ and $0.3\DKL$ in the range $[0, 0.1^{10}]$ do not represent the actual curves. The ranges $[0, 0.1^{10}]$ and $[0.1, 0.5]$ are only for references to the nearby contexts as they do not use the same logarithmic scale as in $[0.1^{10}, 0.1]$.}%
  \label{fig:NearZero}
  \vspace{-4mm}
\end{figure}

\begin{figure*}[t]
    \centering
    \includegraphics[width=180mm]{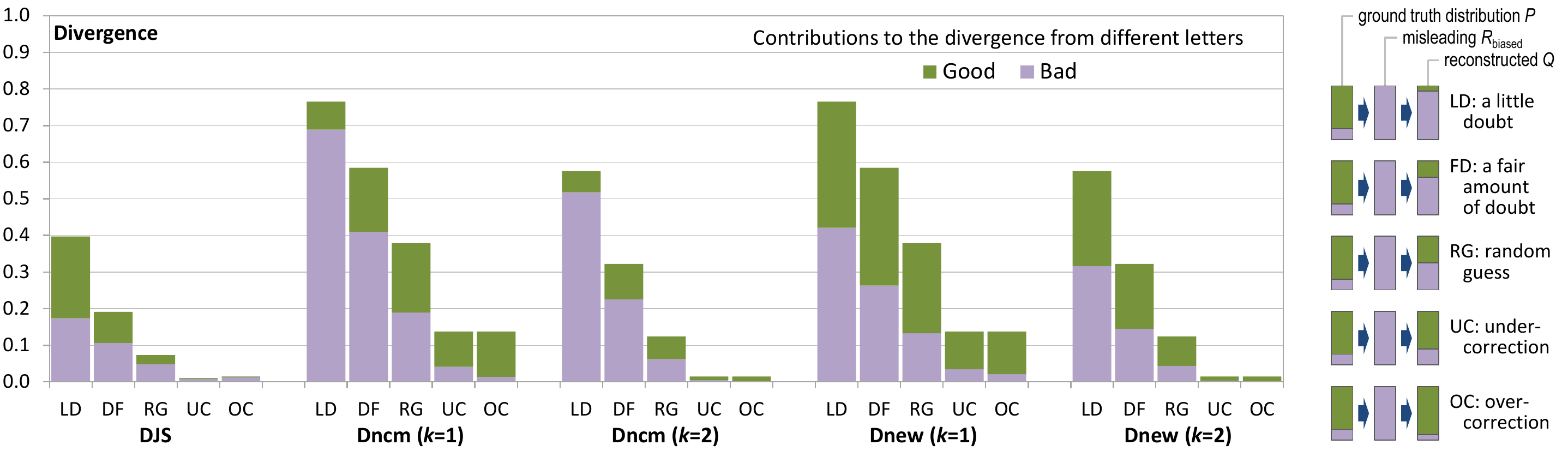}
    \caption{An example scenario with two states \emph{good} and \emph{bad} has a ground truth PMF $P=\{0.8, 0.2\}$. From the output of a biased process that always informs users that the situation is \emph{bad}. Five users, LD, DF, RG, UC and OC, have different knowledge, and thus different divergence. The five candidate measures return different values of divergence. We would like to see which set of values are more intuitive.}
    \label{fig:Compare-GoodBad}
    \vspace{-4mm}
\end{figure*}

\textbf{Criterion 5.} We now consider Fig. \ref{fig:NearZero}, where the candidate measures are visualized in comparison with $\DKL$ and $0.3\DKL$ in a range close to zero, i.e., $[0.1^{10}, 0.1]$.
The ranges $[0, 0.1^{10}]$ and $[0.1, 0.5]$ are there only for references to the nearby contexts as they do not have the same logarithmic scale as that in the range
$[0.1^{10}, 0.1]$.
We can observe that in $[0.1^{10}, 0.1]$ the curve of $0.3\DKL$ rises
as almost quickly as $\DKL$.
This confirms that simply scaling the KL-divergence is not an adequate solution.
The curves of $\DnewA$ and $\DnewB$ converge to their maximum value 1.0 earlier than that of $\DJS$.
If the curve of $0.3\DKL$ is used as a benchmark as in Fig. \ref{fig:P1_0_1}, the curve of $\DnewB$ is closer to $0.3\DKL$ than that of $\DJS$.
We thus score $\DnewB$: 5, $\DJS$: 4, $\DnewA$: 3, $\DM (k=200)$: 3, $\DM (k=200)$: 2, and $\SE(P|Q)$: 1.
Same as Fig. \ref{fig:P1_0_1}, $\Dncm$ has the same curves and thus the same score as $\Dnew$.

\begin{figure*}[t]
    \centering
    \includegraphics[width=180mm]{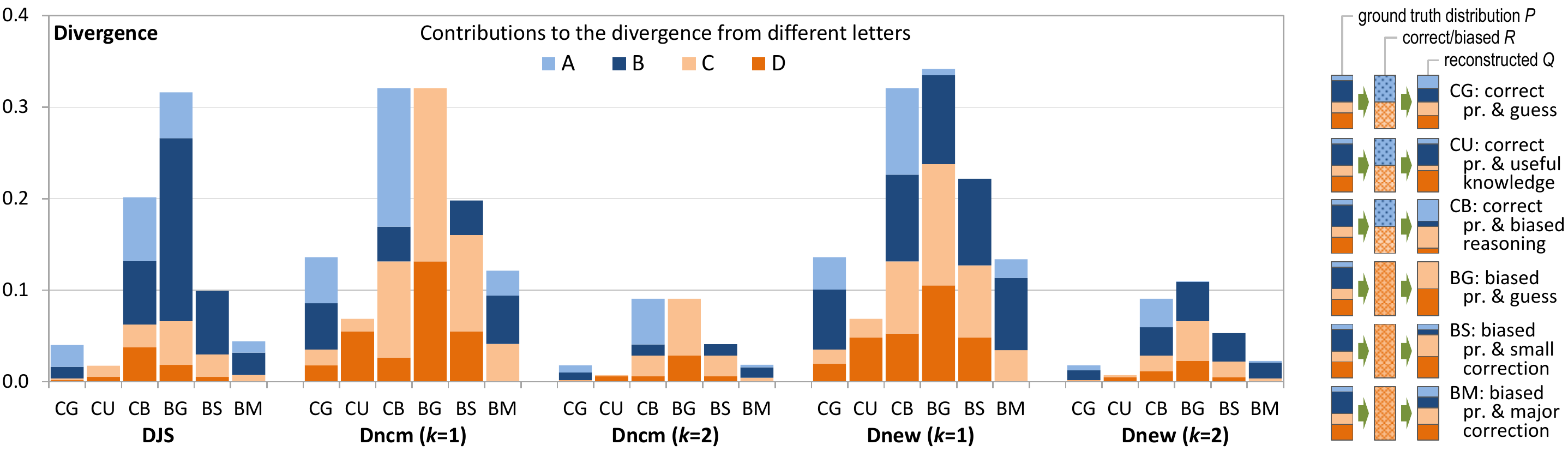}
    \caption{An example scenario with four data values: A, B, C, are D. Two processes (one correct and one biased) aggregated them to two values AB and CD. Users CG, CU, CB attempt to reconstruct [A, B, C, D] from the output [AB, CD] of the correct process, while BG, BS, and BM attempt to do so with the output from the biased processes.
    Five candidate measures compute values of divergence of the six users.}
    \label{fig:Compare-ABCD}
    \vspace{-4mm}
\end{figure*}

The sums of the scores for criteria 1-5 indicate that $\SE(P|Q)$ and $\DM$ are much less favourable than $\DJS$, $\Dnew$, and $\Dncm$.
Because these criteria have more holistic significance than the instance-based analysis for criteria 6-9, we can eliminate $\SE(P|Q)$ and $\DM$ for further consideration.
Ordinal scores in MCDA are typically subjective.
Nevertheless, in our analysis, $\pm 1$ in those scores would not affect the elimination. 

\textbf{Criterion 6.}
For criteria 6-9, we followed the best-worst method of MCDA \cite{Rezaei:2015:O}.
Let us consider a few numerical examples that may represent some practical scenarios.
We use these scenarios to see if the values returned by different divergence measures make sense.
Let $\mathbb{Z}$ be an alphabet with two letters, \emph{good} and \emph{bad}, for describing a scenario (e.g., an object or an event), which has the probability of \emph{good} is $p_1 = 0.8$, and that of \emph{bad} is $p_2 = 0.2$.
In other words, $P = \{0.8, 0.2\}$.
Imagine that a biased process (e.g., a distorted visualization, an incorrect algorithm, or a misleading communication) conveys the information about the scenario always \emph{bad}, i.e., a PMF $R_{\text{biased}} = \{0, 1\}$.
Users at the receiving end of the process may have different knowledge about the actual scenario, and they will make a decision after receiving the output of the process.
For example, there are five users and we have obtained the probability of their decisions as follows:
\begin{itemize}
  \vspace{-2mm}
  \item LD --- The user has a little doubt about the output of the process, and decides \emph{bad} 90\% of the time, and \emph{good} 10\% of the time, i.e., with PMF $Q = \{0.1, 0.9\}$.
  \vspace{-2mm}
  \item FD --- The user has a fair amount of doubt, with $Q = \{0.3, 0.7\}$.
  \vspace{-2mm}
  \item RG --- The user makes a random guess, with $Q = \{0.5, 0.5\}$.
  \vspace{-2mm}
  \item UC --- The user has adequate knowledge about $P$, but under-compensate it slightly, with $Q = \{0.7, 0.3\}$.
  \vspace{-2mm}
  \item OC --- The user has adequate knowledge about $P$, but over-compensate it slightly, with $Q = \{0.9, 0.1\}$.
\end{itemize}
We can use different candidate measures to compute the divergence between $P$ and $Q$.
Fig. \ref{fig:Compare-GoodBad} shows different divergence values returned by these measures, while the transformations from $P$ to $R_\text{biased}$ and then to $Q$ are illustrated on the right margin of the figure.
Each value is decomposed into two parts, one for \emph{good} and one for \emph{bad}. 
All these measures can order these five users reasonably well.
The users UC (under-compensate) and OC (over-compensate) have the same values with $\Dnew$ and $\Dncm$, while $\DJS$ considers OC has slightly more divergence than UC (0.014 vs. 0.010).
$\DJS$ returns relatively low values than other measures.
For UC and OC, $\DJS$, $\DncmA$, and $\DnewB$ return small values $(<0.02)$, which are a bit difficult to estimate.   

$\DncmA$ and $\DncmB$ show strong asymmetric patterns between \emph{good} and \emph{bad}, reflecting the probability values in $Q$.
In other words, the more decisions on \emph{good}, the more \emph{good}-related divergence.
This asymmetric pattern is not in anyway incorrect, as the KL-divergence is also non-commutative and would also produce much stronger asymmetric patterns.
An argument for supporting commutative measures would point out that the higher probability of \emph{good} in $P$ should also influence the balance between the \emph{good}-related divergence.

We decide to score $\DJS$ 3 because of its lower valuation and its non-equal comparison of OU and OC.
We score $\DncmA)$ and $\DnewA$ 5; and $\DncmB$ and $\DnewB$ 4 as the values returned by $\DncmA$ and $\DnewA$ are slightly more intuitive.

\textbf{Criterion 7.} We now consider a slightly more complicated scenario with four pieces of data, A, B, C, and D, which can be defined as an alphabet $\mathbb{Z}$ with four letters.
The ground truth PMF is $P=\{0.1, 0.4, 0.2, 0.3\}$.
Consider two processes that combine these into two classes AB and CD.
These typify clustering algorithms, downsampling processes, discretization in visual mapping, and so on.
One process is considered to be \emph{correct}, which has a PMF for AB and CD as $R_\text{correct}=\{0.5, 0.5\}$, and another \emph{biased} process with $R_{\text{biased}}=\{0, 1\}$.
Let CG, CU, and CH be three users at the receiving end of the \emph{correct} process, and BG, BS, and BM be three other users at the receiving end of the \emph{biased} process.
The users with different knowledge exhibit different abilities to reconstruct the original scenario featuring A, B, C, D from aggregated information about AB and CD.
Similar to the \emph{good}-\emph{bad} scenario, such abilities can be captured by a PMF $Q$. For example, we have:
\begin{itemize}
  \vspace{-2mm}
  \item CG makes random guess, $Q=\{0.25, 0.25, 0.25, 0.25\}$.
  \vspace{-2mm}
  \item CU has useful knowledge, $Q=\{0.1, 0.4, 0.1, 0.4\}$.
  \vspace{-2mm}
  \item CB is highly biased, $Q=\{0.4, 0.1, 0.4, 0.1\}$.
  \vspace{-2mm}
  \item BG makes guess based on $R_{\text{biased}}$, $Q=\{0.0, 0.0, 0.5, 0.5\}$.
  \vspace{-2mm}
  \item BS makes a small adjustment, $Q=\{0.1, 0.1, 0.4, 0.4\}$.
  \vspace{-2mm}
  \item BM makes a major adjustment, $Q=\{0.2, 0.2, 0.3, 0.3\}$.
\end{itemize}
Fig. \ref{fig:Compare-ABCD} compares the divergence values returned by the candidate measures for these six users, while the transformations from $P$ to $R_\text{correct}$ or $R_\text{biased}$, and then to $Q$ are illustrated on the right.
We can observe that $\Dncm$ and $\DnewB$ return values $<0.1$, which seem to be less intuitive. Meanwhile $\DJS$ shows a large portion of divergence from the AB category, while $\DncmA$ and $\DncmB$ show more divergence in the BC category.
In particular, for user BG, $\DncmA$ and $\DncmB$ do not show any divergence in relation to A and B, though BG clearly has reasoned A and B rather incorrectly.
$\DnewA$ and $\DnewB$ show a relatively balanced account of divergence associated with A, B, C, and D.
On balance, we give scores 5, 4, 3, 2, 1 to $\DnewA$, $\DJS$, $\DnewB$, $\DncmA$, and $\DncmB$ respectively.

% With the major shortcomings of $\Dncm (k=1, k=2)$ in this scenario, we can now focus on three commutative measures $\DJS$ and $\Dnew (k=1, k=2)$ in conjunction with two case studies.

% A major disadvantage of Eq.\,\ref{eq:CondSE} is that it requires an additional probability distribution $R$.
% In the context of this work, one would have to estimate three distributions instead of two.

% ====================
\section{Comparing Bounded Measures: Case Studies}
\label{sec:CaseStudies}
To complement the visual analysis in Section \ref{sec:VisualAnalysis}, we conducted two surveys to collect some realistic examples that feature the use of knowledge in visualization.
In addition to providing instances of criteria 8 and 9 for selecting a bounded measure, the surveys were also designed to demonstrate that one could use a few simple questions to estimate the cost-benefit of visualization in relation to individual users.
% Built on the visual analysis in the previous section, we focus on three divergence measures, namely the JS divergence $\DJS$ and two versions of the new divergence, i.e., $\Dnew$ with $k=1$ and $k=2$.
% We denote $\Dnew (k=1)$ as $\mathcal{D}_1$, and $\Dnew (k=2)$ as $\mathcal{D}_2$.

% --------------------------------------------------
\subsection{Volume Visualization (Criterion 8)}
\label{sec:VolVis}
This survey, which involved ten surveyees, was designed to collect some real-world data that reflects the use of knowledge in viewing volume visualization images.
The full set of questions were presented to surveyees in the form of slides, which are included in the supplementary materials.
The full set of survey results is given in Appendix C.
The featured volume datasets were from ``The Volume Library'' \cite{Roettger:2019:web}, and visualization images were either rendered by the authors or from one of the four publications \cite{Nagy:2002:VMV,Correa:2006:TVCG,Wu:2007:TVCG,Jung:2008:web}.

The transformation from a volumetric dataset to a volume-rendered image typically features a noticeable amount of alphabet compression.
Some major algorithmic functions in volume visualization, e.g., iso-surfacing, transfer function, and rendering integral, all facilitate alphabet compression, hence information loss.

In terms of rendering integral, maximum intensity projection (MIP) incurs a huge amount of information loss in comparison with the commonly-used emission-and-absorption integral \cite{Max:2010:book}.
As shown in Fig. \ref{fig:Arteries}, the surface of arteries are depicted more or less in the same color.
The accompanying question intends to tease out two pieces of knowledge, ``curved surface'' and ``with wrinkles and bumps''.
Among the ten surveyees, one selected the correct answer B, eight selected the relatively plausible answer A, and one selected the doubtful answer D.

Let alphabet $\mathbb{Z}=\{\mathrm{A, B, C, D}\}$ contain the four optional answers.
One may assume a ground truth PMF $Q=\{0.1, 0.878, 0.002, 0.02\}$ since there might still be a small probability for a section of artery to be flat or smooth.
The rendered image depicts a misleading impression, implying that answer C is correct or a false PMF $F=\{0, 0, 1, 0\}$.
The amount of alphabet compression is thus $\mathcal{H}(Q) - \mathcal{H}(F) = 0.225$.

When a surveyee gives an answer to the question, it can also be considered as a PMF $P$.
With PMF $Q=\{0.1, 0.878, 0.002, 0.02\}$, different answers thus lead to different values of divergence as follows:

\begin{table}[h!]
  \vspace{-2mm}
  \centering
  \begin{tabular}{@{}l@{\hspace{5mm}}r@{\hspace{3mm}}r@{\hspace{3mm}}r@{\hspace{3mm}}r@{\hspace{3mm}}r@{}}
    \textbf{Divergence for:} & $\DJS$\; & $\DnewA$\; & $\DnewB$\; & $\DncmA$\; & $\DncmB$\;\\[1mm]
    A $(P_a = \{1, 0, 0, 0\}, Q)$: & 0.758 & 0.9087 & 0.833 & 0.926 & 0.856\\
    B $(P_b = \{0, 1, 0, 0\}, Q)$: & 0.064 & 0.1631 & 0.021 & 0.166 & 0.021\\
    C $(P_c = \{0, 0, 1, 0\}, Q)$: & 0.990 & 0.9066 & 0.985 & 0.999 & 0.997\\
    D $(P_d = \{0, 0, 0, 1\}, Q)$: & 0.929 & 0.9086 & 0.858 & 0.986 & 0.971
  \end{tabular}
  \vspace{-2mm}
%  \caption{Caption}
%  \label{tab:my_label}
\end{table}

Without any knowledge, a surveyee would select answer C, leading to the highest value of divergence in terms of any of the three measures.
Based PMF $Q$, we expect to have divergence values in the order of C $>$ D $>$ A $\gg$ B.
$\DJS$, $\DnewB$, $\DncmA$, and $\DncmB$ have produced values in that order, while $\DnewA$ indicates an order of A $>$ D $>$ C $\gg$ B.
This order cannot be interpreted easily, indicating a weakness of $\DnewA$.
% We thus score $\Dnew (k=1)$ 1 in Table \ref{tab:MultiCriteria}.

Together with the alphabet compression $\mathcal{H}(Q) - \mathcal{H}(F) = 0.225$ and the $\mathcal{H}_\text{max}$ of 2 bits, we can also calculate the informative benefit using Eq.\,\ref{eq:CBM-3}.
For surveyees with different answers, the lossy depiction of the surface of arteries brought about different amounts of benefit:

\begin{table}[h!]
  \vspace{-2mm}
  \centering
  \begin{tabular}{@{}l@{\hspace{3mm}}r@{\hspace{1.5mm}}r@{\hspace{1.5mm}}r@{\hspace{1.5mm}}r@{\hspace{1.5mm}}r@{}}
    \textbf{Benefit for:} & $\DJS$\; & $\DnewA$\; & $\DnewB$\; & $\DncmA$\; & $\DncmB$\;\\[1mm]
    A $(P_a = \{1, 0, 0, 0\}, Q)$:   & $-0.889$ & $-1.190$ & $-1.038$ & $-1.224$ & $-1.084$\\
    B $(P_b = \{0, 1, 0, 0\}, Q)$: &  $0.500$ &  $0.302$ &  $0.586$ &  $0.296$ &  $0.585$\\
    C $(P_c = \{0, 0, 1, 0\}, Q)$: & $-1.351$ & $-1.185$ & $-1.097$ & $-1.369$ & $-1.366$\\
    D $(P_d = \{0, 0, 0, 1\}, Q)$:  & $-1.230$ & $-1.189$ & $-1.088$ & $-1.343$ & $-1.314$
  \end{tabular}
  \vspace{-2mm}
%  \caption{Caption}
%  \label{tab:my_label}
\end{table}

All five sets of values indicate that only those surveyees who gave answer C would benefit from such lossy depiction produced by MIP, signifying the importance of user knowledge in visualization. However, the values returned for A, C, and D by $\DnewA$ are almost indistinguishable and in an undesirable order.

One may also consider the scenarios where flat or smooth surfaces are more probable.
For example, if the ground truth PMF were $Q'=\{0.30, 0.57, 0.03, 0.10\}$ and $\mathcal{H}(Q') = 1.467$, the amounts of benefit would become:

\begin{table}[h!]
  \vspace{-2mm}
  \centering
  \begin{tabular}{@{}l@{\hspace{3mm}}r@{\hspace{1.5mm}}r@{\hspace{1.5mm}}r@{\hspace{1.5mm}}r@{\hspace{1.5mm}}r@{}}
    \textbf{Benefit for:} & $\DJS$\; & $\DnewA$\; & $\DnewB$\; & $\DncmA$\; & $\DncmB$\;\\[1mm]
    A $(P_a = \{1, 0, 0, 0\}, Q')$: &  $0.480$ &  $0.086$ &  $0.487$ & $-0.064$ &  $0.317$\\
    B $(P_b = \{0, 1, 0, 0\}, Q')$: &  $0.951$ &  $0.529$ &  $1.044$ &  $0.435$ &  $0.978$\\
    C $(P_c = \{0, 0, 1, 0\}, Q')$: & $-0.337$ & $-0.038$ &  $0.212$ & $-0.489$ & $-0.446$\\
    D $(P_d = \{0, 0, 0, 1\}, Q')$: & $-0.049$ & $-0.037$ &  $0.257$ & $-0.385$ & $-0.245$
  \end{tabular}
  \vspace{-2mm}
%  \caption{Caption}
%  \label{tab:my_label}
\end{table}

\noindent Because the ground truth PMF $Q'$ would be less certain, the knowledge of ``curved surface'' and ``with wrinkles and bumps'' would become more useful.
Further, because the probability of flat and smooth surfaces would have increased, an answer C would not be as bad as when it is with the original PMF $Q$.
Among the five measures, $\DJS$, $\DnewB$, $\DncmA$, and $\DncmB$ returned values indicating the same order of benefit, i.e., B $>$ A $>$ D $>$ C, which is consistent with PMF $Q'$.
Only $\DnewA$ orders C and D differently.

We can also observe that these measures occupy different ranges of real values.
$\DnewB$ appears to be more generous in valuing the benefit of visualization, while $\DncmA$ is less generous. We will examine this phenomenon with a more compelling example in Section \ref{sec:London}. 

% The above example of MIP rendering shows that to those users with the appropriate knowledge, the missing information in a visualization image is not really ``lost''.
% Using the categorization of visual multiplexing \cite{Chen:2014:CGF}, the information about ``curved surface'' and ``with wrinkles and bumps'' is conveyed using a \emph{hollow visual channel}.
% Volume visualization features some other forms of visual multiplexing.
% The viewers' ability to de-multiplex depends on their knowledge, which can now be estimated quantitatively.

\begin{figure}[t]
\centering
\includegraphics[width=\linewidth]{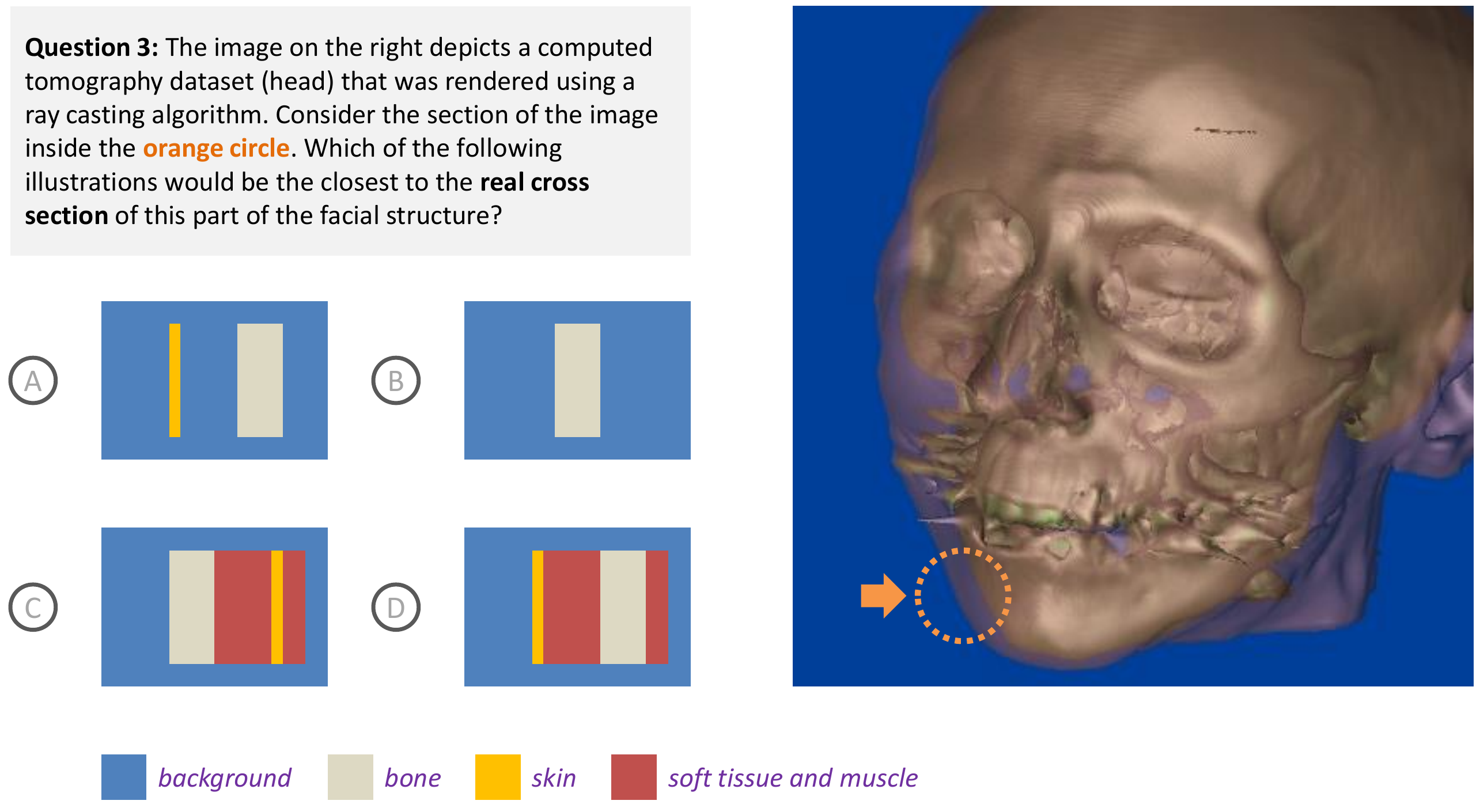}
\caption{Two iso-surfaces of a volume dataset were rendered using the ray casting method. A question about the tissue configuration in the orange circle can tease out a viewer's knowledge about the translucent depiction and the missing information.}
\label{fig:Isosurfaces}
\vspace{-4mm}
\end{figure}

Fig. \ref{fig:Isosurfaces} shows another volume-rendered image used in the survey.
Two iso-surfaces of a head dataset are depicted with \emph{translucent occlusion}, which is a type of visual multiplexing \cite{Chen:2014:CGF}.
Meanwhile, the voxels for soft tissue and muscle are not depicted at all, which can also been regarded as using a \emph{hollow visual channel}. 
The visual representation has been widely used, and the viewers are expected to use their knowledge to infer the 3D relationships between the two iso-surfaces as well as the missing information about soft tissue and muscle.
The question that accompanies the figure is for estimating such knowledge.

Although the survey offers only four options, it could in fact offer many other configurations as optional answers.
Let us consider four color-coded segments similar to the configurations in answers C and D.
Each segment could be one of four types: bone, skin, soft tissue and muscle, or background.
There are a total of $4^4=256$ configurations.
If one had to consider the variation of segment thickness, there would be many more options.
Because it would not be appropriate to ask a surveyee to select an answer from 256 options, a typical assumption is that the selected four options are representative.
In other words, considering that the 256 options are letters of an alphabet, any unselected letter has a probability similar to one of the four selected options. 

For example, we can estimate a ground truth PMF $Q$ such that among the 256 letters,
\begin{itemize}
  \vspace{-2mm}
  \item Answer A and four other letters have a probability 0.01,
  \vspace{-2mm}
  \item Answer B and 64 other letters have a probability 0.0002,
  \vspace{-2mm}
  \item Answer C and 184 other letters have a probability 0.0001,
  \vspace{-2mm}
  \item Answer D has a probability 0.9185.
\end{itemize}
We have the entropy of this alphabet $\mathcal{H}(Q)=0.85$.
Similar to the previous example, we can estimate the values of divergence as:

\begin{table}[h!]
  \vspace{-2mm}
  \centering
  \begin{tabular}{@{}l@{\hspace{3mm}}r@{\hspace{2.3mm}}r@{\hspace{2.3mm}}r@{\hspace{2.3mm}}r@{\hspace{2.3mm}}r@{}}
    Divergence for: & $\DJS$\; & $\DnewA$\; & $\DnewB$\; & $\DncmA$\; & $\DncmB$\;\\[1mm]
    A: $P = \{1, \ddddot{_4}, 0, \ddddot{_{64}}, 0, \ddddot{_{184}}, 0\}$
        & 0.960 & 0.933 & 0.903 & 0.993 & 0.986\\
    B: $P = \{0, \ddddot{_4}, 1, \ddddot{_{64}}, 0, \ddddot{_{184}}, 0\}$
        & 0.999 & 0.932 & 0.905 & 1.000 & 1.000\\
    C: $P = \{0, \ddddot{_4}, 0, \ddddot{_{64}}, 1, \ddddot{_{184}}, 0\}$
        & 0.999 & 0.932 & 0.905 & 1.000 & 1.000\\
    D: $P = \{0, \ddddot{_4}, 0, \ddddot{_{64}}, 0, \ddddot{_{184}}, 1\}$
        & 0.042 & 0.109 & 0.009 & 0.113 & 0.010
  \end{tabular}
  \vspace{-2mm}
%  \caption{Caption}
%  \label{tab:my_label}
\end{table}

\noindent where $\ddddot{_{n}}$ denotes $n$ zeros.
$\DJS$, $\DnewB$, $\DncmA$ and $\DncmB$ returned values indicating the same order of divergence, i.e., C $\sim$ B $>$ A $\gg$ D, which is consistent with PMF $Q'$.
Only $\DnewA$ returns an order A $>$ B $\sim$ C $\gg$ D.
This reinforces the observation by the previous example (i.e., Fig. \ref{fig:Arteries}) about the characteristics of ordering of the five measures.

For both examples (Figs. \ref{fig:Arteries} and \ref{fig:Isosurfaces}), because both $\DJS$, $\DnewB$, $\DncmA$ and $\DncmB$ have consistently returned sensible values, we give a score of 5 to each of them in Table \ref{tab:MultiCriteria}.
$\DnewA$ appears to be often incompatible with other measures in terms of ordering, we score $\DnewA$ 1. 

% With the maximum entropy being 8 bits, we can estimate the amounts of informative benefit as:
%
% \begin{align*}
%  &\text{with } \mathcal{D}_{\text{JS}},
%    &\text{A}: -6.826, \text{ B}: -7.139, \text{ C}: -7.144, \text{ D}: 0.514\\
%  &\text{with } \DnewB,
%    &\text{A}: -6.374, \text{ B}: -6.392, \text{ C}: -6.392, \text{ D}: 0.777
% \end{align*}
%

% such as \emph{partial occlusion}, , \emph{continuous field}, \emph{shifting visual channels}, and \emph{assuming a priori knowledge}

\begin{figure*}[t]
  \centering
  \includegraphics[height=46mm]{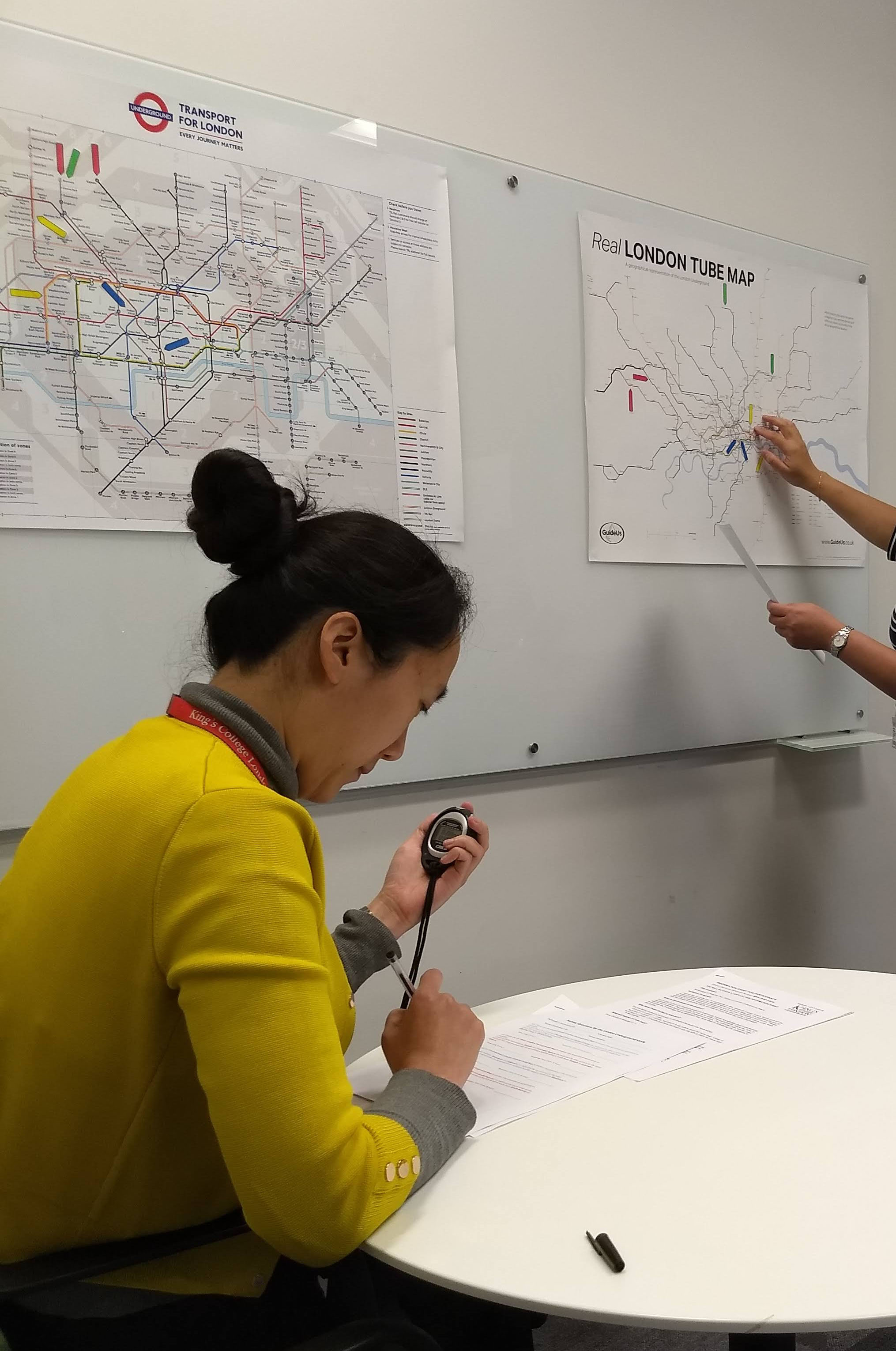}
  \includegraphics[height=46mm]{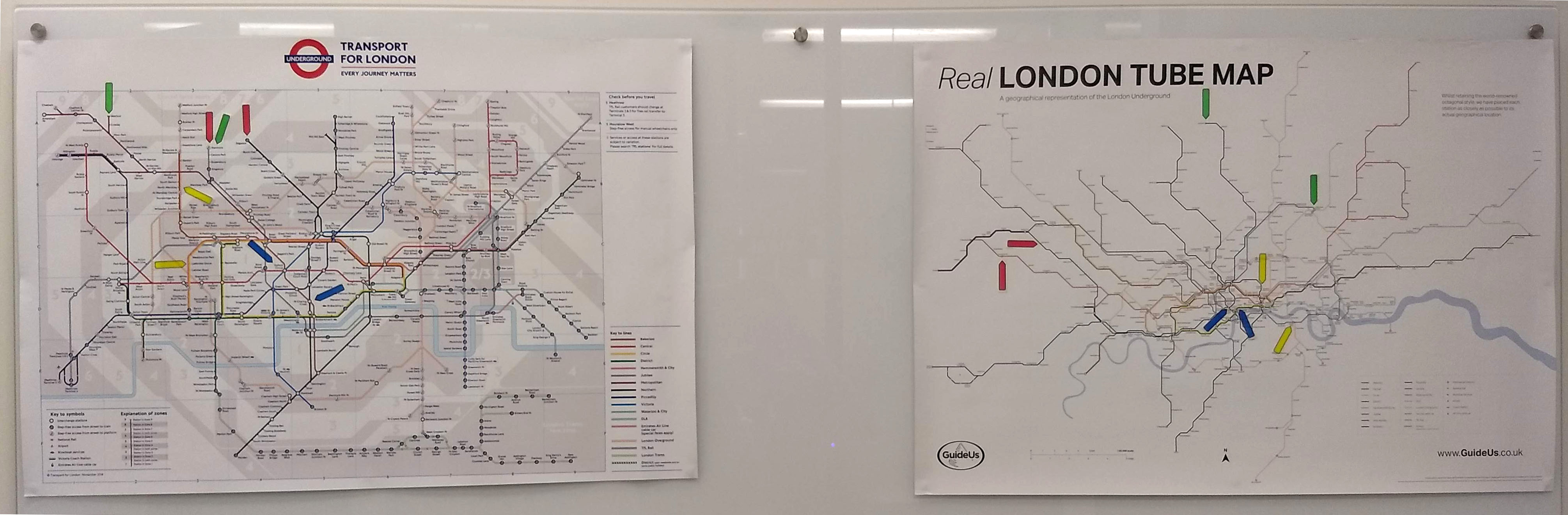}
  \caption{A survey for collecting data that reflects the use of some knowledge in viewing two types of London underground maps.}
  \label{fig:LondonMaps}
  \vspace{-4mm}
\end{figure*}

% --------------------------------------------------
\subsection{London Underground Map (Criterion 9)}
\label{sec:London}
This survey was designed to collect some real-world data that reflects the use of knowledge in viewing different London underground maps.
It involved sixteen surveyees, twelve at King's College London (KCL) and four at University of Oxford.
Surveyees were interviewed individually in a setup as shown in Fig. \ref{fig:LondonMaps}.  
Each surveyee was asked to answer 12 questions using either a geographically-faithful map or a deformed map, followed by two further questions about their familiarity of a metro system and London.
A \pounds5 Amazon voucher was offered to each surveyee as an appreciation of their effort and time.
The survey sheets and the full set of survey results are given in Appendix D.

Harry Beck first introduced geographically-deformed design of the London underground maps in 1931.
Today almost all metro maps around the world adopt this design concept.
Information-theoretically, the transformation of a geographically-faithful map to such a geographically-deformed map causes a significant loss of information.
Naturally, this affects some tasks more than others.

For example, the distances between stations on a deformed map are not as useful as in a faithful map.
The first four questions in the survey asked surveyees to estimate how long it would take to walk (i) from \emph{Charing Cross} to \emph{Oxford Circus},
(ii) from \emph{Temple} and \emph{Leicester Square},
(iii) from \emph{Stanmore} to \emph{Edgware}, and
(iv) from \emph{South Rulslip} to \emph{South Harrow}.
On the deformed map, the distances between the four pairs of the stations are all about 50mm.
On the faithful map, the distances are (i) 21mm, (ii) 14mm, (iii) 31mm, and (iv) 53mm respectively. 
According to the Google map, the estimated walk distance and time are (i) 0.9 miles, 20 minutes; (ii) 0.8 miles, 17 minutes; (iii) 1.6 miles, 32 minutes; and (iv) 2.2 miles, 45 minutes respectively.

The average range of the estimations about the walk time by the 12 surveyees at KCL are: (i) 19.25 [8, 30], (ii) 19.67 [5, 30], (iii) 46.25 [10, 240], and (iv) 59.17 [20, 120] minutes.
The estimations by the four surveyees at Oxford are:
(i) 16.25 [15, 20], (ii) 10 [5, 15], (iii) 37.25 [25, 60], and (iv) 33.75 [20, 60] minutes.
The values correlate better to the Google estimations than what would be implied by the similar distances on the deformed map.
Clearly some surveyees were using some knowledge to make better inference.

Let $\mathbb{Z}$ be an alphabet of integers between 1 and 256.
The range is chosen partly to cover the range of the answers in the survey, and partly to round  
up the maximum entropy $\mathbb{Z}$ to 8 bits.
For each pair of stations, we can define a PMF using a skew normal distribution peaked at the Google estimation $\xi$.
As an illustration, we coarsely approximate the PMF as $Q = \{q_i \; | \; 1 \le i \le 256 \}$, where
\begin{equation*}
  q_i = \begin{cases}
    0.01/236    & \text{if }\; 1 \le i \le \xi - 8  \hspace{10mm} (\textit{wild guess})\\
    0.026        & \text{if }\; \xi - 7 \le i \le \xi - 3 \quad \;\;\, (\textit{close})\\
    0.12        & \text{if }\; \xi - 2 \le i \le \xi + 2 \quad \;\;\, (\textit{spot on})\\
    0.026        & \text{if }\; \xi + 3 \le i \le \xi + 12 \quad (\textit{close})\\
    0.01/236    & \text{if }\; \xi + 13 \le i \le 256  \quad \;\;\, (\textit{wild guess})
  \end{cases}
\end{equation*}
Using the same way in the previous case study, we can estimate the divergence and the benefit of visualization for an answer in each range. Recall our observation of the phenomenon in Section \ref{sec:VolVis} that the measurements by $DJS$, $\DnewA$, $\DnewB$, $\DncmA$ and $\DncmB$ occupy different ranges of values, with $\DnewB$ be the most generous in measuring the benefit of visualization.
With the entropy of the alphabet as $\mathcal{H}(Q) \approx 3.6$ bits and the maximum entropy being 8 bits, the benefit values obtained for this example exhibit a similar but more compelling pattern:%

\begin{table}[h!]
  \vspace{-2mm}
  \centering
  \begin{tabular}{@{}lrrrrr@{}}
    Benefit for: & $\DJS$\; & $\DnewA$\; & $\DnewB$\; & $\DncmA$\; & $\DncmB$\;\\[1mm]
    \emph{spot on}    & $-1.765$ & $-0.418$ & \textbf{0.287} & $-3.252$ & $-2.585$\\
    \emph{close}      & $-3.266$ & $-0.439$ & \textbf{0.033} & $-3.815$ & $-3.666$\\
    \emph{wild guess} & $-3.963$ & $-0.416$ & $-0.017$ & $-3.966$ & $-3.965$
  \end{tabular}
  \vspace{-2mm}
%  \caption{Caption}
%  \label{tab:my_label}
\end{table}

\noindent Only $\DnewB$ has returned positive benefit values for \emph{spot on} and \emph{close} answers. Since it is not intuitive to say that those surveyees who gave good answers benefited from visualization negatively, clearly only the measurements returned by $\DnewB$ are intuitive. In addition, the ordering resulting from $\DnewA$ is again inconsistent with others.

For instance, surveyee P9, who has lived in a city with a metro system for a period of 1-5 years and lived in London for several months, made similarly good estimations about the walking time with both types of underground maps.
With one \emph{spot on} answer and one \emph{close} answer under each condition, the estimated benefit on average is $0.160$ bits if one uses $\DnewB$ and is negative if one uses any of the other four measures.
Meanwhile, surveyee P3, who has lived in a city with a metro system for two months, provided all four answers in the \emph{wild guess} category, leading to negative benefit values by all five measures.

Similarly, among the first set of four questions in the survey, Questions 1 and 2 are about stations near KCL, and Questions 3 and 4 are about stations more than 10 miles away from KCL.
The local knowledge of the surveyees from KCL clearly helped their answers.
Among the answers given by the twelve surveyees from KCL,
\begin{itemize}
    \vspace{-2mm}
    \item For Question 1, four \emph{spot on}, five \emph{close}, and three \emph{wild guess} --- the average benefit is $-2.940$ with $\mathcal{D}_{\text{JS}}$ or $0.105$ with $\DnewB$.
    \vspace{-2mm}
    \item For Question 2, two \emph{spot on}, nine \emph{close}, and one \emph{wild guess} ---
    the average benefit is $-3.074$ with $\mathcal{D}_{\text{JS}}$ or $0.071$ with $\DnewB$.
    \vspace{-2mm}
    \item For Question 3, three \emph{close}, and nine \emph{wild guess} ---
    the average benefit is $-3.789$ with $\mathcal{D}_{\text{JS}}$ or $-0.005$ with $\DnewB$.
    \vspace{-2mm}
    \item For Question 4, two \emph{spot on}, one \emph{close}, and nine \emph{wild guess} --- the average benefit is with $-3.539$ $\mathcal{D}_{\text{JS}}$ or $0.038$ with $\DnewB$.
\end{itemize}

The average benefit values returned by $\DnewA$, $\DncmA$, and $\DncmB$ are all negative for these four questions. Hence, unless $\DnewB$ is used, all other measures would semantically imply that both types of the London underground maps would have negative benefit. 
We therefore give $\DnewB$ a 5 score and $\DJS$, $\DncmA$, and $\DncmB$ a 3 score each in Table \ref{tab:MultiCriteria}.
We score $\DnewA$ 1 as it also exhibits an ordering issue.  

When we consider answering each of Questions 1$\sim$4 as performing a visualization task, we can estimate the cost-benefit ratio of each process.
As the survey also collected the time used by each surveyee in answering each question, the cost in Eq.\,\ref{eq:CBM-1} can be approximated with the mean response time.
For Questions 1$\sim$4, the mean response times by the surveyees at KCL are 9.27, 9.48, 14.65, and 11.40 seconds respectively.
Using the benefit values based on $\DnewB$, the cost-benefit ratios are thus 0.0113, 0.0075, -0.0003, and 0.0033 bits/second respectively. While these values indicate the benefits of the local knowledge used in answering Questions 1 and 2, they also indicate that when the local knowledge is absent in the case of Questions 3 and 4, the deformed map (i.e., Question 3) is less cost-beneficial.

% ====================
\section{Conclusions}
\label{sec:Conclusions}
In this paper, we have considered the need to improve the mathematical formulation of an information-theoretic measure for analyzing the cost-benefit of visualization as well as other processes in a data intelligence workflow \cite{Chen:2016:TVCG}.
The concern about the original measure is its unbounded term based on the KL-divergence.
We have obtained a proof that as long as the input and output alphabets of a process have a finite number of letters, the divergence measure used in the cost-benefit formula should be bounded.

We have considered a number of bounded measures to replace the unbounded term, including a new divergence measure $\Dnew$ and its variant $\Dncm$.
We have conducted multi-criteria decision analysis to select the best measure among these candidates.
In particular, we have used visualization to aid the observation of the mathematical properties of the candidate measures, assisting in the analysis of four criteria.
We have conducted two case studies, both in the form of surveys.
One consists of questions about volume visualizations, while the other features visualization tasks performed in conjunction with two types of London Underground maps.
The case studies allowed us to test some most promising candidate measures with the real world data collected in the two surveys, providing important evidence to two important aspects of the multi-criteria analysis.
From Table \ref{tab:MultiCriteria}, we can observe the process of narrowing down from eight candidate measures to five measures, and then to one.
Taking the importance of the criteria into account, we consider that candidate $\Dnew (k=2)$ is ahead of $\DJS$, critically because $\DJS$ often yields negative benefit values even when the benefit of visualization is clearly there.
We therefore propose to revise the original cost-benefit ratio in \cite{Chen:2016:TVCG} to the following:
\begin{equation} \label{eq:CBM-4}
  \begin{split}
          \frac{\text{Benefit}}{\text{Cost}} &= \frac{\text{Alphabet Compression} - \text{Potential Distortion}}{\text{Cost}}\\
          &= \frac{\SE(\mathbb{Z}_i) - \SE(\mathbb{Z}_{i+1}) - \SE_{\text{max}}(\mathbb{Z}_i) \mathcal{D}^2_{\text{new}}(\mathbb{Z}'_i || \mathbb{Z}_i)}{\text{Cost}}
  \end{split}
\end{equation}

This cost-benefit measure was developed in the field of visualization, for optimizing visualization processes and visual analytics workflows.
It is now being improved by using visual analysis and with the survey data collected in the context of visualization applications.
We would like to continue our theoretical investigation into the mathematical properties of the new divergence measure.
Meanwhile, having a bounded cost-benefit measure offers many new opportunities of developing tools for aiding the measurement and using such tools in practical applications, especially in visualization and visual analytics.

% ================

% The authors wish to thank A, B, and C. This work was supported
% in part by a grant from XYZ (\# 12345-67890).}

%\bibliographystyle{abbrv}
\bibliographystyle{abbrv-doi}

\bibliography{EstimatePD}

\begin{thebibliography}{10}

\bibitem{Biswas:2013:TVCG}
A.~Biswas, S.~Dutta, H.-W. Shen, and J.~Woodring.
\newblock An information-aware framework for exploring multivariate data sets.
\newblock {\em IEEE Transactions on Visualization and Computer Graphics},
  19(12):2683--2692, 2013.

\bibitem{Bordoloi:2005:Vis}
U.~Bordoloi and H.-W. Shen.
\newblock View selection for volume rendering.
\newblock In {\em Proc. IEEE Visualization}, pp. 487--494, 2005.

\bibitem{Bramon:2012:TVCG}
R.~Bramon, I.~Boada, A.~Bardera, Q.~Rodr\'{i}guez, M.~Feixas, J.~Puig, and
  M.~Sbert.
\newblock Multimodal data fusion based on mutual information.
\newblock {\em IEEE Transactions on Visualization and Computer Graphics},
  18(9):1574--1587, 2012.

\bibitem{Bramon:2013:CGF}
R.~Bramon, M.~Ruiz, A.~Bardera, I.~Boada, M.~Feixas, and M.~Sbert.
\newblock An information-theoretic observation channel for volume
  visualization.
\newblock {\em Computer Graphics Forum}, 32(3pt4):411--420, 2013.

\bibitem{Bramon:2013:JBHI}
R.~Bramon, M.~Ruiz, A.~Bardera, I.~Boada, M.~Feixas, and M.~Sbert.
\newblock Information theory-based automatic multimodal transfer function
  design.
\newblock {\em IEEE Journal of Biomedical and Health Informatics},
  17(4):870--880, 2013.

\bibitem{Bruckner:2010:CGF}
S.~Bruckner and T.~M\"{o}ller.
\newblock Isosurface similarity maps.
\newblock {\em Computer Graphics Forum}, 29(3):773--782, 2010.

\bibitem{Chen:2019:CGF}
M.~Chen and D.~S. Ebert.
\newblock An ontological framework for supporting the design and evaluation of
  visual analytics systems.
\newblock {\em Computer Graphics Forum}, 38(3):131--144, 2019.

\bibitem{Chen:2020:book}
M.~Chen and D.~J. Edwards.
\newblock `isms' in visualization.
\newblock In M.~Chen, H.~Hauser, P.~Rheingans, and G.~Scheuermann, eds., {\em
  Foundations of Data Visualization}. Springer, 2020.

\bibitem{Chen:2016:book}
M.~Chen, M.~Feixas, I.~Viola, A.~Bardera, H.-W. Shen, and M.~Sbert.
\newblock {\em Information Theory Tools for Visualization}.
\newblock A K Peters, 2016.

\bibitem{Chen:2019:TVCG}
M.~Chen, K.~Gaither, N.~W. John, and B.~McCann.
\newblock Cost-benefit analysis of visualization in virtual environments.
\newblock {\em IEEE Transactions on Visualization and Computer Graphics},
  25(1):32--42, 2019.

\bibitem{Chen:2016:TVCG}
M.~Chen and A.~Golan.
\newblock What may visualization processes optimize?
\newblock {\em IEEE Transactions on Visualization and Computer Graphics},
  22(12):2619--2632, 2016.

\bibitem{Chen:2017:CGA}
M.~Chen, G.~Grinstein, C.~R. Johnson, J.~Kennedy, and M.~Tory.
\newblock Pathways for theoretical advances in visualization.
\newblock {\em IEEE Computer Graphics and Applications}, 37(4):103--112, 2017.

\bibitem{Chen:2010:TVCG}
M.~Chen and H.~J\"anicke.
\newblock An information-theoretic framework for visualization.
\newblock {\em IEEE Transactions on Visualization and Computer Graphics},
  16(6):1206--1215, 2010.

\bibitem{Chen:2019:arXiv}
M.~Chen and M.~Sbert.
\newblock On the upper bound of the kullback-leibler divergence and cross
  entropy.
\newblock arXiv:1911.08334, 2019.

\bibitem{Chen:2014:CGF}
M.~Chen, S.~Walton, K.~Berger, J.~Thiyagalingam, B.~Duffy, H.~Fang,
  C.~Holloway, and A.~E. Trefethen.
\newblock Visual multiplexing.
\newblock {\em Computer Graphics Forum}, 33(3):241--250, 2014.

\bibitem{Correa:2006:TVCG}
C.~Correa, D.~Silver, and M.~Chen.
\newblock Feature aligned volume manipulation for illustration and
  visualization.
\newblock {\em IEEE Transactions on Visualization and Computer Graphics},
  12(5):1069--1076, 2006.

\bibitem{Cover:2006:book}
T.~M. Cover and J.~A. Thomas.
\newblock {\em Elements of Information Theory}.
\newblock John Wiley \& Sons, 2006.

\bibitem{Dasgupta:2012:CGF}
A.~Dasgupta, M.~Chen, and R.~Kosara.
\newblock Conceptualizing visual uncertainty in parallel coordinates.
\newblock {\em Computer Graphics Forum}, 31(3):1015–1024, 2012.

\bibitem{Feixas:2001:CGF}
M.~Feixas, E.~{del Acebo}, P.~Bekaert, and M.~Sbert.
\newblock An information theory framework for the analysis of scene complexity.
\newblock {\em Computer Graphics Forum}, 18(3):95--106, 1999.

\bibitem{Feixas:2009:AP}
M.~Feixas, M.~Sbert, and F.~Gonz\'{a}lez.
\newblock A unified information-theoretic framework for viewpoint selection and
  mesh saliency.
\newblock {\em ACM Transactions on Applied Perception}, 6(1):1--23, 2009.

\bibitem{Golin:2020:web}
M.~J. Golin.
\newblock Lecture 17: Huffman coding.
\newblock
  \url{http://home.cse.ust.hk/faculty/golin/COMP271Sp03/Notes/MyL17.pdf},
  accessed in March 2020.

\bibitem{Gumhold:2002:Vis}
S.~Gumhold.
\newblock Maximum entropy light source placement.
\newblock In {\em Proc. IEEE Visualization}, pp. 275--282, 2002.

\bibitem{Haseli:2020:IJMSEM}
G.~Haseli, R.~Sheikh, and S.~S. Sana.
\newblock Base-criterion on multi-criteria decision-making method and its
  applications.
\newblock {\em International Journal of Management Science and Engineering
  Management}, 15(2):79--88, 2020.

\bibitem{Ishizaka:2013:book}
A.~Ishizaka and P.~Nemery.
\newblock {\em Multi-Criteria Decision Analysis: Methods and Software}.
\newblock John Wiley \& Sons, 2013.

\bibitem{Jaenicke:2010:CGA}
H.~J\"anicke and G.~Scheuermann.
\newblock Visual analysis of flow features using information theory.
\newblock {\em IEEE Computer Graphics and Applications}, 30(1):40--49, 2010.

\bibitem{Jaenicke:2007:TVCG}
H.~J\"anicke, A.~Wiebel, G.~Scheuermann, and W.~Kollmann.
\newblock Multifield visualization using local statistical complexity.
\newblock {\em IEEE Transactions on Visualization and Computer Graphics},
  13(6):1384--1391, 2007.

\bibitem{Jung:2008:web}
Y.~Jung.
\newblock instantreality 1.0.
\newblock https://doc.instantreality.org/tutorial/volume-rendering/, last
  accessed in 2019.

\bibitem{Kijmongkolchai:2017:CGF}
N.~Kijmongkolchai, A.~Abdul-Rahman, and M.~Chen.
\newblock Empirically measuring soft knowledge in visualization.
\newblock {\em Computer Graphics Forum}, 36(3):73--85, 2017.

\bibitem{Kullback:1951:AMS}
S.~Kullback and R.~A. Leibler.
\newblock On information and sufficiency.
\newblock {\em Annals of Mathematical Statistics}, 22(1):79--86, 1951.

\bibitem{Lin:1991:TIT}
J.~Lin.
\newblock Divergence measures based on the shannon entropy.
\newblock {\em IEEE Transactions on Information Theory}, 37:145–151, 1991.

\bibitem{Max:2010:book}
N.~Max and M.~Chen.
\newblock Local and global illumination in the volume rendering integral.
\newblock In H.~Hagen, ed., {\em Scientific Visualization: Advanced Concepts}.
  Schloss Dagstuhl, Wadern, Germany, 2010.

\bibitem{Moser:2012:book}
S.~M. Moser.
\newblock {\em A Student's Guide to Coding and Information Theory}.
\newblock Cambridge University Press, 2012.

\bibitem{Nagy:2002:VMV}
Z.~Nagy, J.~Schneide, and R.~Westerman.
\newblock Interactive volume illustration.
\newblock In {\em Proc. Vision, Modeling and Visualization}, 2002.

\bibitem{Ng:2004:IV}
C.~U. Ng and G.~Martin.
\newblock Automatic selection of attributes by importance in relevance feedback
  visualisation.
\newblock In {\em Proc. Information Visualisation}, pp. 588--595, 2004.

\bibitem{Purchase:2008:LNCS}
H.~C. Purchase, N.~Andrienko, T.~J. Jankun-Kelly, and M.~Ward.
\newblock Theoretical foundations of information visualization.
\newblock In {\em Information Visualization: Human-Centered Issues and
  Perspectives}, Springer LNCS 4950, pp. 46--64. 2008.

\bibitem{Rezaei:2015:O}
J.~Rezaei.
\newblock Best-worst multi-criteria decision-making method.
\newblock {\em Omega}, 53:49--57, 2015.

\bibitem{Rigau:2005:SMA}
J.~Rigau, M.~Feixas, and M.~Sbert.
\newblock Shape complexity based on mutual information.
\newblock In {\em Proc. IEEE Shape Modeling and Applications}, 2005.

\bibitem{Roettger:2019:web}
S.~Roettger.
\newblock The volume library.
\newblock http://schorsch.efi.fh-nuernberg.de/data/volume/, last accessed in
  2019.

\bibitem{Ruiz:2011:TVCG}
M.~Ruiz, A.~Bardera, I.~Boada, I.~Viola, M.~Feixas, and M.~Sbert.
\newblock Automatic transfer functions based on informational divergence.
\newblock {\em IEEE Transactions on Visualization and Computer Graphics},
  17(12):1932--1941, 2011.

\bibitem{Shannon:1948:BSTJ}
C.~E. Shannon.
\newblock A mathematical theory of communication.
\newblock {\em Bell System Technical Journal}, 27:379--423, 1948.

\bibitem{Streeb:2019:TVCG}
D.~Streeb, M.~El-Assady, D.~Keim, and M.~Chen.
\newblock Why visualize? untangling a large network of arguments.
\newblock {\em IEEE Transactions on Visualization and Computer Graphics}, early
  view, 2019.
\newblock 10.1109/TVCG.2019.2940026.

\bibitem{Takahashi:2005:Vis}
S.~Takahashi and Y.~Takeshima.
\newblock A feature-driven approach to locating optimal viewpoints for volume
  visualization.
\newblock In {\em Proc. IEEE Visualization}, pp. 495--502, 2005.

\bibitem{Tam:2017:TVCG}
G.~K.~L. Tam, V.~Kothari, and M.~Chen.
\newblock An analysis of machine- and human-analytics in classification.
\newblock {\em IEEE Transactions on Visualization and Computer Graphics},
  23(1), 2017.

\bibitem{Vazquez:2004:CGF}
P.-P. V\'{a}zquez, M.~Feixas, M.~Sbert, and W.~Heidrich.
\newblock Automatic view selection using viewpoint entropy and its application
  to image-based modelling.
\newblock {\em Computer Graphics Forum}, 22(4):689--700, 2004.

\bibitem{Viola:2019:book}
I.~Viola, M.~Chen, and T.~Isenberg.
\newblock Visual abstraction.
\newblock In M.~Chen, H.~Hauser, P.~Rheingans, and G.~Scheuermann, eds., {\em
  Foundations of Data Visualization}. Springer, 2020.
\newblock Preprint at arXiv:1910.03310, 2019.

\bibitem{Viola:2006:TVCG}
I.~Viola, M.~Feixas, M.~Sbert, and M.~E. Gr{\"o}ller.
\newblock Importance-driven focus of attention.
\newblock {\em IEEE Transactions on Visualization and Computer Graphics},
  12(5):933--940, 2006.

\bibitem{Wang:2006:TVCG}
C.~Wang and H.-W. Shen.
\newblock {LOD Map} - a visual interface for navigating multiresolution volume
  visualization.
\newblock {\em IEEE Transactions on Visualization and Computer Graphics},
  12(5):1029--1036, 2005.

\bibitem{Wang:2011:E}
C.~Wang and H.-W. Shen.
\newblock Information theory in scientific visualization.
\newblock {\em Entropy}, 13:254--273, 2011.

\bibitem{Wang:2008:TVCG}
C.~Wang, H.~Yu, and K.-L. Ma.
\newblock Importance-driven time-varying data visualization.
\newblock {\em IEEE Transactions on Visualization and Computer Graphics},
  14(6):1547--1554, 2008.

\bibitem{Wei:2013:CGF}
T.-H. Wei, T.-Y. Lee, and H.-W. Shen.
\newblock Evaluating isosurfaces with level-set-based information maps.
\newblock {\em Computer Graphics Forum}, 32(3):1--10, 2013.

\bibitem{Wu:2007:TVCG}
Y.~{Wu} and H.~{Qu}.
\newblock Interactive transfer function design based on editing direct volume
  rendered images.
\newblock {\em IEEE Transactions on Visualization and Computer Graphics},
  13(5):1027--1040, 2007.

\bibitem{Xu:2010:TVCG}
L.~Xu, T.~Y. Lee, and H.~W. Shen.
\newblock An information-theoretic framework for flow visualization.
\newblock {\em IEEE Transactions on Visualization and Computer Graphics},
  16(6):1216--1224, 2010.

\end{thebibliography}

% ================
\newpage
\noindent\huge%
\textsc{\textsf{\textbf{Appendices}}}

\noindent\LARGE%
\textbf{A Bounded Measure for Estimating\\the Benefit of Visualization}

\noindent\large%
~\\
Min Chen, University of Oxford, UK\\
Mateu Sbert, University of Girona, Spain\\
Alfie Abdul-Rahman, King's College London, UK\\
Deborah Silver, Rutgers University, USA\\

\normalsize%

\appendix
% =================================================
\section{Explanation of the Original Cost-Benefit Measure}
\label{app:OriginalTheory}
This appendix contains an extraction from a previous publication \cite{Chen:2019:CGF}, which provides a relatively concise but informative description of the cost-benefit ratio proposed in \cite{Chen:2016:TVCG}. The inclusion of this is to minimize the readers' effort to locate such an explanation. The extraction has been slightly modified.
In addition, at the end of this appendix, we provide a relatively informal and somehow conversational discussion about using this measure to explain why visualization is useful.

Chen and Golan introduced an information-theoretic metric for measuring the cost-benefit ratio of a visual analytics (VA) workflow or any of its component processes \cite{Chen:2016:TVCG}.
The metric consists of three fundamental measures that are abstract representations of a variety of qualitative and quantitative criteria used in practice, including
operational requirements (e.g., accuracy, speed, errors, uncertainty, provenance, automation),
analytical capability (e.g., filtering, clustering, classification, summarization),
cognitive capabilities (e.g., memorization, learning, context-awareness, confidence), and so on.
The abstraction results in a metric with the desirable mathematical simplicity \cite{Chen:2016:TVCG}.
The qualitative form of the metric is as follows:
\begin{equation}
\label{eq:CBR}
\frac{\textit{Benefit}}{\textit{Cost}} = \frac{\textit{Alphabet Compression} - \textit{Potential Distortion}}{\textit{Cost}}
\end{equation}

The metric describes the trade-off among the three measures:

\begin{itemize}
\vspace{-1mm}
\item
\emph{Alphabet Compression} (AC) measures the amount of entropy reduction (or information loss) achieved by a process.
As it was noticed in \cite{Chen:2016:TVCG}, most visual analytics processes (e.g., statistical aggregation, sorting, clustering, visual mapping, and interaction), feature many-to-one mappings from input to output, hence losing information.
Although information loss is commonly regarded harmful, it cannot be all bad if it is a general trend of VA workflows.
Thus the cost-benefit metric makes AC a positive component.
\vspace{-1mm}
\item
\emph{Potential Distortion} (PD) balances the positive nature of AC by measuring the errors typically due to information loss. Instead of measuring mapping errors using some third party metrics, PD measures the potential distortion when one reconstructs inputs from outputs.
The measurement takes into account humans' knowledge that can be used to improve the reconstruction processes. For example, given an average mark of 62\%, the teacher who taught the class can normally guess the distribution of the marks among the students better than an arbitrary person.
\vspace{-1mm}
\item
\emph{Cost} (Ct) of the forward transformation from input to output and the inverse transformation of reconstruction provides a further balancing factor in the cost-benefit metric in addition to the trade-off between AC and PD. In practice, one may measure the cost using \emph{time} or a monetary measurement.
\end{itemize}

\vspace{2mm}
\noindent\textbf{Why is visualization useful?}
There have been many arguments about why visualization is useful. 
Streeb et al. collected a large number of arguments and found many arguments are in conflict with each other \cite{Streeb:2019:TVCG}.
Chen and Edwards presented an overview of schools of thought in the field of visualization, and showed that the ``why'' question is a bone of major contention \cite{Chen:2020:book}.

The most common argument about ``why'' question is because visualization offers insight or helps humans to gain insight. When this argument is used outside the visualization community, there are often counter-arguments that statistics and algorithms can offer insight automatically and often with better accuracy and efficiency. There are also concerns that visualization may mislead viewers, which cast further doubts about the usefulness of visualization, while leading to a related argument that ``visualization must be accurate'' in order for it to be useful.
The accuracy argument itself is not bullet-proof since there are many types of uncertainty in a visualization process, from uncertainty in data, to that caused by visual mapping, and to that during perception and cognition \cite{Dasgupta:2012:CGF}.
Nevertheless, it is easier to postulate that visualization must be accurate, as it seems to be counter-intuitive to condone the idea that ``visualization can be inaccurate,'' not mentioning the idea of ``visualization is normally inaccurate,'' or ``visualization should be inaccurate.''

The word ``inaccurate'' is itself an abstraction of many different types of inaccuracy.
Misrepresentation truth is a type of inaccuracy.
Such acts are mostly wrong, but some (such as wordplay and sarcasm) may cause less harm.
Converting a student's mark in the range of [0, 100] to the range of [A, B, C, D, E, F] is another type of inaccuracy.
This is a common practice, and must be useful.
From an information-theoretic perspective, these two types of inaccuracy are information loss.

In their paper \cite{Chen:2016:TVCG}, Chen and Golan observed that statistics and algorithms usually lose more information than visualization. Hence, this provides the first hint about the usefulness of visualization. They also noticed that like wordplay and sarcasm, the harm of information loss can be alleviated by knowledge. For someone who can understand a workplay (e.g., a pun) or can sense a sarcastic comment, the misrepresentation can be corrected by that person at the receiving end. This provides the second hint about the usefulness of visualization because any ``misrepresentation'' in visualization may be corrected by a viewer with appropriate knowledge.

On the other hand, statistics and algorithms are also useful, and sometimes more useful than visualization. Because statistics and algorithms usually cause more information loss, some aspects of information loss must be useful.
One important merit of losing information in one process is that the succeeding process has less information to handle, and thus incurs less cost.
This is why Chen and Golan divided information loss into two components, a positive component called alphabet compression and a negative component called potential distortion \cite{Chen:2016:TVCG}.

The positive component explains why statistics, algorithms, visualization, and interaction are useful because they all lose information.
The negative component explains why they are sometimes less useful because information loss may cause distortion during information reconstruction.
Both components are moderated by the cost of a process (i.e., statistics, algorithms, visualization, or interaction) in losing information and reconstructing the original information.
Hence, given a dataset, the best visualization is the one that loses most information while causing the least distortion.
This also explains why visual abstraction is effective when the viewers have adequate knowledge to reconstruct the lost information and may not be effective otherwise \cite{Viola:2019:book}.

The central thesis by Chen and Golan \cite{Chen:2016:TVCG} may appear to be counter-intuitive to many as it suggests ``inaccuracy is a good thing'', partly because the word ``inaccuracy'' is an abstraction of many meanings and itself features information loss. Perhaps the reason for the conventional wisdom is that it is relatively easy to think that ``visualization must be accurate''. To a very small extent, this is a bit like the easiness to think ``the earth is flat'' a few centuries ago, because the evidence for supporting that wisdom was available everywhere, right in front of everyone at that time.
Once we step outside the field of visualization, we can see the phenomena of inaccuracy everywhere, in statistics and algorithms as well as in visualization and interaction.
All these suggest that ``the earth may not be flat,'' or ``inaccuracy can be a good thing.''

In summary, the cost-benefit measure by Chen and Golan \cite{Chen:2016:TVCG} explains that when visualization is useful, it is because visualization has a better trade-off than simply reading the data, simply using statistics alone, or simply relying on algorithms alone.
The ways to achieve a better trade-off include: (i) visualization may lose some information to reduce the human cost in observing and analyzing the data, (ii) it may lose some information since the viewers have adequate knowledge to recover such information or can acquire such knowledge at a lower cost, (iii) it may preserve some information because it reduces the reconstruction distortion in the current and/or succeeding processes, and (iv) it may preserve some information because the viewers do not have adequate knowledge to reconstruct such information or it would cost too much to acquire such knowledge.  

% =================================================
\section{Formulae of the Basic and Relevant Information-Theoretic Measures}
\label{app:InfoTheory}
This section is included for self-containment. Some readers who have the essential knowledge of probability theory but are unfamiliar with information theory may find these formulas useful.

Let $\mathbb{Z} = \{ z_1, z_2, \ldots, z_n \}$ be an alphabet and $z_i$ be one of its letters.
$\mathbb{Z}$ is associated with a probability distribution or probability mass function (PMF) $P(\mathbb{Z}) = \{ p_1, p_2, \ldots, p_n \}$ such that
$p_i = p(z_i) \ge 0$ and $\sum_{1}^n p_i = 1$. The \textbf{Shannon Entropy} of $\mathbb{Z}$ is:

\[
  \mathcal{H}(\mathbb{Z}) = \mathcal{H}(P)= - \sum_{i=1}^n p_i \log_2 p_i \quad \text{(unit: bit)}
\]

Here we use base 2 logarithm as the unit of bit is more intuitive in context of computer science and data science.

An alphabet $\mathbb{Z}$ may have different PMFs in different conditions.
Let $P$ and $Q$ be such PMFs. The \textbf{KL-Divergence} $\mathcal{D}_{KL}(P||Q)$ describes the difference between the two PMFs in bits:
\[
  \mathcal{D}_{KL}(P||Q) = \sum_{i=1}^n p_i \log_2 \frac{p_i}{q_i} \quad \text{(unit: bit)}
\]
$\mathcal{D}_{KL}(P||Q)$ is referred as the divergence of $P$ from $Q$.
This is not a metric since $\mathcal{D}_{KL}(P||Q) \equiv \mathcal{D}_{KL}(Q||P)$ cannot be assured.

Related to the above two measures, \textbf{Cross Entropy} is defined as:
\[
  \mathcal{H}(P, Q) = \mathcal{H}(P) + \mathcal{D}_{KL}(P||Q) = - \sum_{i=1}^n p_i \log_2 q_i \quad \text{(unit: bit)}
\]
Sometimes, one may consider $\mathbb{Z}$ as two alphabets $\mathbb{Z}_a$ and $\mathbb{Z}_b$ with the same ordered set of letters but two different PMFs.
In such case, one may denote the KL-Divergence as $\mathcal{D}_{KL}(\mathbb{Z}_a||\mathbb{Z}_b)$, and the cross entropy as $\mathcal{H}(\mathbb{Z}_a, \mathbb{Z}_b)$.

% =================================================
\section{Conceptual Boundedness of $\CE(P, Q)$ and $\mathcal{D}_\text{KL}$}
\label{app:Proof}
For readers who are not familiar with information-theoretic notations and measures, it is helpful to read Appendix \ref{app:InfoTheory} first.
According to the mathematical definition of cross entropy:
\begin{equation} \label{eq:CrossEntropy}
    \mathcal{H}(P, Q) = - \sum_{i=1}^n p_i \log_2 q_i = \sum_{i=1}^n p_i \log_2 \frac{1}{q_i}
\end{equation}
\noindent $\SE(P,Q)$ is of course unbounded.
When $q_i \rightarrow 0$, we have $\log_2 \frac{1}{q_i} \rightarrow \infty$. As long as $p_i \neq 0$ and is independent of $q_i$, $\SE(P,Q) \rightarrow \infty$.
Hence the discussion in this appendix is not about a literal proof that $\SE(P,Q)$ is unbounded when this mathematical formula is applied without any change.
It is about that the concept of cross entropy implies that it should be bounded when $n$ is a finite number.

\vspace{2mm}\noindent\textbf{Definition 1.}
\emph{Given an alphabet $\mathbb{Z}$ with a true PMF $P$, cross-entropy $\SE(P,Q)$ is the average number of bits required when encoding $\mathbb{Z}$ with an alternative PMF $Q$.}
\vspace{2mm}

This is a widely-accepted and used definition of cross-entropy in the literature of information theory.
The concept can be observed from the formula of Eq.\,\ref{eq:CrossEntropy}, where $\log_2 (1/q_i)$ is considered as the mathematically-supposed length of a codeword that is used to encode letter $z_i \in \mathbb{Z}$ with a probability value $q_i \in Q$. Because $\sum_{i=1}^n p_i = 1$, $\SE(P,Q)$ is thus the weighted or probabilistic average length of the codewords for all letters in $\mathbb{Z}$.

Here a \emph{codeword} is the digital representation of a letter in an alphabet. A \emph{code} is a collection of the codewords for all letters in an alphabet. In communication and computer science, we usually use binary codes as digital representations for alphabets, such as ASCII code and variable-length codes.

However, when a letter $z_i \in \mathbb{Z}$ is given a probability value $q_i$, it is not necessary for $z_i$ to be encoded using a codeword of length $\log_2 (1/q_i)$ bits, or more precisely, the nearest integer above or equal to it, i.e., $\lceil log_2 (1/q_i) \rceil$ bits, since a binary codeword cannot have fractional bits digitally.
For example, consider a simple alphabet $\mathbb{Z} = \{z_1, z_2\}$.
Regardless what PMF is associated with $\mathbb{Z}$, $\mathbb{Z}$ can always be encoded with a 1-bit code, e.g., codeword 0 for $z_1$ and codeword 1 for $z_2$, as long as neither of the two probability values in $P$ is zero, i.e., $p_1 \neq 0$ and $p_2 \neq 0$.
However, if we had followed Eq.\,\ref{eq:CrossEntropy} literally, we would have created codes similar to the following examples:
\begin{itemize}
    \item if $P = \{ \frac{1}{2}, \frac{1}{2} \}$, codeword 0 for $z_1$ and codeword 1 for $z_2$;
    \item if $P = \{ \frac{3}{4}, \frac{1}{4} \}$, codeword 0 for $z_1$ and codeword 10 for $z_2$;
    \item $\cdots$
    \item if $P = \{ \frac{63}{64}, \frac{1}{64} \}$, codeword 0 for $z_1$ and codeword 111111 for $z_2$;
    \item $\cdots$
\end{itemize}

\begin{figure} [t]
    \centering
    \includegraphics[width=88mm]{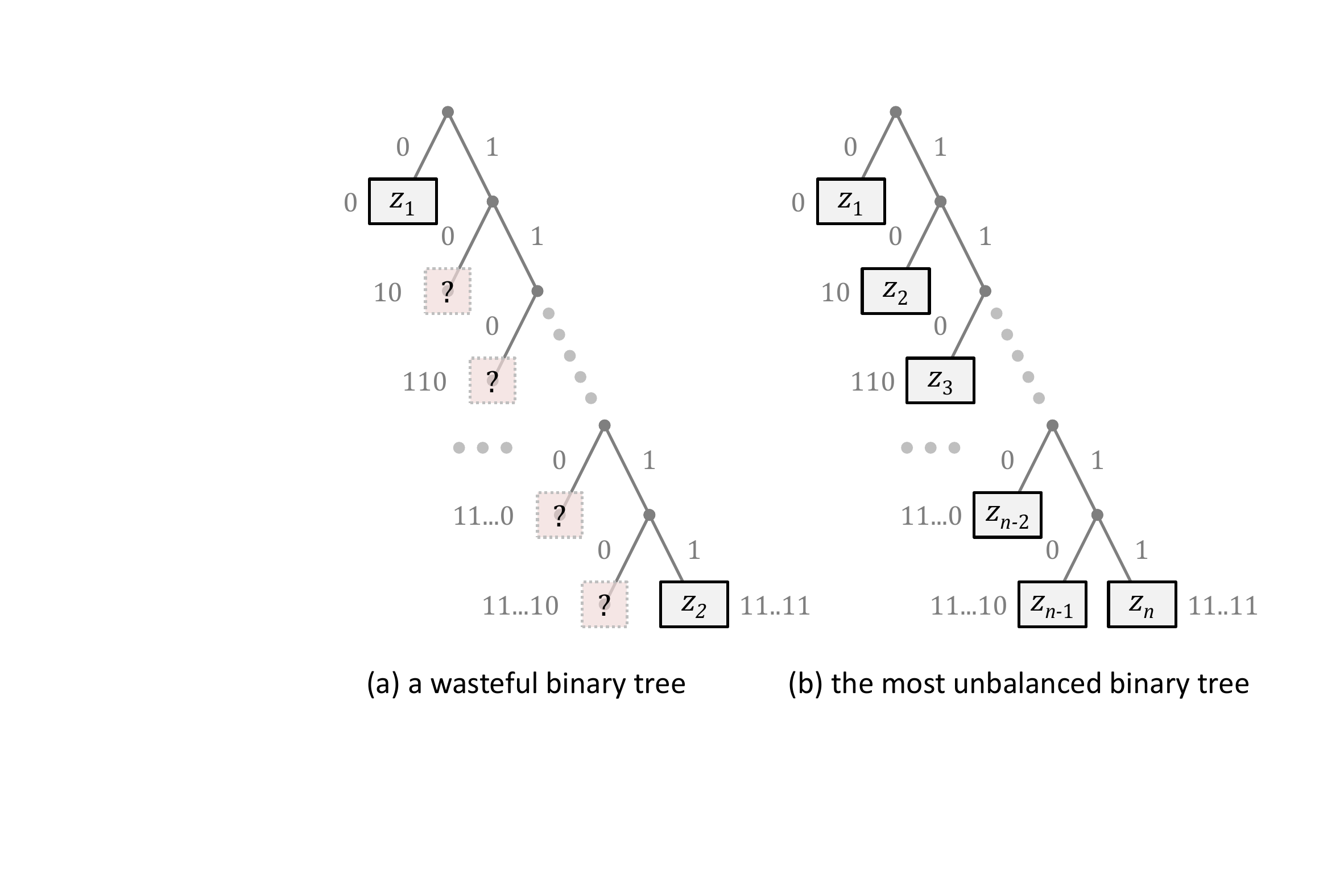}
    \caption{Two examples of binary codes illustrated as binary trees.}
    \label{fig:BinaryTree}
\end{figure}
As shown in Fig. \ref{fig:BinaryTree}(a), such a code is very wasteful.
Hence, in practice, encoding $\mathbb{Z}$ according to Eq.\,\ref{eq:CrossEntropy} literally is not desirable. Note that the discussion about encoding is normally conducted in conjunction with the Shannon entropy. Here we use the cross entropy formula for our discussion to avoid a deviation from the flow of reasoning.

Let $\mathbb{Z}$ be an alphabet with a finite number of letters, $\{ z_1, z_2, \ldots, z_n \}$, and $\mathbb{Z}$ is associated with a PMF, $Q$, such that: 
\begin{equation} \label{eq:CodeP-APP}
\begin{split}
  q(z_n) &= \epsilon, \quad\text{(where $0 < \epsilon < 2^{-(n-1)}$}),\\
  q(z_{n-1}) &= (1-\epsilon)2^{-(n-1)},\\
  q(z_{n-2}) &= (1-\epsilon)2^{-(n-2)},\\
  &\cdots\\
  q(z_{2}) &= (1-\epsilon)2^{-2},\\
  q(z_{1}) &= (1-\epsilon)2^{-1} + (1-\epsilon)2^{-(n-1)}.
\end{split}
\end{equation}
We can encode this alphabet using the Huffman encoding that is a practical binary coding scheme and adheres the principle to obtain a code with the Shannon entropy as the average length of codewords \cite{Moser:2012:book}.
Entropy coding is designed to minimize the average number of bits per letter when one transmits a ``very long'' sequence of letters in the alphabet over a communication channel.
Here the phrase ``very long'' implies that the string exhibits the above PMF $Q$ (Eq.\,\ref{eq:CodeP-APP}).
In other words, given an alphabet $\mathbb{Z}$ and a PMF $Q$, the Huffman encoding algorithm creates an optimal code with the lowest average length of codewords when the code is used to transmit a ``very long'' sequence of letters in $\mathbb{Z}$.
One example of such a code for the above PMF $Q$ is:
\begin{equation} \label{eq:Code-APP}
\begin{split}
  z_1&: 0, \qquad z_2: 10, \qquad z_3: 110\\
  &\cdots\\
  z_{n-1}&: 111\ldots10 \quad\text{(with $n-2$ ``1''s and one ``0'') }\\
  z_n&: 111\ldots11 \quad\text{(with $n-1$ ``1''s and no ``0'') }
\end{split}
\end{equation}
Fig. \ref{fig:BinaryTree}(b) shows illustrates such a code using a binary tree.
In this way, $z_n$, which has the smallest probability value, will always be assigned a codeword with the maximum length of $n-1$.

\vspace{2mm}\noindent\textbf{Lemma 1.}
\emph{Let $\mathbb{Z}$ be an alphabet with $n$ letters and $\mathbb{Z}$ is associated with a PMF $P$.
If $\mathbb{Z}$ is encoded using the aforementioned entropy coding, the maximum length of any codeword for $z_i \in \mathbb{Z}$ is always $\leq n-1$.}
\vspace{2mm}

We can prove this lemma by confirming that when one creates a binary code for an $n$-letter alphabet $\mathbb{Z}$, the binary tree shown in Fig. \ref{fig:BinaryTree}(b) is the worst unbalanced tree without any wasteful leaf nodes.
Visually, we can observe that the two letters with the lowest values always share the lowest internal node as their parent node. The remaining $n-2$ letters are to be hung on the rest binary subtree.
Because the subtree is not allowed to waste leaf space, the $n-2$ leaf nodes can be comfortably hung on the root and up to $n-3$ internal node.
A formal proof can be obtained using induction.
For details, readers may find Golin's lecture notes useful \cite{Golin:2020:web}.
See also \cite{Cover:2006:book} for related mathematical theorems.

\vspace{2mm}\noindent\textbf{Theorem 1.}
\emph{Let $\mathbb{Z}$ be an alphabet with a finite number of letters and $\mathbb{Z}$ is associated with two PMFs, $P$ and $Q$. With the Huffman encoding, conceptually the cross entropy $\SE(P,Q)$ should be bounded.}
\vspace{2mm}

Let $n$ be the number of letters in $\mathbb{Z}$. 
According to \textbf{Lemma 1}, when $\mathbb{Z}$ is encoded in conjunction with PMF $Q$ using the Huffman encoding, the maximum codeword length is $\leq n-1$.
In other words, in the worst case scenario, there is letter $z_k \in \mathbb{Z}$ that has the lowest probability value $q_k$, i.e., $q_k \leq q_j \forall j=1, 2, \ldots n \text{ and } j \neq k$.
With the Huffman encoding, $z_k$ will be encoded with the longest codeword of up to $n-1$ bits.

According to \textbf{Definition 1}, there is a true PMF $P$.
Let $L(z_i, q_i)$ be the codeword length of $z_i \in \mathbb{Z}$ determined by the Huffman encoding.
We can write a conceptual cross entropy formula as:
\[
  \SE(P, Q) = \sum_{i=1}^n p_i \cdot L(z_i, q_i) \leq \sum_{i=1}^n p_i \cdot L(z_k, q_k) \leq n-1
\]
\noindent where $q_k$ is the lowest probability value in $Q$ and $z_k$ is encoded with a codeword of up to $n-1$ bits (i.e., $L(z_k, q_k) \leq n-1$).
Hence conceptually $\SE(P, Q)$ is bounded by $n-1$ if the Huffman encoding is used.
Since we can find a bounded solution for any $n$-letter alphabet with any PMF, the claim of unboundedness has been falsified.
$\blacksquare$

\vspace{2mm}\noindent\textbf{Corollary 1.}
\emph{Let $\mathbb{Z}$ be an alphabet with a finite number of letters and $\mathbb{Z}$ is associated with two PMFs, $P$ and $Q$. With the Huffman encoding, conceptually the KL-divergence $\DKL(P\|Q)$ should be bounded.}
\vspace{2mm}

For an alphabet $\mathbb{Z}$ with a finite number of letters, the Shannon entropy $\mathcal{H}(P)$ is bounded regardless any PMF $P$.
The upper bound of $\mathcal{H}(P)$ is $\log_2 n$, where $n$ is the number of letters in $\mathbb{Z}$. 
Since we have
\begin{align*}
    &\mathcal{H}(P, Q) = \mathcal{H}(P) + \mathcal{D}_{KL}(P||Q)\\
    &\mathcal{D}_{KL}(P||Q) = \mathcal{H}(P, Q) - \mathcal{H}(P)
\end{align*}
\noindent using \textbf{Theorem 1}, we can infer that with the Huffman encoding, conceptually $\mathcal{D}_{KL}(P||Q)$ is also bounded.
$\blacksquare$

\vspace{2mm}
\noindent\textbf{Further discussion.}
The code created using Huffman encoding is also considered to be optimal for source coding (i.e., assuming without the need for error correction and detection). A formal proof can be found in \cite{Golin:2020:web}.

Let $\mathbb{Z}$ be an $n$-letter alphabet, and $Q$ be its associated PMF. When we use the Shannon entropy to determine the length of each codeword mathematically, we have:
\[
    L(z_i, q_i) = \lceil \log_2 \frac{1}{q_i} \rceil, \quad z_i \in \mathbb{Z}, q_i \in Q
\]
As we showed before, the length of a codeword can be infinitely long if $q_i \rightarrow 0$. Huffman encoding makes the length finite as long as $n$ is finite. This difference between the mathematically-literal entropy encoding and Huffman encoding is important to our proof that conceptually $\SE(P, Q)$ and $\DKL(P\|Q)$ are bounded.

However, we should not draw a conclusion that there is much difference between the communication efficiency gained based on the mathematically-literal entropy encoding and that gained using the Huffman encoding.
In fact, in terms of  the average length of codewords, they differ by less than one bit since both lie between $\SE(Q)$ and $\SE(Q)+1$ \cite{Cover:2006:book}, although their difference in terms of the maximum length of individual letters can be very different.

For example, if $\mathbb{Z}$ is a two-letter alphabet, and its PMF $Q$ is $\{0.999, 0.001\}$, the Huffman encoding results in a code with one bit for each letter, while the mathematically-literal entropy encoding results in 1 bit for $z_1 \in \mathbb{Z}$ and 10 bits for $z_2 \in \mathbb{Z}$. The probabilistic average length of the two codewords, which indicate the communication efficiency, is 1 bit for the Huffman encoding, and 1.009 bits for the mathematically-literal entropy encoding, while the entropy $\SE(Q)$ is 0.0114 bits. As predicted, $0.0114 < 1 < 1.009 < 1.0114$.

Consider another example with a five-letter alphabet and $Q = \{0.45, 0.20, 0.15, 0.15, 0.05\}$.
The mathematically-literal entropy encoding assigns five codewords with lengths of $\{2, 3, 3, 3, 5\}$, while the Huffman encoding assigns codewords with lengths of $\{1, 3, 3, 3, 3\}$.
The probabilistic average length of the former is 2.65, while that of the Huffman encoding is 2.1, while the entropy $\SE(Q)$ is 2.0999.
As predicted, $2.0999 < 2.1 < 2.65 < 3.0999$.

\begin{table*}[t]
  \centering
  \caption{The answers by ten surveyees to the questions in the volume visualization survey. The surveyees are ordered from left to right according to their own ranking about their knowledge of volume visualization. Correct answers are indicated by letters in brackets. The upper case letters (always in brackets) are the most appropriate answers, while the lower case letters with brackets are acceptable answers as they are correct in some circumstances. The lower case letters without brackets are incorrect answers.}
  \begin{tabular}{@{}l|cccccccccc@{}}
  & \multicolumn{8}{c}{\textbf{Surveyee's ID}}\\
 \textbf{Questions (with correct answers in brackets)}
     & S1 & S2 & S3 & S4 & S5 & S6 & S7 & S8 & P9 & P10\\
  \hline
  1. Use of different transfer functions (D), dataset: Carp
     & (D) & (D) & (D) & (D) & (D) & c & b & (D) & a & c\\
  2. Use of translucency in volume rendering (C), dataset: Engine Block
     & (C) & (C) & (C) & (C) & (C) & (C) & (C) & (C) & d & (C)\\
  3. Omission of voxels of soft tissue and muscle (D), dataset: CT head
     & (D) & (D) & (D) & (D) & b & b & a & (D) & a & (D)\\
  4. sharp objects in volume-rendered CT data (C), dataset: CT head
     & (C) & (C) & a & (C) & a & b & d & b & b & b\\ 
  5. Loss of 3D information with MIP (B, a), dataset: Aneurism
     & (a) & (B) & (a) & (a) & (a) & (a) & D & (a) & (a) & (a)\\
  6. Use of volume deformation (A), dataset: CT head
     & (A) & (A) & b & (A) & (A) & b & b & (A) & b & b\\
  7. Toe nails in non-photo-realistic volume rendering (B, c): dataset: Foot
     & (c) & (c) & (c) & (B) & (c) & (B) & (B) & (B) & (B) & (c)\\
  8. Noise in non-photo-realistic volume rendering (B): dataset: Foot
     & (B) & (B) & (B) & (B) & (B) & (B) & a & (B) & c & (B)\\
  \hline
  9. Knowledge about 3D medical imaging technology [1 lowest. 5 highest]
     & 4 & 3 & 4 & 5 & 3 & 3 & 3 & 3 & 2 & 1\\
  10. Knowledge about volume rendering techniques [1 lowest. 5 highest]
     & 5 & 5 & 4-5 & 4 & 4 & 3 & 3 & 3 & 2 & 1\\
  \hline
  \end{tabular}
  \label{tab:VolVis}
  \vspace{-2mm}
\end{table*}

% =================================================
\section{Survey Results of Useful Knowledge in Volume Visualization}
\label{app:VolVis}
This survey consists of eight questions presented as slides.
The questionnaire is given as part of the supplementary materials. 
The ten surveyees are primarily colleagues from the UK, Spain, and the USA.
They include doctors and experts of medical imaging and visualization, as well as several persons who are not familiar with the technologies of medical imaging and data visualization.
Table \ref{tab:VolVis} summarizes the answers from these ten surveyees.

There is also a late-returned survey form that was not included in the analysis. As a record, the answers in this survey form are:
1: c,
2: d,
3: (D),
4: a,
5: (a),
6: (A),
7: (c),
8: (B),
9: 5,
10: 4.
The upper case letters (always in brackets) are the most appropriate answers, while the lower case letters with brackets are acceptable answers as they are correct in some circumstances. The lower case letters without brackets are incorrect answers.

% =================================================
\section{Survey Results of Useful Knowledge in Viewing London Underground Maps}
\label{app:London}

\begin{figure}[t]
    \centering
    \includegraphics[width=\linewidth]{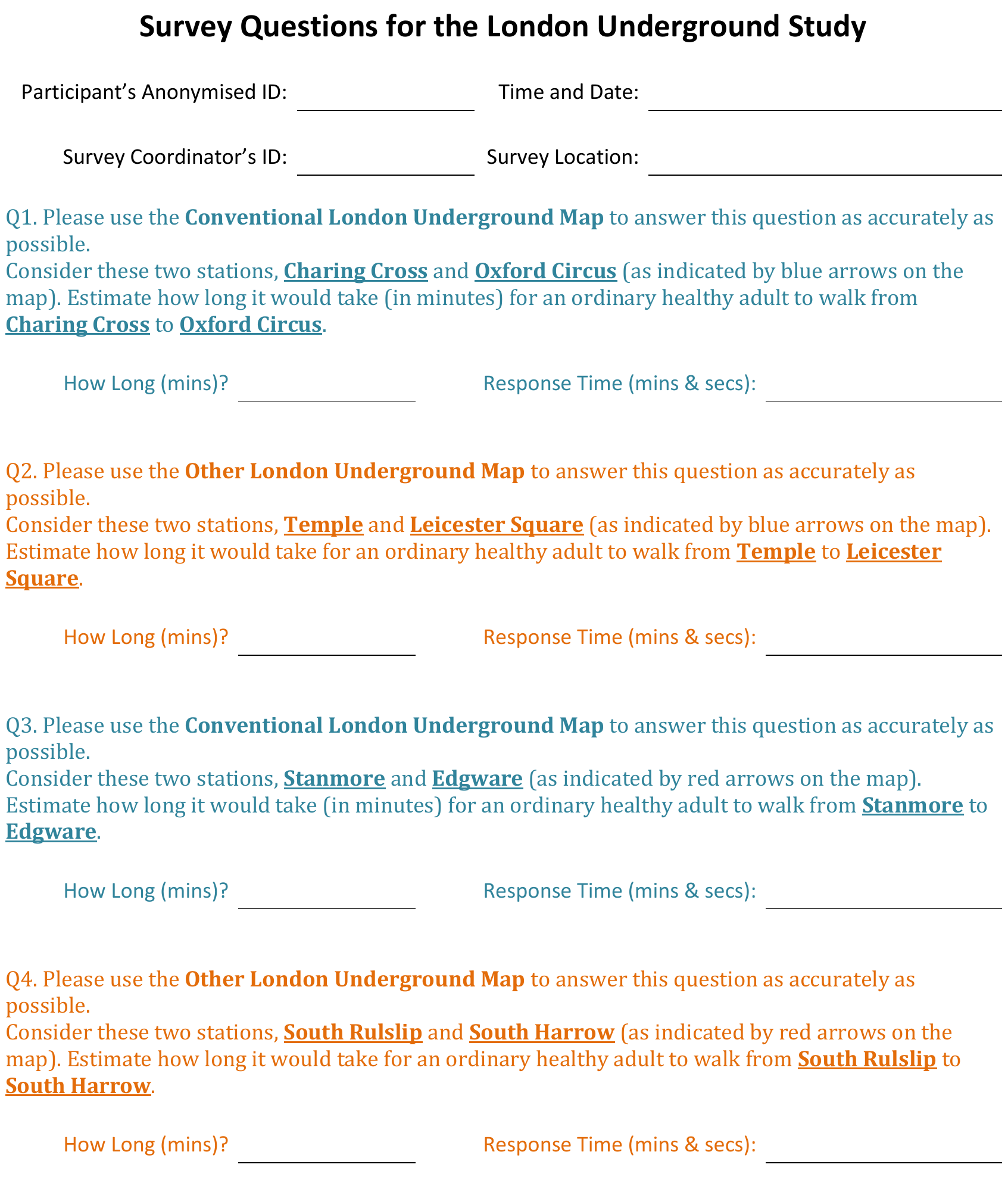}
    \caption{London underground survey: question sheet 1 (out of 3).}
    \label{fig:LondonSheet1}
    \vspace{-4mm}
\end{figure}

\begin{figure}[t]
    \centering
    \includegraphics[width=\linewidth]{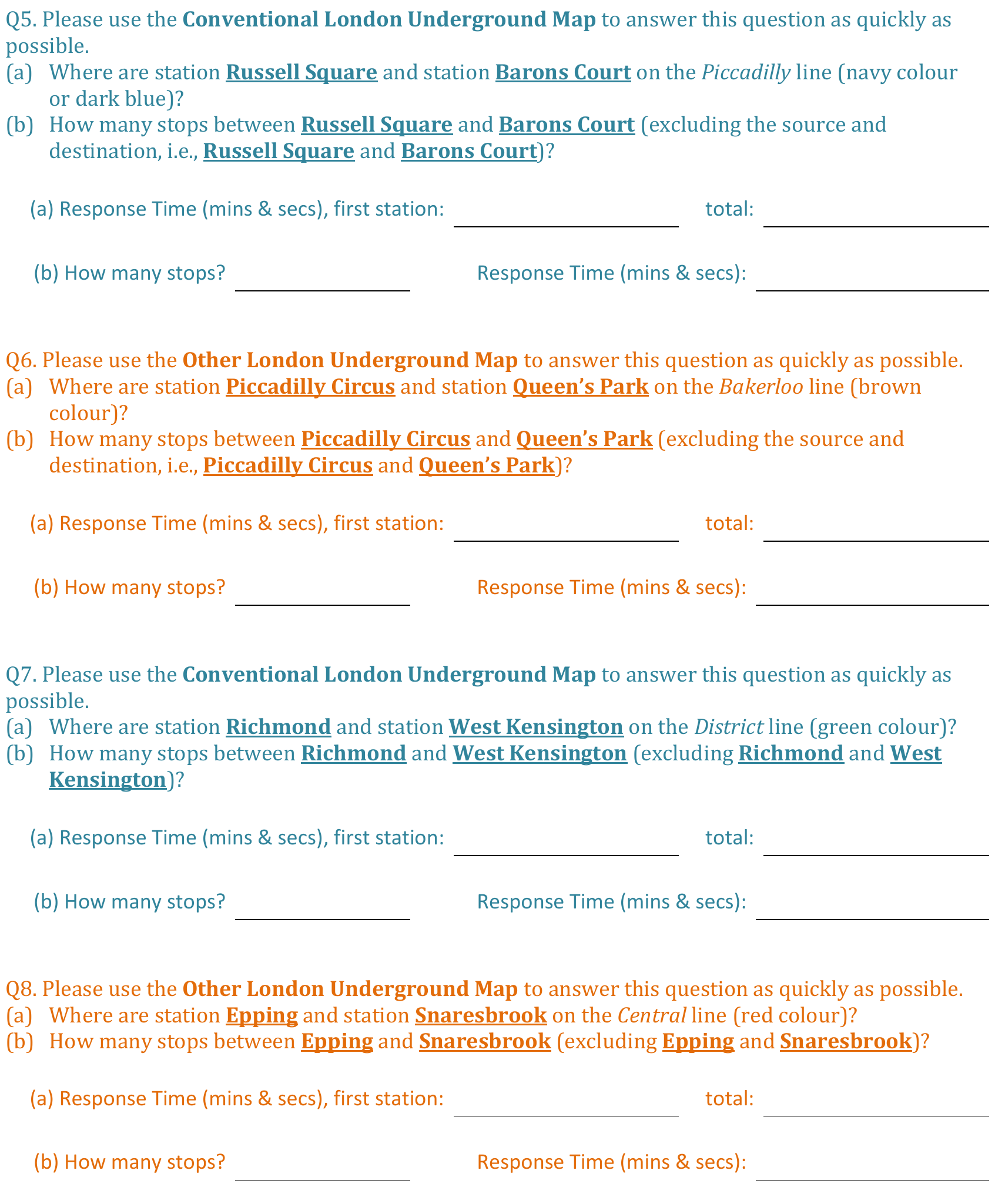}
    \caption{London underground survey: question sheet 2 (out of 3).}
    \label{fig:LondonSheet2}
\end{figure}

\begin{figure}[t]
    \centering
    \includegraphics[width=\linewidth]{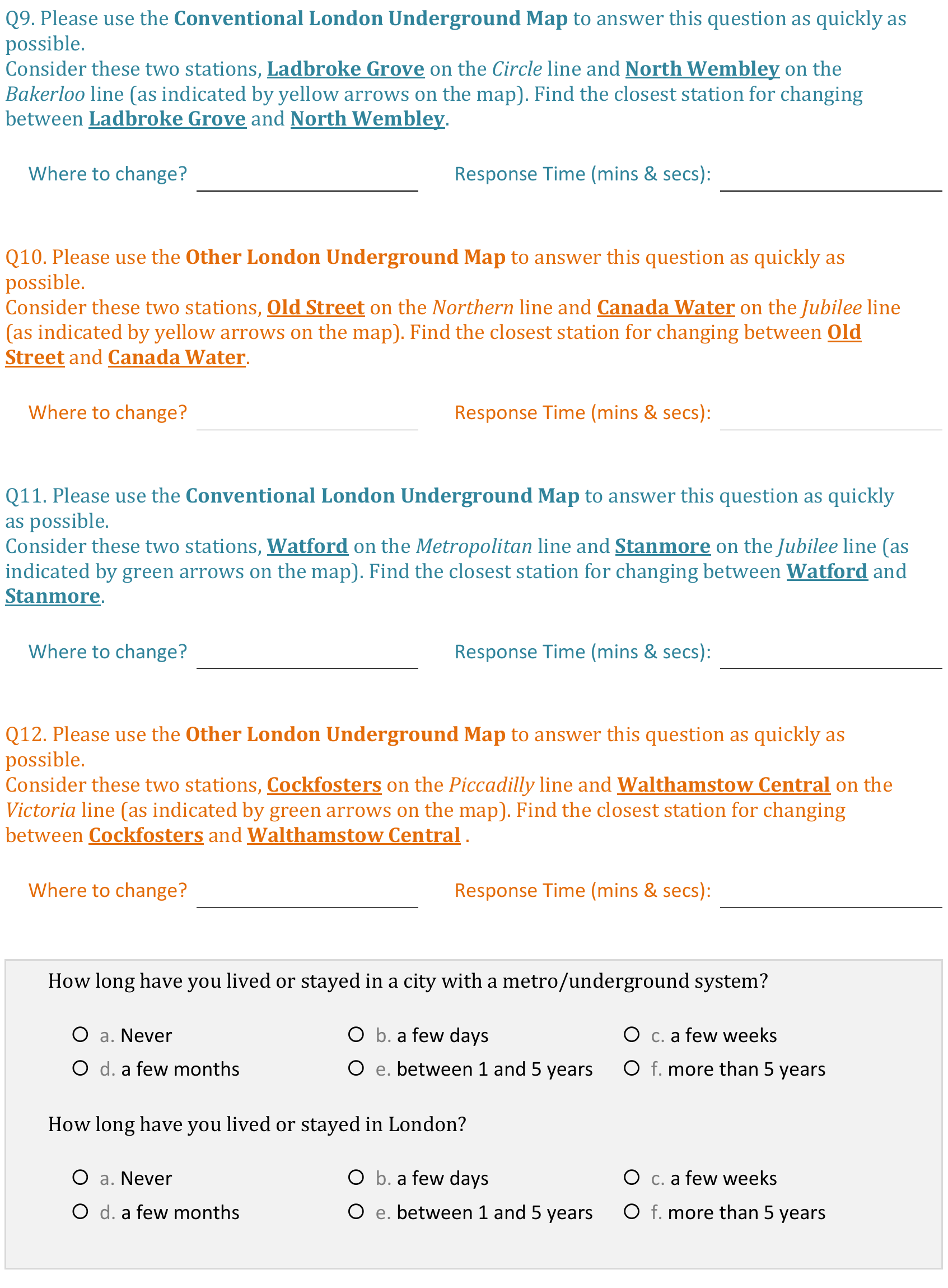}
    \vspace{-4mm}
    \caption{London underground survey: question sheet 3 (out of 3).}
    \label{fig:LondonSheet3}
    \vspace{-2mm}
\end{figure}

Figures \ref{fig:LondonSheet1}, \ref{fig:LondonSheet2}, and \ref{fig:LondonSheet3} show the questionnaire used in the survey about two types of London Underground maps.
Table \ref{tab:MapStudyKCL} summarizes the data from the answers by the 12 surveyees at King's College London, while Table \ref{tab:MapStudyOU} summarizes the data from the answers by the four surveyees at University Oxford.

In Section \ref{sec:London}, we have discussed Questions 1$\sim$4 in some detail.
In the survey, Questions 5$\sim$8 constitute the second set.   
Each question asks surveyees to first identify two stations along a given underground line, and then determine how many stops between the two stations.
All surveyees identified the stations correctly for all four questions, and most have also counted the stops correctly.
In general, for each of these cases, one can establish an alphabet of all possible answers in a way similar to the example of walking distances.
However, we have not observed any interesting correlation between the correctness and the surveyees' knowledge about metro systems or London.

With the third set of four questions, each questions asks surveyees to identify the closest station for changing between two given stations on different lines.
All surveyees identified the changing stations correctly for all questions.

\begin{figure}[t]
  \centering
  \includegraphics[width=80mm]{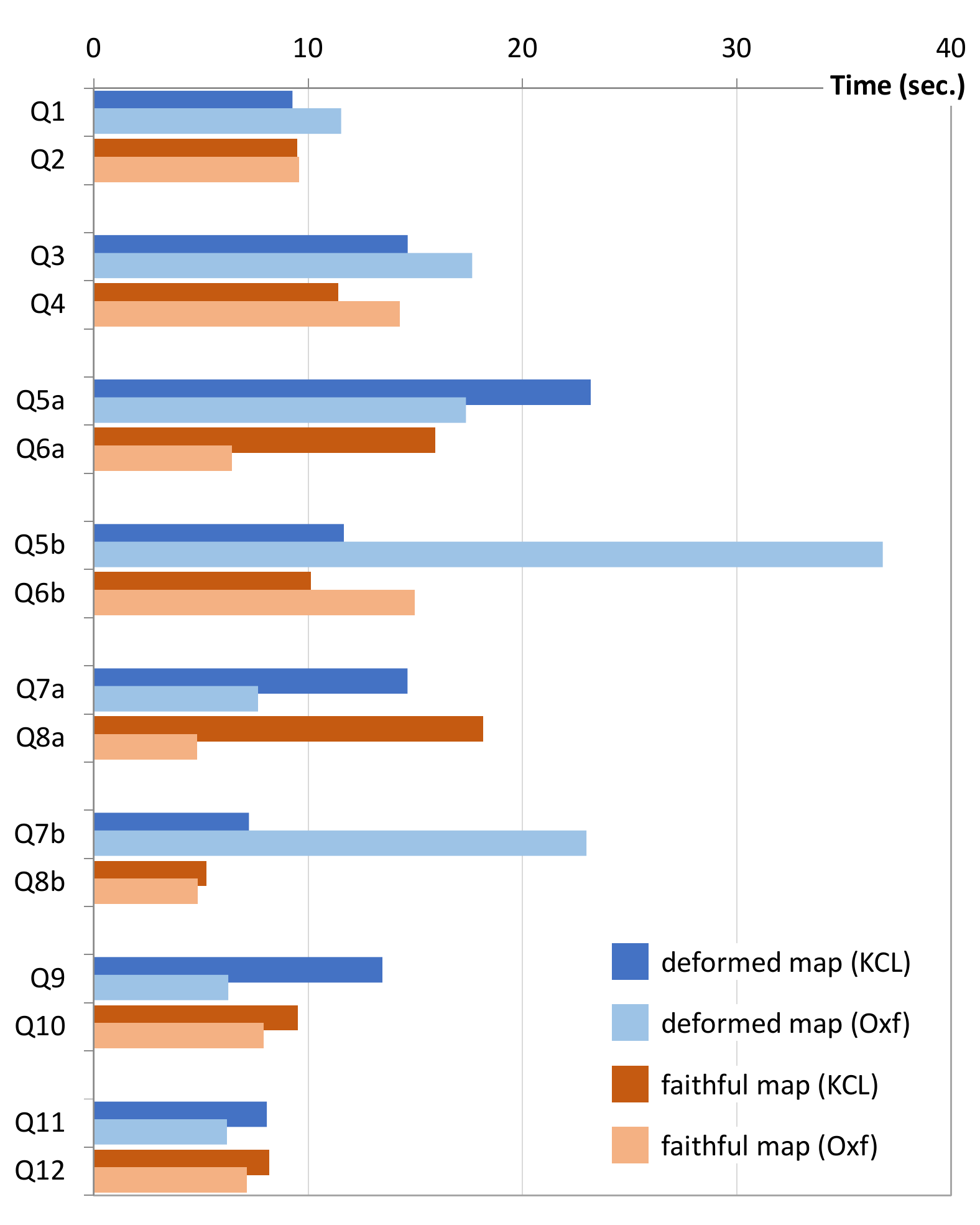}
  \vspace{-2mm}
  \caption{The average time used by surveyees for answering each of the 12 questions. The data does not indicate any significant advantage of using the geographically-deformed map.}
  \label{fig:MapStudy-RT}
  \vspace{-4mm}
\end{figure}

The design of Questions 5$\sim$12 was also intended to collect data that might differentiate the deformed map from the faithful map in terms of the time required for answering questions.
As shown in Figure \ref{fig:MapStudy-RT}, the questions were paired, such that the two questions feature the same level of difficulties.
Although the comparison seems to suggest that the faithful map might have some advantage in the setting of this survey, we cannot be certain about this observation as the sample size is not large enough.
In general, we cannot draw any meaningful conclusion about the cost in terms of time.
We hope to collect more real world data about the timing cost of visualization processes for making further advances in applying information theory to visualization.

Meanwhile, we consider that the space cost is valid consideration.
While both maps have a similar size (i.e., deformed map: 850mm$\times$580mm, faithful map: 840mm$\times$595mm), their font sizes for station labels are very different.
For long station names, ``High Street Kensington'' and ``Totteridge \& Whetstone'', the labels on the deformed map are of 35mm and 37mm in length, while those on the faithful map are of 17mm and 18mm long.
Taking the height into account, the space used for station labels in the deformed map is about four times of that in the faithful map.
In other worlds, if the faithful map were to display its labels with the same font size, the cost of the space would be four times of that of the deformed map.

\begin{table*}[t]
  \centering
  \caption{The answers by twelve surveyees at King's College London to the questions in the London underground survey.}
  \scalebox{1}{
  \begin{tabular}{@{}l@{}r@{\hspace{5mm}}c@{\hspace{3mm}}c@{\hspace{3mm}}c@{\hspace{3mm}}%
  c@{\hspace{3mm}}c@{\hspace{3mm}}c@{\hspace{3mm}}c@{\hspace{3mm}}c@{\hspace{3mm}}%
  c@{\hspace{3mm}}c@{\hspace{3mm}}c@{\hspace{3mm}}c@{\hspace{5mm}}r@{}}
  & & \multicolumn{12}{c}{\textbf{Surveyee's ID}}\\
  \multicolumn{2}{@{}l}{\textbf{Questions}} 
     & P1 & P2 & P3 & P4 & P5 & P6 & P7 & P8 & P9 & P10 & P11 & P12 & mean\\
  \hline
  Q1: & answer (min.) & 8 & 30 & 12 & 16 & 20 & 15 & 10 & 30
      & 20 & 20 & 20 & 30 & 19.25\\
      & time (sec.) & 06.22 & 07.66 & 09.78 & 11.66 & 03.72 & 04.85 & 08.85 & 21.12
      & 12.72 & 11.22 & 03.38 & 10.06 & 09.27\\
  \hline
  Q2: & answer (min.) & 15 & 30 & 5 & 22 & 15 & 14 & 20 & 20 
      & 25 & 25 & 25 & 20 & 19.67\\
      & time (sec.) & 10.25 & 09.78 & 06.44 & 09.29 & 12.12 & 06.09 & 17.28 & 06.75
      & 12.31 & 06.85 & 06.03 & 10.56 & 09.48\\
  \hline
  Q3: & answer (min.) & 20 & 45 & 10 & 70 & 20 & 20 & 20 & 35
      & 25 & 30 & 20 & 240 & 46.25\\
      & time (sec.) & 19.43 & 13.37 & 10.06 & 09.25 & 14.06 & 10.84 & 12.46 & 19.03
      & 11.50 & 16.09 & 11.28 & 28.41 & 14.65\\
  \hline
  Q4: & answer (min.) & 60 & 60 & 35 & 100 & 30 & 20 & 45 & 35
      & 45 & 120 & 40 & 120 & 59.17\\
      & time (sec.) & 11.31 & 10.62 & 10.56 & 12.47 & 08.21 & 07.15 & 18.72 & 08.91
      & 08.06 & 12.62 & 03.88 & 24.19 & 11.39\\
  \hline
  Q5: & time 1 (sec.) & 22.15 & 01.75 & 07.25 & 03.78 & 14.25 & 37.68 & 06.63 & 13.75
      & 19.41 & 06.47 & 03.41 & 34.97 & 14.29\\
      & time 2 (sec.) & 24.22 & 08.28 & 17.94 & 05.60 & 17.94 & 57.99 & 21.76 & 20.50
      & 27.16 & 13.24 & 22.66 & 40.88 & 23.18\\
      & answer (10) & 10 & 10 & 10 & 9 & 10 & 10 & 10 & 10
      & 9 & 10 & 10 & 10 \\
      & time (sec.) & 06.13 & 28.81 & 08.35 & 06.22 & 09.06 & 06.35 & 09.93 & 12.69
      & 10.47 & 05.54 & 08.66 & 27.75 & 11.66\\
  \hline
  Q6: & time 1 (sec.) & 02.43 & 08.28 & 01.97 & 08.87 & 05.06 & 02.84 & 06.97 & 10.15
      & 18.10 & 21.53 & 03.00 & 07.40 & 08.05\\
      & time 2 (sec.) & 12.99 & 27.69 & 04.81 & 10.31 & 15.97 & 04.65 & 17.56 & 16.31
      & 20.25 & 24.69 & 15.34 & 20.68 & 15.94\\
      & answer (9) & 9 & 10 & 9 & 9 & 4 & 9 & 9 & 9
      & 8 & 9 & 9 & 9\\
      & time (sec.) & 07.50 & 06.53 & 04.44 & 16.53 & 19.41 & 05.06 & 13.47 & 07.03
      & 12.44 & 04.78 & 07.91 & 16.34 & 10.12\\
  \hline
  Q7: & time 1 (sec.) & 17.37 & 08.56 & 01.34 & 03.16 & 08.12 & 01.25 & 21.75 & 15.56
      & 02.81 & 07.84 & 02.22 & 46.72 & 11.39\\
      & time 2 (sec.) & 17.38 & 13.15 & 02.34 & 03.70 & 08.81 & 02.25 & 22.75 & 26.00
      & 17.97 & 10.37 & 03.18 & 47.75 & 14.64\\
      & answer (7) & 7 & 7 & 7 & 7 & 6 & 7 & 7 & 7
      & 6 & 7 & 7 & 7\\
      & time (sec.) & 07.53 & 06.34 & 03.47 & 03.87 & 02.75 & 04.09 & 02.16 & 04.94
      & 26.88 & 05.31 & 06.63 & 12.84 & 07.23\\ 
  \hline
  Q8: & time 1 (sec.) & 12.00 & 08.50 & 06.09 & 02.88 & 08.62 & 14.78 & 19.12 & 08.53
      & 12.50 & 10.22 & 12.50 & 20.00 & 11.31\\
      & time 2 (sec.) & 13.44 & 10.78 & 23.37 & 09.29 & 13.03 & 36.34 & 23.55 & 09.50
      & 13.53 & 10.23 & 32.44 & 22.60 & 18.18\\
      & answer (6) & 6 & 6 & 6 & 6 & 6 & 6 & 6 & 6
      & 6 & 6 & 6 & 6\\
      & time (sec.) & 02.62 & 05.94 & 02.15 & 04.09 & 04.94 & 07.06 & 07.50 & 04.90
      & 04.37 & 04.53 & 05.47 & 09.43 & 05.25\\
  \hline
  Q9: & answer (P) & P & P & P & P & P & P & P & P
      & P & P & P & P & \\
      & time (sec.) & 35.78 & 02.87 & 07.40 & 13.03 & 06.97 & 52.15 & 13.56 & 02.16
      & 08.13 & 09.06 & 01.93 & 08.44 & 13.46\\
  \hline
  Q10: & answer (LB) & LB & LB & LB & LB & LB & LB & LB & LB
      & LB & LB & LB & LB & \\
      & time (sec.) & 05.50 & 03.13 & 12.04 & 14.97 & 07.00 & 26.38 & 11.31 & 03.38
      & 06.75 & 07.47 & 06.50 & 09.82 & 09.52\\
  \hline
  Q11: & answer (WP) & WP & WP & WP & WP & WP & WP & WP &WP
      & WP & WP & WP & WP & \\
      & time (sec.) & 06.07 & 05.35 & 07.72 & 05.00 & 04.32 & 23.72 & 05.25 & 03.07
      & 10.66 & 05.37 & 02.94 & 17.37 & 08.07\\
  \hline
  Q12: & answer (FP) & FP & FP & FP & FP & FP & FP & FP & FP
      & FP & FP & FP & FP & \\
      & time (sec.) & 05.16 & 02.56 & 11.78 & 08.62 & 03.60 & 19.72 & 11.28 & 03.94
      & 20.72 & 01.56 & 02.50 & 06.84 & 08.19\\
  \hline
  \multicolumn{2}{@{}l}{live in metro city} & $>$5yr & $>$5yr & mths & 1-5yr
    & $>$5yr & 1-5yr & weeks & $>$5yr & 1-5yr & $>$5yr & mths & mths\\ 
  \multicolumn{2}{@{}l}{live in London} & $>$5yr & $>$5yr & mths & 1-5yr
    & 1-5yr & mths & mths & mths & mths & mths & mths & mths\\ 
  \hline
  \end{tabular}
  }
  \label{tab:MapStudyKCL}
\end{table*}
\clearpage

\begin{table}[t]
  \centering
  \caption{The answers by four surveyees at University of Oxford to the questions in the London underground survey.}
  %\scalebox{1}{
  \begin{tabular}{@{}l@{}r@{\hspace{5mm}}c@{\hspace{3mm}}c@{\hspace{3mm}}%
  c@{\hspace{3mm}}c@{\hspace{5mm}}r@{}}
  & & \multicolumn{4}{c}{\textbf{Surveyee's ID}}\\
  \multicolumn{2}{@{}l}{\textbf{Questions}} 
     & P13 & P14 & P15 & P16 & mean\\
  \hline
  Q1: & answer (min.) & 15 & 20 & 15 & 15 & 16.25 \\
      & time (sec.) & 11.81 & 18.52 & 08.18 & 07.63 & 11.52 \\
  \hline
  Q2: & answer (min.) & 5 & 5 & 15 & 15 & 10.00\\
      & time (sec.) & 11.10 & 02.46 & 13.77 & 10.94 & 09.57 \\
  \hline
  Q3: & answer (min.) & 35 & 60 & 30 & 25 & 37.50 \\
      & time (sec.) & 21.91 & 16.11 & 10.08 & 22.53 & 17.66 \\
  \hline
  Q4: & answer (min.) & 20 & 30 & 60 & 25 & 33.75\\
      & time (sec.) & 13.28 & 16.21 & 08.71 & 18.87 & 14.27 \\
  \hline
  Q5: & time 1 (sec.) & 17.72 & 07.35 & 17.22 & 09.25 & 12.89 \\
      & time 2 (sec.) & 21.06 & 17.00 & 19.04 & 12.37 & 17.37\\
      & answer (10) & 10 & 8 & 10 & 10 & \\
      & time (sec.) & 04.82 & 02.45 & 02.96 & 15.57 & 06.45\\
  \hline
  Q6: & time 1 (sec.) & 35.04 & 38.12 & 11.29 & 07.55 & 23.00 \\
      & time 2 (sec.) & 45.60 & 41.32 & 20.23 & 40.12 & 36.82 \\
      & answer (9) & 9 & 10 & 9 & 8 & \\
      & time (sec.) & 03.82 & 13.57 & 08.15 & 34.32 & 14.97 \\
  \hline
  Q7: & time 1 (sec.) & 01.05 & 02.39 & 09.55 & 11.19 & 06.05 \\
      & time 2 (sec.) & 02.15 & 05.45 & 09.58 & 13.47 & 07.66 \\
      & answer (7) & 10 & 6 & 7 & 7 & \\
      & time (sec.) & 01.06 & 01.60 & 02.51 & 14.06 & 04.81 \\ 
  \hline
  Q8: & time 1 (sec.) & 08.74 & 26.14 & 20.37 & 15.01 & 17.57 \\
      & time 2 (sec.) & 16.50 & 30.55 & 27.01 & 17.91 & 22.99 \\
      & answer (6) & 6 & 6 & 6 & 6 \\
      & time (sec.) & 09.30 & 03.00 & 02.11 & 04.94 & 04.48 \\
  \hline
  Q9: & answer (P) & P & P & P & P & \\
      & time (sec.) & 05.96 & 09.38 & 04.56 & 05.16 & 06.27\\
  \hline
  Q10: & answer (LB) & LB & LB & LB & LB & \\
      & time (sec.) & 12.74 & 07.77 & 01.30 & 09.94 & 07.94 \\
  \hline
  Q11: & answer (WP) & WP & WP & WP & WP & \\
      & time (sec.) & 09.84 & 04.43 & 03.39 & 07.18 & 06.21 \\
  \hline
  Q12: & answer (FP) & FP & FP & FP & FP & \\
      & time (sec.) & 06.22 & 10.46 & 06.78 & 05.10 & 07.14 \\
  \hline
  \multicolumn{2}{@{}l}{live in metro city} & never & days & days & days \\ 
  \multicolumn{2}{@{}l}{live in London} & never & days & days & days \\ 
  \hline
  \end{tabular}
  %}
  \label{tab:MapStudyOU}
\end{table}

\begin{table}[h]
    \section{EuroVis 2020 Reviews and Revision Report}
    Please see the following 11 pages.\\
    \label{tab:Fix}
\end{table}

\vfill\break

\begin{figure*}[t]
    \centering
    \includegraphics[width=160mm]{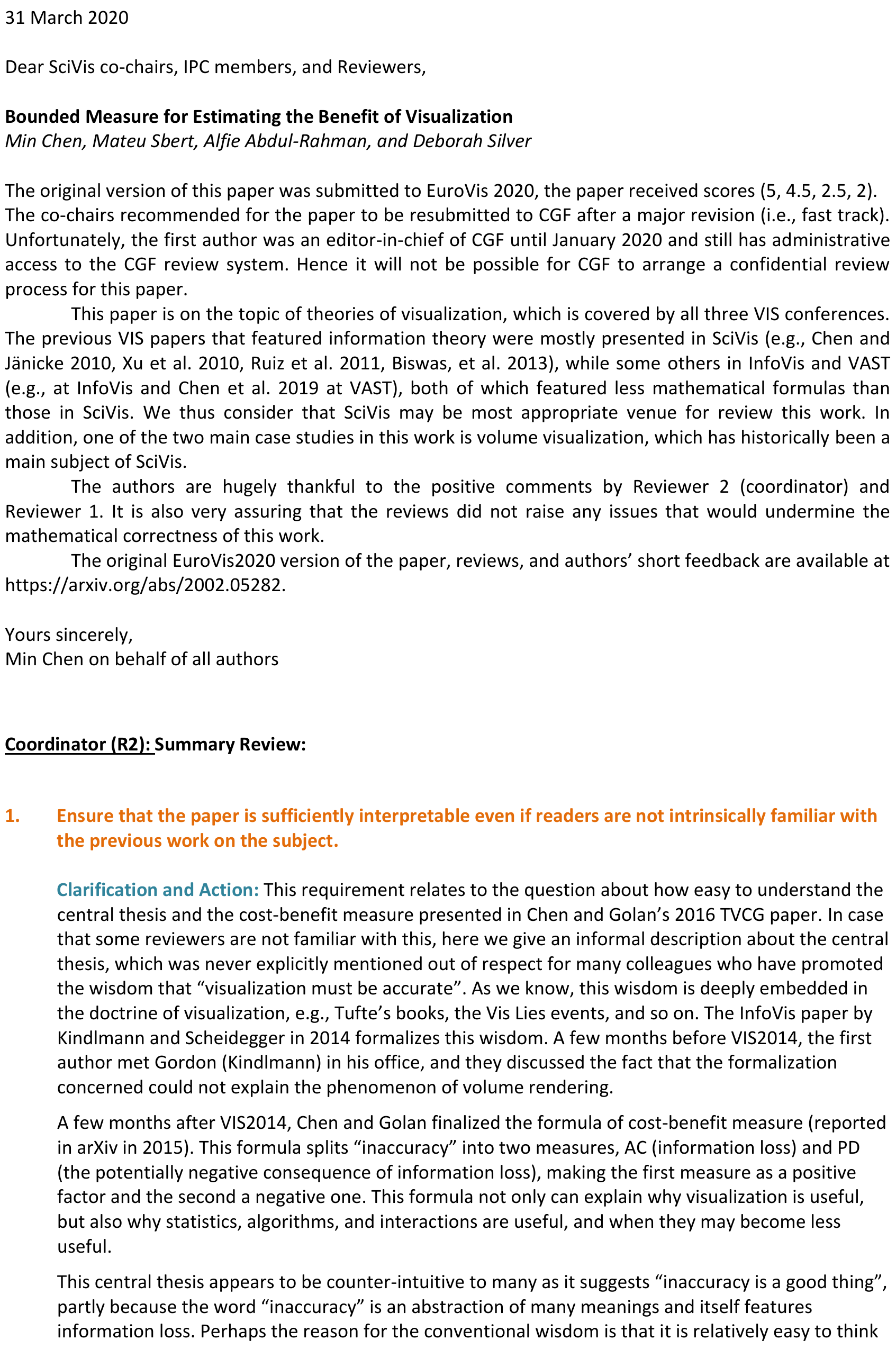}
    \label{ReportP1}
    \vspace{-8mm}
\end{figure*}
\clearpage

\begin{figure*}[t]
    \centering
    \includegraphics[width=160mm]{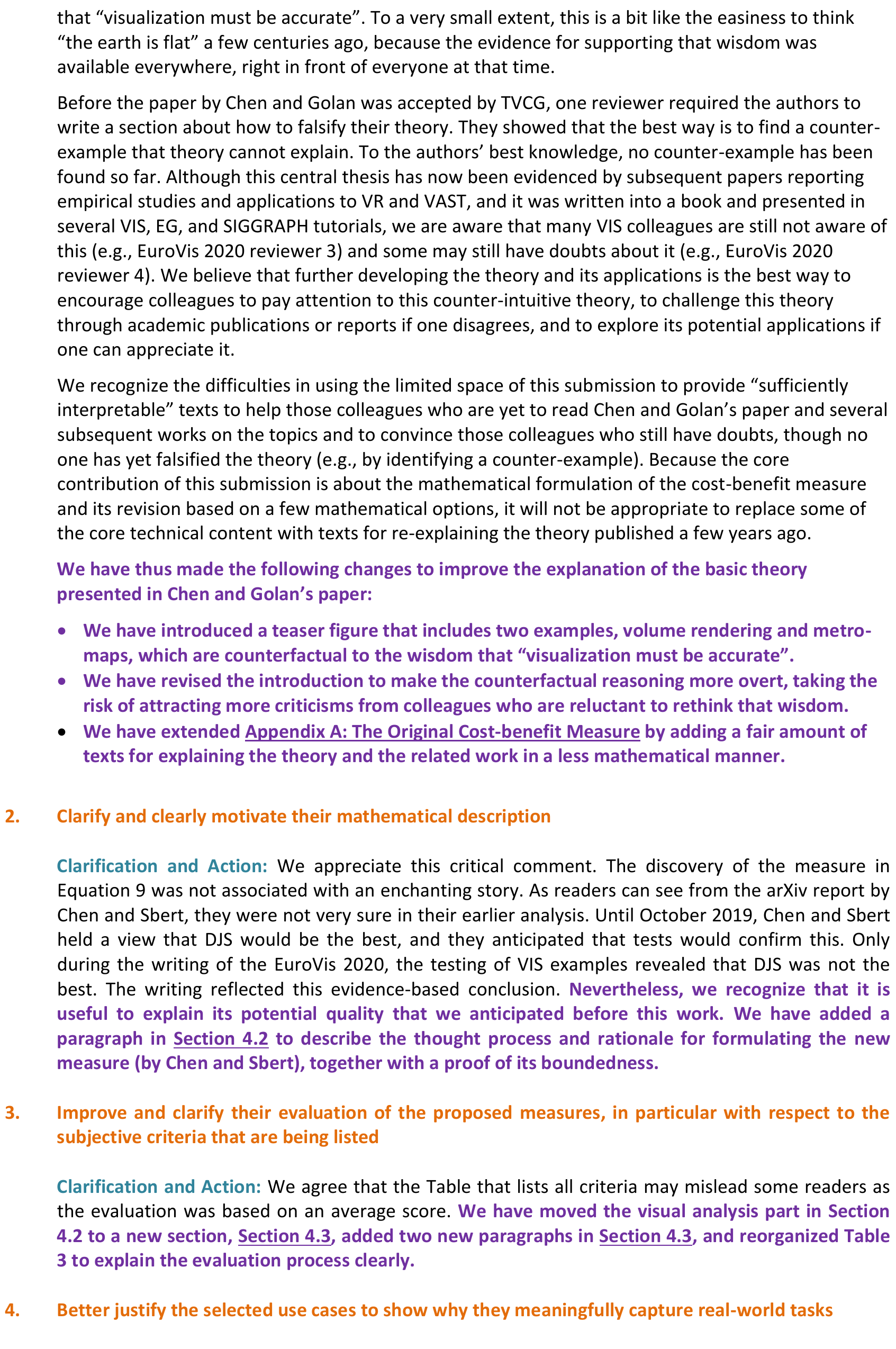}
    \label{ReportP2}
    \vspace{-8mm}
\end{figure*}
\clearpage

\begin{figure*}[t]
    \centering
    \includegraphics[width=160mm]{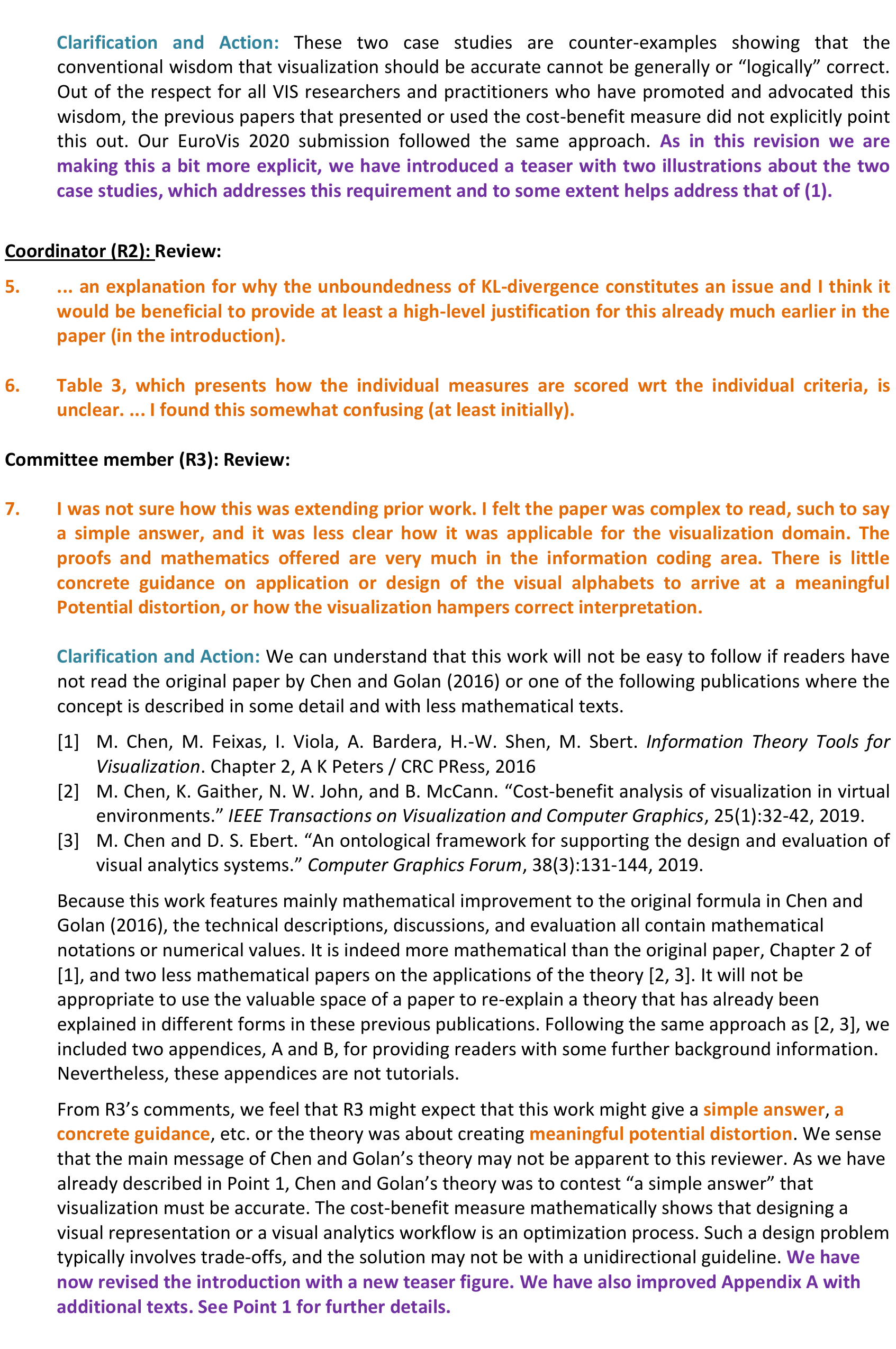}
    \label{ReportP3}
    \vspace{-8mm}
\end{figure*}
\clearpage

\begin{figure*}[t]
    \centering
    \includegraphics[width=160mm]{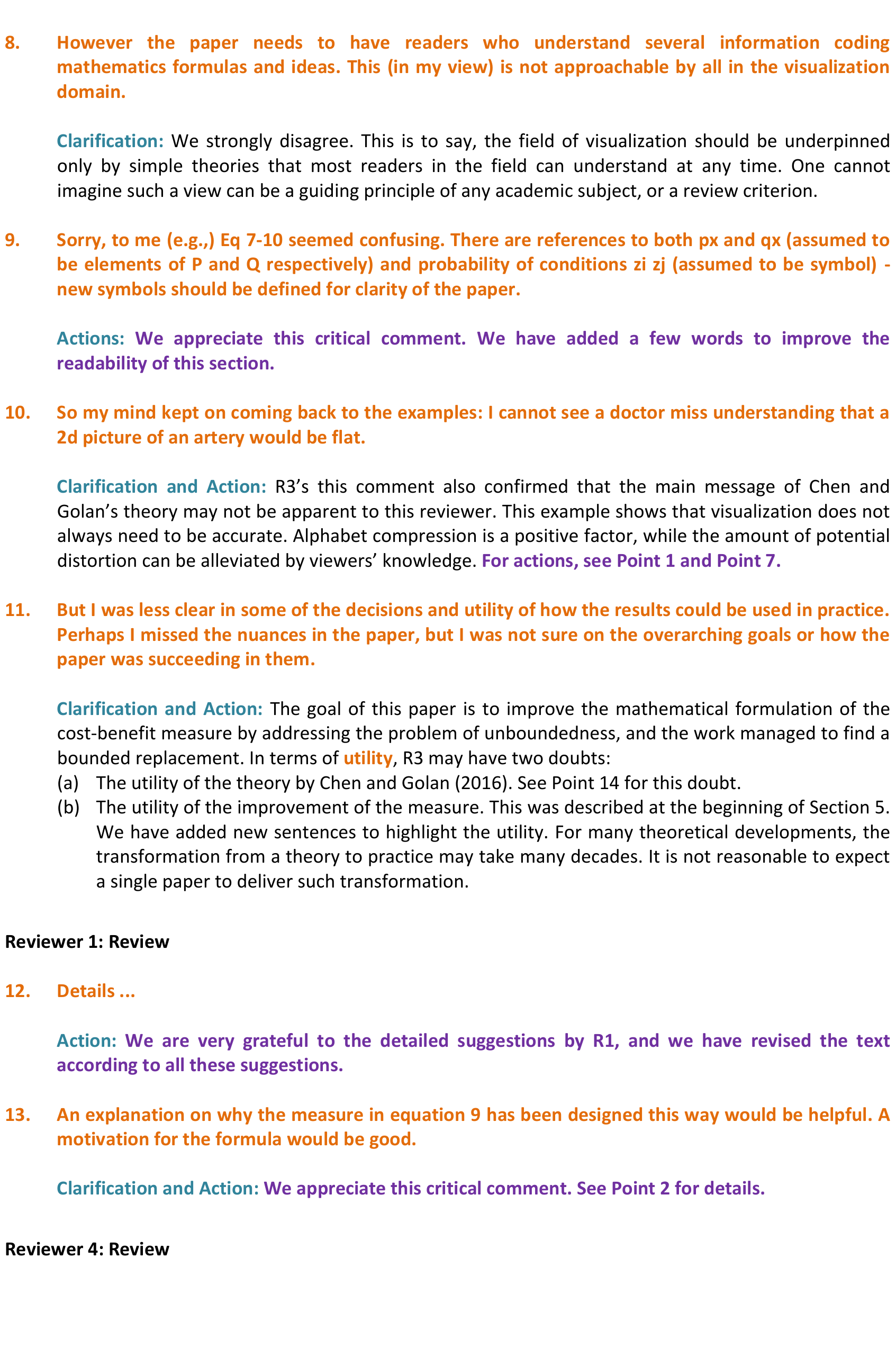}
    \label{ReportP4}
    \vspace{-8mm}
\end{figure*}
\clearpage

\begin{figure*}[t]
    \centering
    \includegraphics[width=160mm]{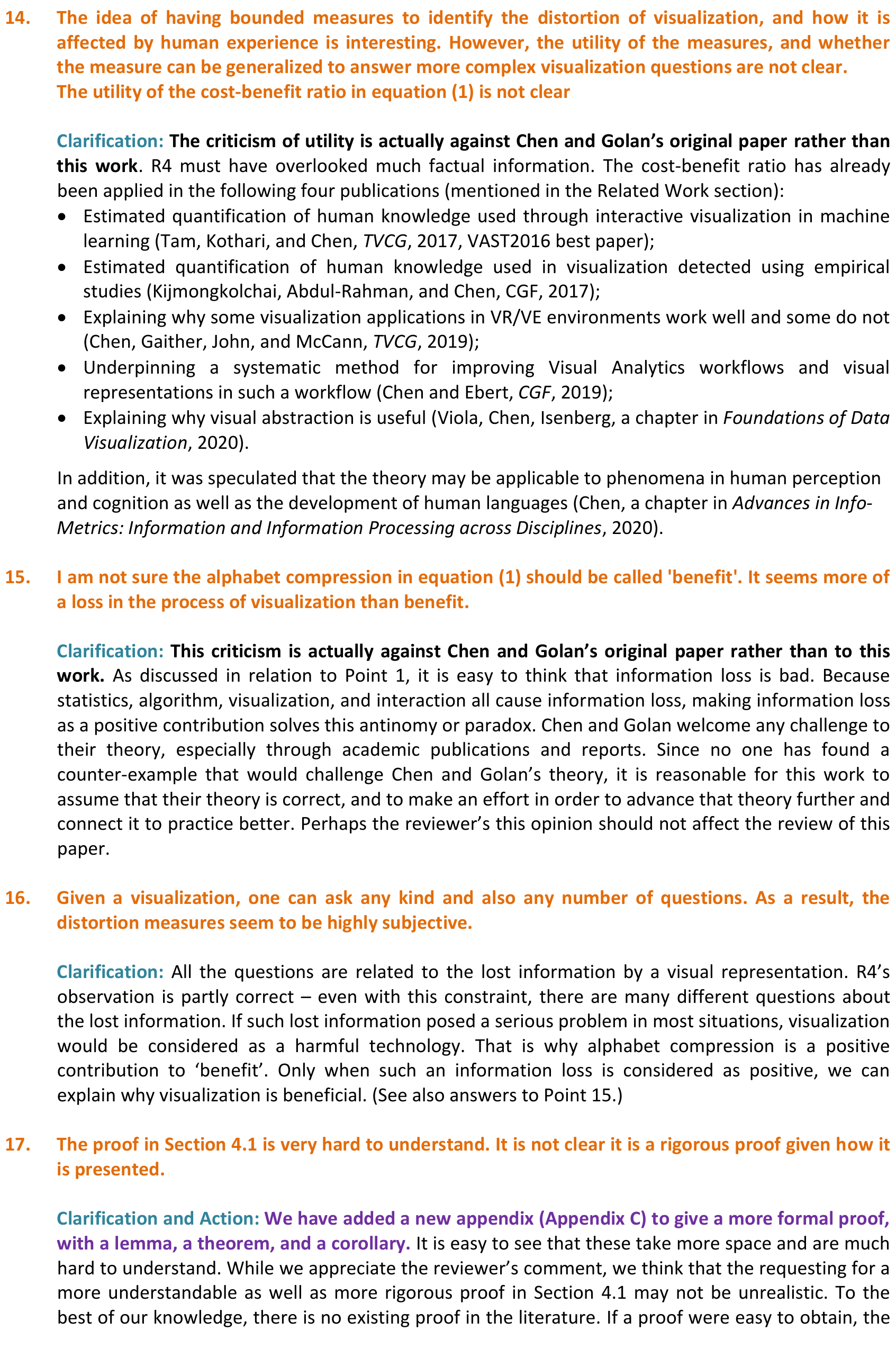}
    \label{ReportP5}
    \vspace{-8mm}
\end{figure*}
\clearpage

\begin{figure*}[t]
    \centering
    \includegraphics[width=160mm]{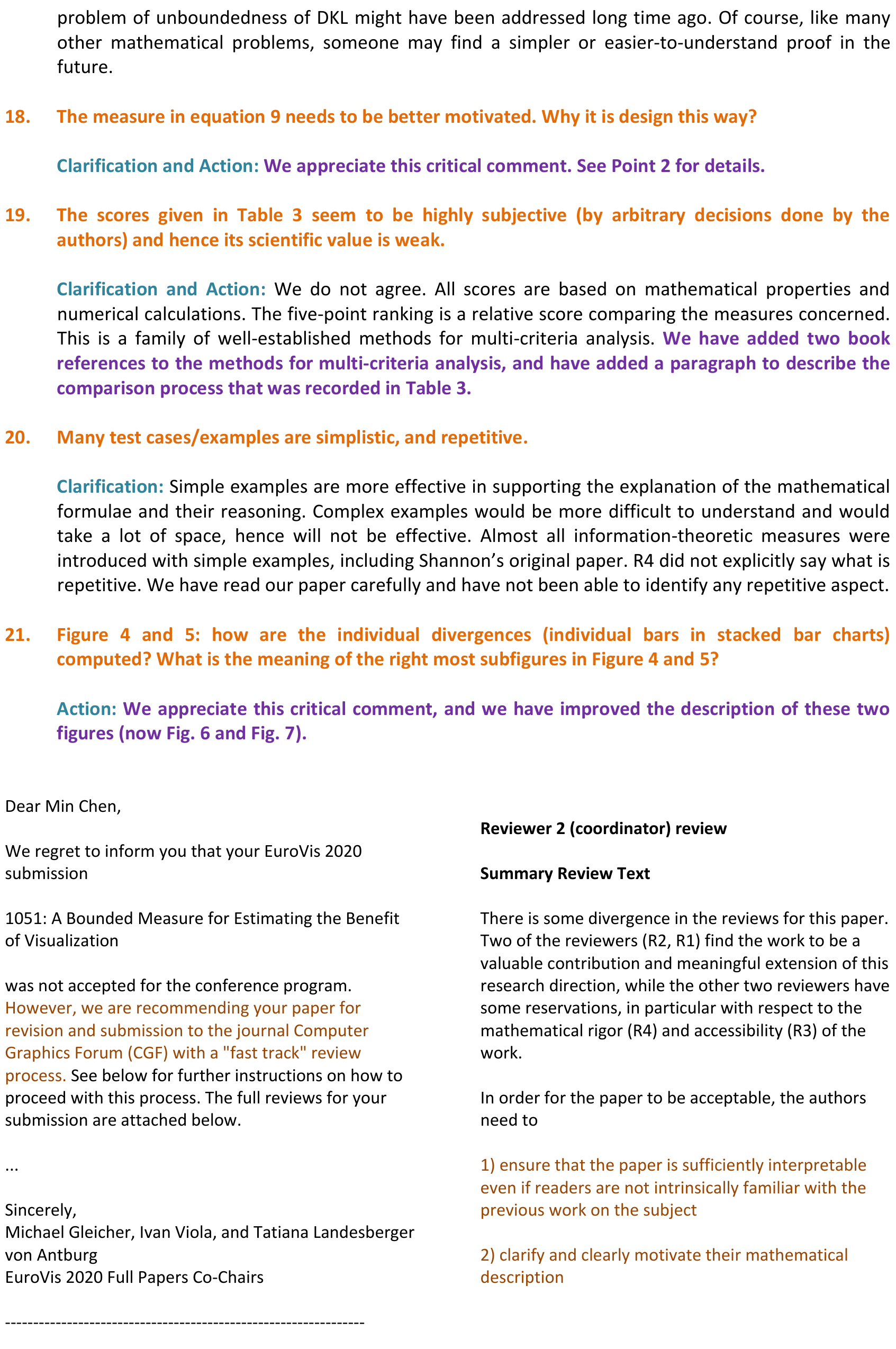}
    \label{ReportP6}
    \vspace{-8mm}
\end{figure*}
\clearpage

\begin{figure*}[t]
    \centering
    \includegraphics[width=160mm]{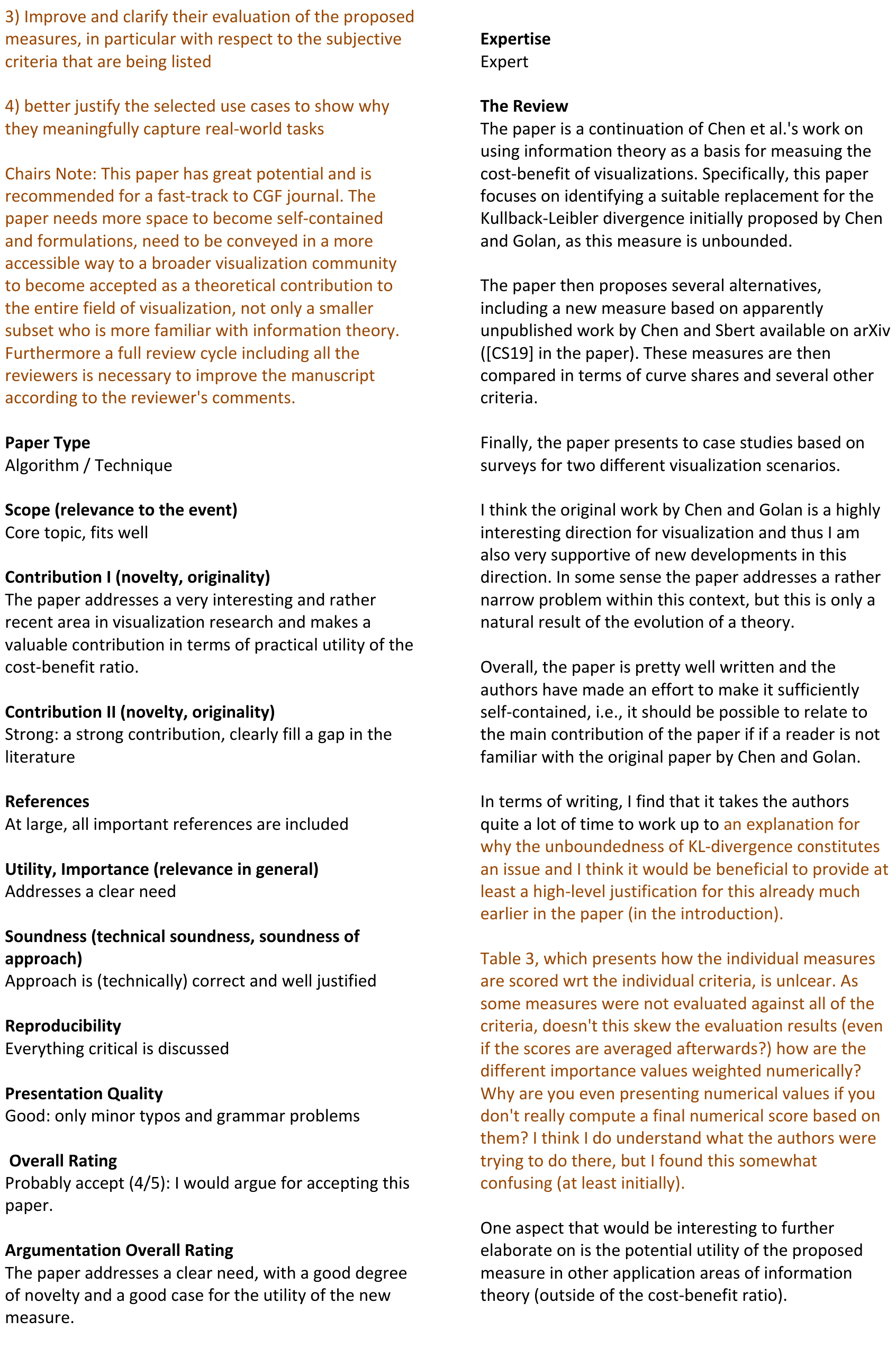}
    \label{ReportP7}
    \vspace{-8mm}
\end{figure*}
\clearpage

\begin{figure*}[t]
    \centering
    \includegraphics[width=160mm]{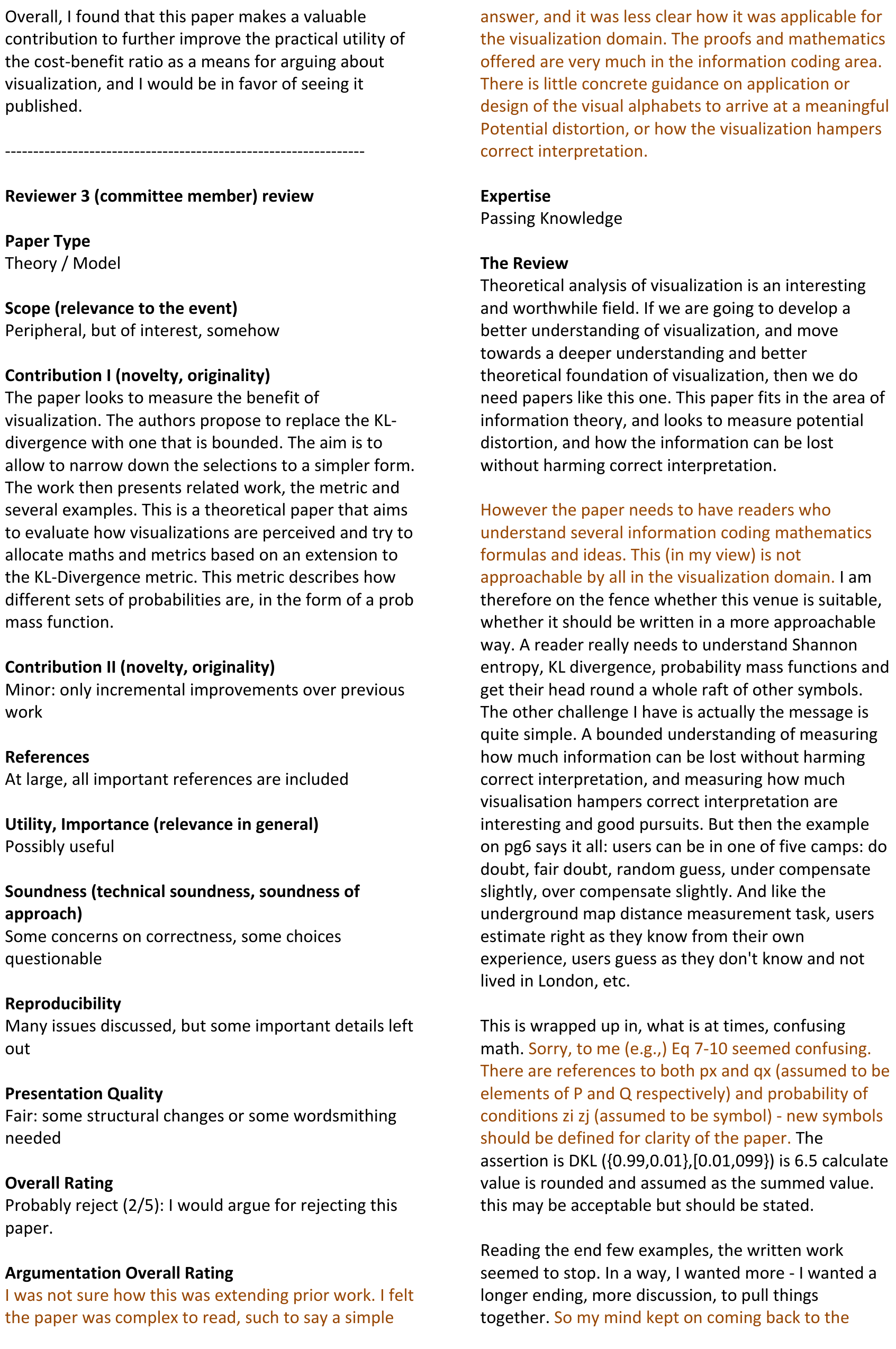}
    \label{ReportP8}
    \vspace{-8mm}
\end{figure*}
\clearpage

\begin{figure*}[t]
    \centering
    \includegraphics[width=160mm]{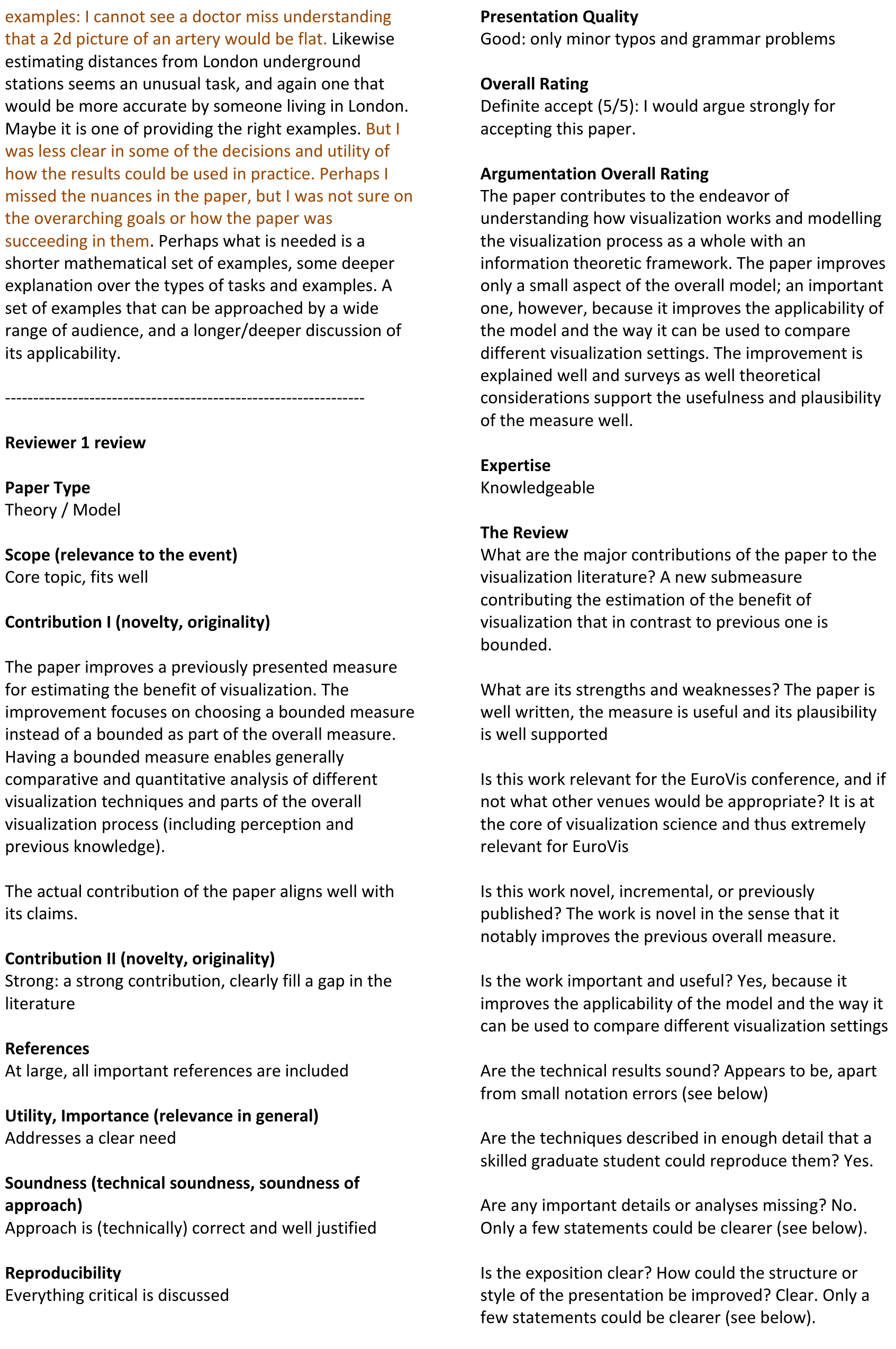}
    \label{ReportP9}
    \vspace{-8mm}
\end{figure*}
\clearpage

\begin{figure*}[t]
    \centering
    \includegraphics[width=160mm]{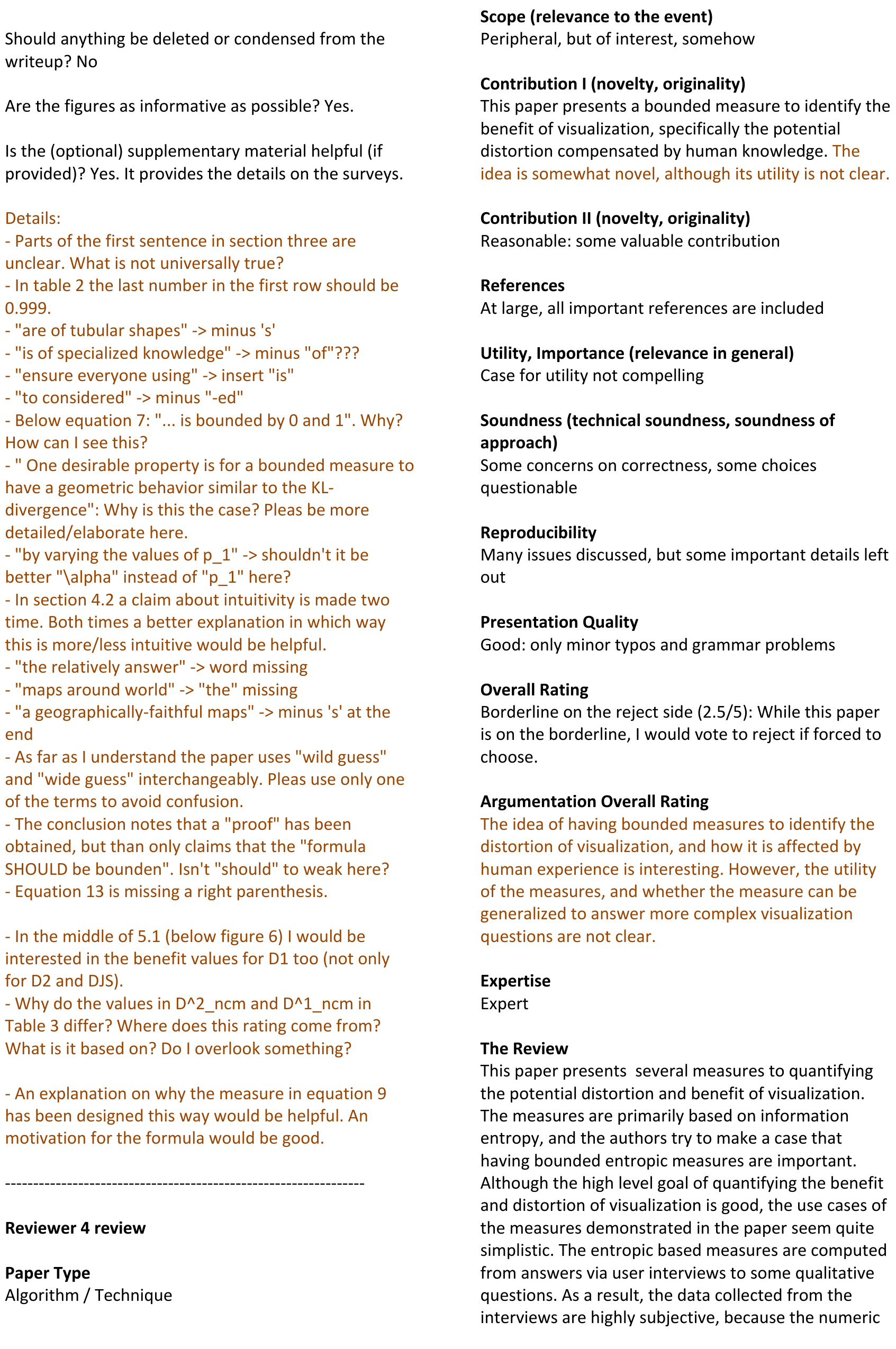}
    \label{ReportP10}
    \vspace{-8mm}
\end{figure*}
\clearpage

\begin{figure*}[t]
    \centering
    \includegraphics[width=160mm]{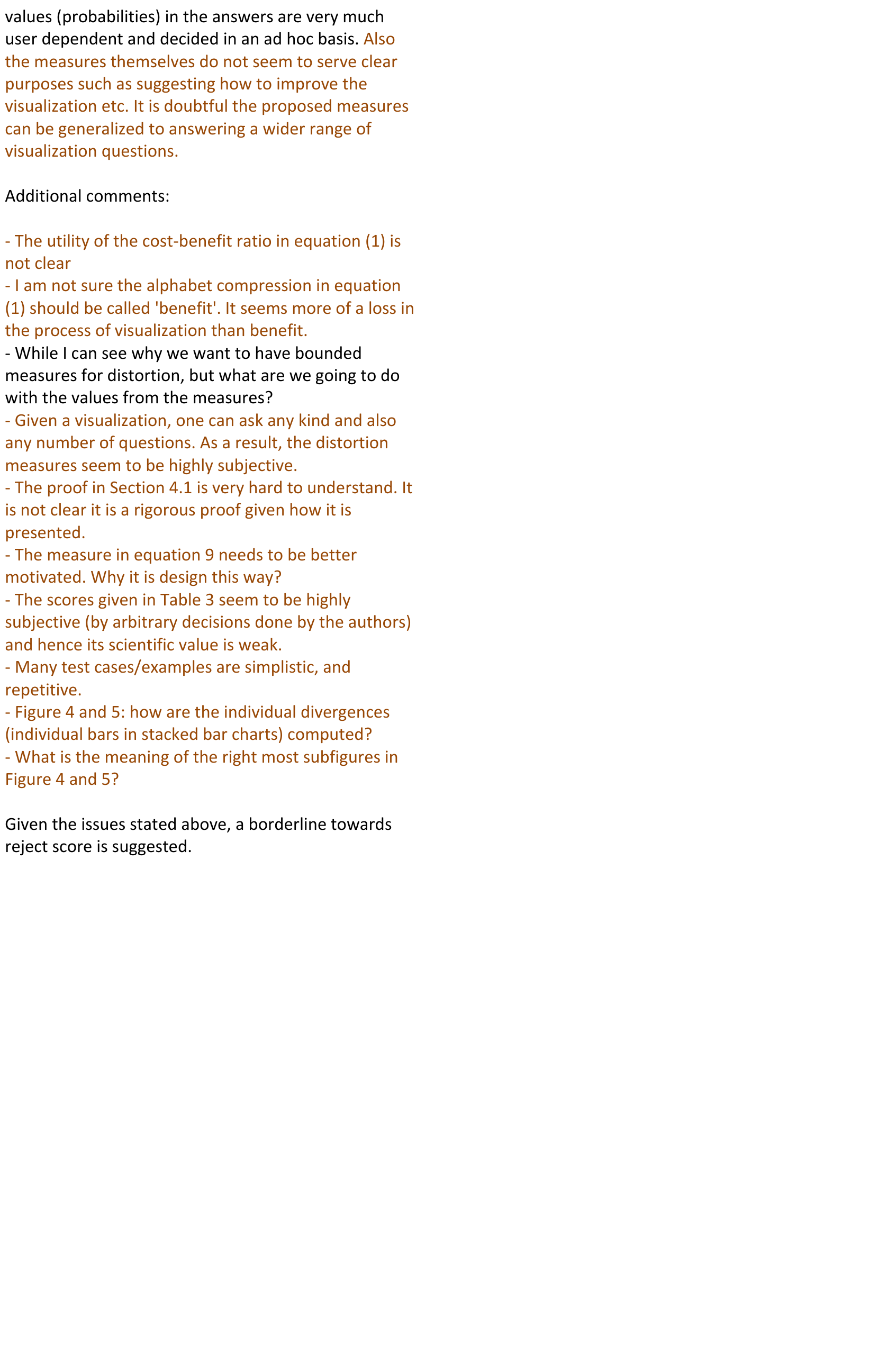}
    \label{ReportP11}
    \vspace{-8mm}
\end{figure*}
\clearpage

% ================

\end{document}